\documentclass[Afour,sageh,times]{sagej}

\usepackage{moreverb,url}

\usepackage[colorlinks,bookmarksopen,bookmarksnumbered,citecolor=red,urlcolor=red]{hyperref}

\usepackage[utf8]{inputenc} 
\usepackage[T1]{fontenc}    
\usepackage{hyperref}       
\usepackage{url}            
\usepackage[table]{xcolor}

\usepackage{booktabs}       
\usepackage{subcaption}
\usepackage{amsfonts}       
\usepackage{nicefrac}       
\usepackage{microtype}      
\usepackage{xcolor}         

\usepackage{times}
\usepackage{epsfig}
\usepackage{amssymb}
\usepackage{mathtools}
\usepackage{enumitem}
\usepackage{setspace}
\usepackage{booktabs}
\usepackage{color,xcolor}
\usepackage{bbding}
\usepackage{diagbox}
\usepackage{bm}
\usepackage{multirow}
\usepackage{caption}
\usepackage{makecell}
\usepackage{afterpage}
\usepackage{placeins}
\usepackage{float}
\usepackage{xspace}
\usepackage{natbib}

\newcommand\BibTeX{{\rmfamily B\kern-.05em \textsc{i\kern-.025em b}\kern-.08em
T\kern-.1667em\lower.7ex\hbox{E}\kern-.125emX}}

\newcommand{\mycolor}{black}

\newcommand{\nickname}{\mbox{FMimic}\xspace}

\def\volumeyear{2024}

\begin{document}


\title{\nickname: Foundation Models are Fine-grained Action Learners from Human Videos}

\author{
Guangyan Chen$^1$, Meiling Wang$^1$, Te Cui$^1$, Yao Mu$^2$, Haoyang Lu$^1$,  Zicai Peng$^1$, Mengxiao Hu$^1$, Tianxing Zhou$^1$, Mengyin Fu$^{1}$, Yi Yang$^{1*}$, Yufeng Yue$^{1*}$
}

\affiliation{\affilnum{1}Beijing Institute of Technology, Beijing, P. R. China
\\
\affilnum{2}The University of Hong Kong, Hong Kong, P. R. China
}

\corrauth{Yufeng Yue$^{*}$ and Yi Yang$^{*}$ are co-corresponding authors, 
Beijing Institute of Technology, China.}

\email{yueyufeng@bit.edu.cn, yi\_yang@bit.edu.cn}

\runninghead{The International Journal of Robotics Research}




\begin{abstract}
Visual imitation learning (VIL) provides an efficient and intuitive strategy for robotic systems to acquire novel skills. \textcolor{\mycolor}{Recent advancements in foundation models, particularly Vision Language Models (VLMs), have demonstrated remarkable capabilities in visual and linguistic reasoning for VIL tasks}. {Despite this progress, existing approaches primarily utilize these models for learning high-level plans from human demonstrations}, relying on pre-defined motion primitives for executing physical interactions, which remains a major bottleneck for robotic systems. In this work, \textcolor{\mycolor}{we present \nickname, a novel paradigm that harnesses foundation models to directly learn generalizable skills at even fine-grained action levels, using only a limited number of human videos}. Specifically, {our approach first grounds human-object movements from demonstration videos, then employs a skill learner to delineate motion properties through keypoints and waypoints, acquiring fine-grained action skills via hierarchical constraint representations}. 
\textcolor{\mycolor}{In unseen scenarios, the learned skills are updated through keypoint transfer and iterative comparison within the skill adapter}, enabling efficient skill adaptation. To achieve high-precision manipulation, the skill refiner optimizes the extracted and transferred interactions for enhanced precision, {while employing iterative master-slave contact refinement for pose estimation}, facilitating the acquisition and accomplishment of even highly constrained manipulation tasks. Our concise approach enables \nickname to effectively learn fine-grained actions from human videos, obviating the reliance on predefined primitives.
Extensive experiments demonstrate that our \nickname delivers strong performance with a single human video, and significantly outperforms all other methods with five videos. \textcolor{\mycolor}{Furthermore, our method exhibits significant improvements of over 39\% and 29\% in RLBench multi-task experiments and real-world manipulation tasks, respectively, and exceeds baselines by more than 34\% in high-precision tasks and 47\% in long-horizon tasks.} Code and videos are available on \href{https://fmimic-page.github.io/}{\textcolor{blue}{our homepage}}.
\end{abstract}

\keywords{{Visual imitation learning, Multimodal language models, Vision language models, Robotic manipulation, Code generation}}

\maketitle




\section{Introduction}
Visual Imitation Learning (VIL) has demonstrated remarkable efficacy in addressing various visual control tasks within intricate environments 
\citep{ mandlekar2020learning, tung2021learning, zeng2021transporter}.
{Diverging from conventional approaches that rely heavily on precisely labeled robot actions, which necessitates substantial human effort for data collection. Researchers increasingly turn to learning from readily available human demonstration videos to mitigate data collection challenges. }

\textcolor{\mycolor}{Existing methods for skill acquisition leveraging video data can be broadly categorized into two primary approaches}. {One typical approach develops efficient visual representations for robotic manipulation through self-supervised learning from large volumes of videos \citep{grauman2022ego4d, xiao2022masked, nair2022r3m, shaw2023videodex}.}
{Another approach focuses on learning task-relevant priors to guide robot behaviors or derive heuristic reward functions for reinforcement learning 
\citep{bahl2022human, shaw2023videodex, bahl2022human, sieb2020graph}.}
\textcolor{\mycolor}{However, these approaches often encounter challenges in both learning and executing precise manipulations, and struggle to generalize the learned skills to unseen environments. Therefore, efficiently acquiring precise and generalizable skills from limited videos remains a significant challenge.}

\begin{figure*}[t!]
\setlength{\abovecaptionskip}{-0.07cm}
\begin{center}
\includegraphics[width=\textwidth]{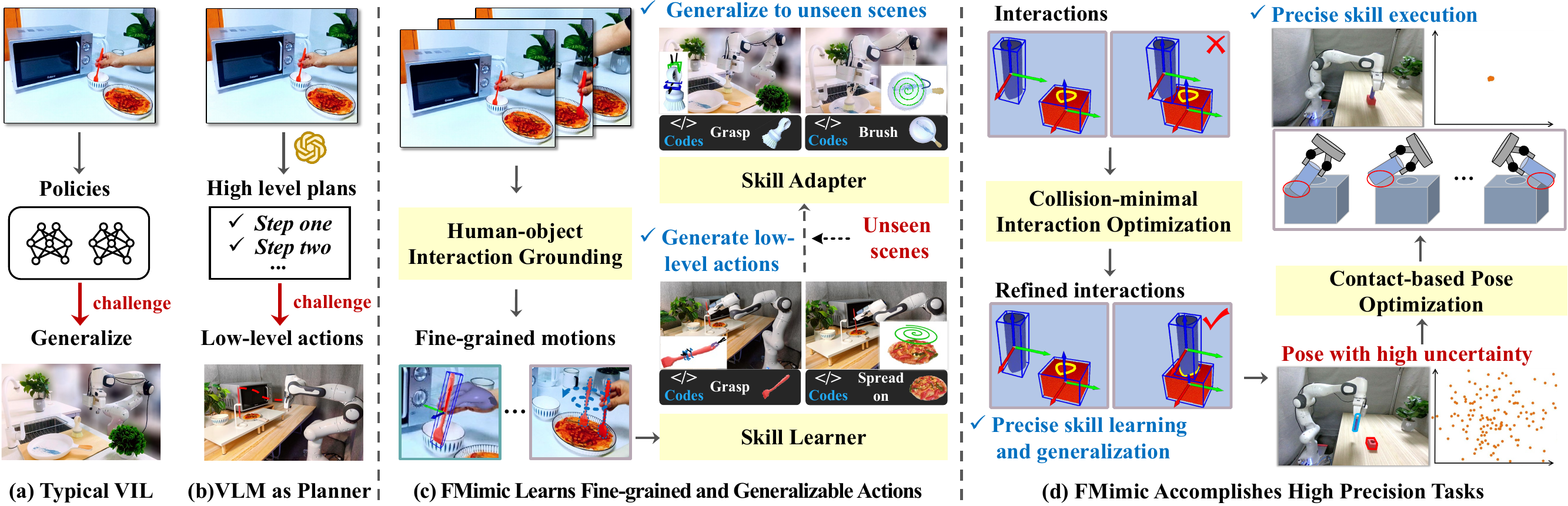}
\end{center}
\caption{{ Illustration of our \nickname.}  \textcolor{black}{(a) Typical VIL methods struggle to generalize to unseen environments, and (b) current methods naively utilize VLMs as planners, encounter difficulties in generating low-level actions. (c) \nickname grounds human videos to obtain action movements, and learns skills with fine-grained actions, while the skill adapter updates skills for generalization. (d) \nickname optimizes the interactions and the pose estimation results, accomplishing even high-precision tasks.} }
\label{Figure1}
\end{figure*}

{Foundation models present a compelling avenue for learning generalizable skills by leveraging their extensive prior knowledge derived from broad data.}
\textcolor{\mycolor}{Recent advances in vision-language models (VLMs) offer particularly promising tools in this regard, with their emergent conceptual understanding, commonsense knowledge, and sophisticated reasoning abilities}. {However, current VIL methods 
\citep{chen2023human, wake2023gpt,patel2023pretrained}
merely utilize VLMs for high-level planning while remaining dependent on pre-defined motion primitives}. This reliance on individual skill acquisition is often considered a major bottleneck due to the lack of large-scale robotic data. \textcolor{\mycolor}{
This limitation prompts a fundamental question}: \textit{Can we leverage foundation models, such as VLMs, to learn even at the fine-grained action level directly from human videos, eliminating the reliance on predefined primitives?} 
\textcolor{\mycolor}{Beyond skill acquisition and generalization, the ability to solve tasks with high precision remains a critical requirement in robotic applications. This requirement, however, persists as an open research challenge for current VIL methods. This observation leads to our second question}: \textit{Can we leverage the acquired actions to accomplish even high-precision tasks?}

\textcolor{\mycolor}{To answer the first question, we exploit the application of VLMs in visual imitation learning for fine-grained actions. However, several key challenges must be addressed: }
(\uppercase\expandafter{\romannumeral1}) {Despite recent advancements in VLMs, their capability to recognize and interpret low-level actions in video sequences remains limited}. To overcome this obstacle, {we propose a human-object interaction grounding module that estimates human-object motions for subsequent analysis. This approach effectively transforms the complex action recognition task into a pattern reasoning problem, which aligns more naturally with the current capabilities of VLMs.}
(\uppercase\expandafter{\romannumeral2})  
\textcolor{\mycolor}{Motion data inherently contains substantial redundancy, which hinders models from extracting valuable information.} 
To overcome this challenge, grounded motions are distilled into keypoints and waypoints to compactly capture motion properties. {These elements are then analyzed by VLMs through hierarchical constraint representations.
%
}
This approach effectively reduces redundancy and facilitates a comprehensive understanding, {enabling our method to effectively learn skills from a limited set of human videos}.
(\uppercase\expandafter{\romannumeral3}) Demonstration and execution scenes may involve different objects and tasks, impeding direct skill transfer. To this end, a skill adapter is proposed, which accurately transfers keypoints to novel objects via the region-to-keypoint mapping approach, thus generalizing the learned affordances to the unseen environments. \textcolor{\mycolor}{Furthermore, it updates skills with an iterative comparison strategy, which iteratively contrasts with the demonstrated knowledge, facilitating the adaptation of learned skills to novel scenes.}

\textcolor{black}{
{Building on this, we take a step further to answer the second question for achieving high-precision task execution}. \textcolor{\mycolor}{However, two critical limitations emerge: The limited precision of grounded and transferred interactions impedes accurate skill learning and generalization}. Furthermore, {pose estimation based on visual recognition often fails to achieve the requisite precision for accurate skill execution.}
To enhance the effectiveness of \nickname in high-precision tasks,
we propose a skill refiner with contact-aided optimization, {which elevates the previous vision-centric framework into a multi-modal architecture, seamlessly integrating visual and contact  feedback. Specifically, our collision-minimal interaction optimization approach is designed to refine both grounded and transferred interactions}, enabling the acquisition and generalization of skills applicable to high-precision tasks with stringent constraints. \textcolor{\mycolor}{The iterative contact-based pose estimation procedure then optimizes the perceived relative poses directly}, achieving effective pose optimization and precise skill execution. Additionally, {we propose an information gain maximization-based contact selection method}, optimizing the effectiveness of individual contact iterations and enhancing generalization capabilities for unfamiliar objects.
}

Based on the above analysis, \textcolor{\mycolor}{we present \nickname, a novel approach that leverages foundation models to directly learn even fine-grained action levels from a limited number of human videos, and generalize these skills to diverse novel scenes}. As shown in Figure \ref{Figure1}, {our method parses videos into multiple segments and precisely captures corresponding movements using the human-object interaction grounding module}. \textcolor{black}{Subsequently, a skill learner identifies keypoints and waypoints from grounded motions, then extracts knowledge employing hierarchical constraint representations, deriving skills with fine-grained actions}. {When deployed in novel environments}, \textcolor{black}{a skill adapter transfers extracted keypoints to novel environments in a region-to-keypoint mapping manner, and the learned skills are revised and updated with an iterative comparison strategy.}  
\textcolor{\mycolor}{To further enhance the ability of \nickname in even high-precision tasks, the skill refiner optimizes both extracted and transferred interactions for improved precision}, and the pose estimation results are optimized through iterative master-slave contact, {enabling \nickname to effectively learn and execute even highly constrained manipulation tasks.
Consequently, our \nickname demonstrates effectiveness in acquiring even complex manipulation skills from a limited number of human videos, and robustly generalizes them to novel scenes.}
Our method outperforms other methods by over 39\% on the RLBench multi-task experiments. In real-world manipulation tasks, \nickname achieves an improvement exceeding 29\% in seen environments, 38\% in unseen environments, and 32\% in unseen tasks. \textcolor{\mycolor}{Furthermore, it excels in high-precision and long-horizon tasks, demonstrating substantial enhancements of 34\% and 47\%.}

{A previous version was accepted at NeurIPS 2024 \citep{chen2024VLMimic}. This work substantially extends the conference version, introducing several significant advancements:(\uppercase\expandafter{\romannumeral1})}
We advance from the object-centric paradigm to an expressive keypoint-centric paradigm, facilitating skill acquisition and generalization. \textcolor{\mycolor}{To achieve this, we develop a keypoint-waypoint extraction approach to distill essential interactions, and propose a region-to-keypoint mapping strategy for precise keypoint transfer.}
(\uppercase\expandafter{\romannumeral2}) We elevate the vision-based framework into a comprehensive multi-modal architecture, integrating visual and contact feedback through our proposed skill refiner. {It refines the interactions to enable precise skill acquisition, and optimizes pose estimation results through contact iterations. }(\uppercase\expandafter{\romannumeral3}) \textcolor{\mycolor}{We broaden the application scope of \nickname beyond daily tasks to  encompass high-precision tasks}. Extensive experiments are conducted across all 8 benchmarks, complemented by detailed ablation studies and robustness analyses, to provide a comprehensive understanding of \nickname.

\section{Related Work}
\subsection{Learning from Human videos}
\textcolor{black}{Conventional learning approaches necessitate access to expert demonstrations, encompassing observations and precise actions for each timestep. {Drawing inspiration from human cognitive processes, learning from observation presents efficient and intuitive methods for robots to develop new skills}. \textcolor{\mycolor}{Recent research has extensively explored the utilization of large-scale human video data to improve robot policy learning 
\citep{grauman2022ego4d, xiao2022masked, nair2022r3m, shaw2023videodex}}.}
Representative methods, R3M \citep{nair2022r3m} and MVP \citep{xiao2022masked}, {which employ the internet-scale Ego4D dataset \citep{grauman2022ego4d} to develop visual representations for downstream imitation learning tasks}. {Another thread of work \citep{bahl2022human, sieb2020graph, sharma2019third, kumar2023graph, smith2019avid, chen2025unify, chen2025graphMimic} 
focuses on extracting task-relevant priors from videos to guide robot behaviors or derive heuristic reward functions for reinforcement learning.}
Learning by watching \citep{xiong2021learning} learns human-to-robot translation, and the resulting representations are used to guide robots to learn robotic skills. 
{WHIRL \citep{bahl2022human} infers trajectories and interaction details to establish a prior, though it still relies on real-world exploration for policy learning and requires a large number of iterations to converge.}
\textcolor{black}{GraphIRL \citep{kumar2023graph} employs graph abstraction on the videos followed by temporal matching to measure the task progress}, and a dense reward function is employed to train reinforcement learning algorithms. \textcolor{\mycolor}{Despite these advancements in the field, efficiently acquiring generalizable skills from limited demonstration videos remains a significant challenge.}

\begin{figure*}[t]
\setlength{\abovecaptionskip}{-0.13cm}
\begin{center}
\includegraphics[width=0.99\textwidth]{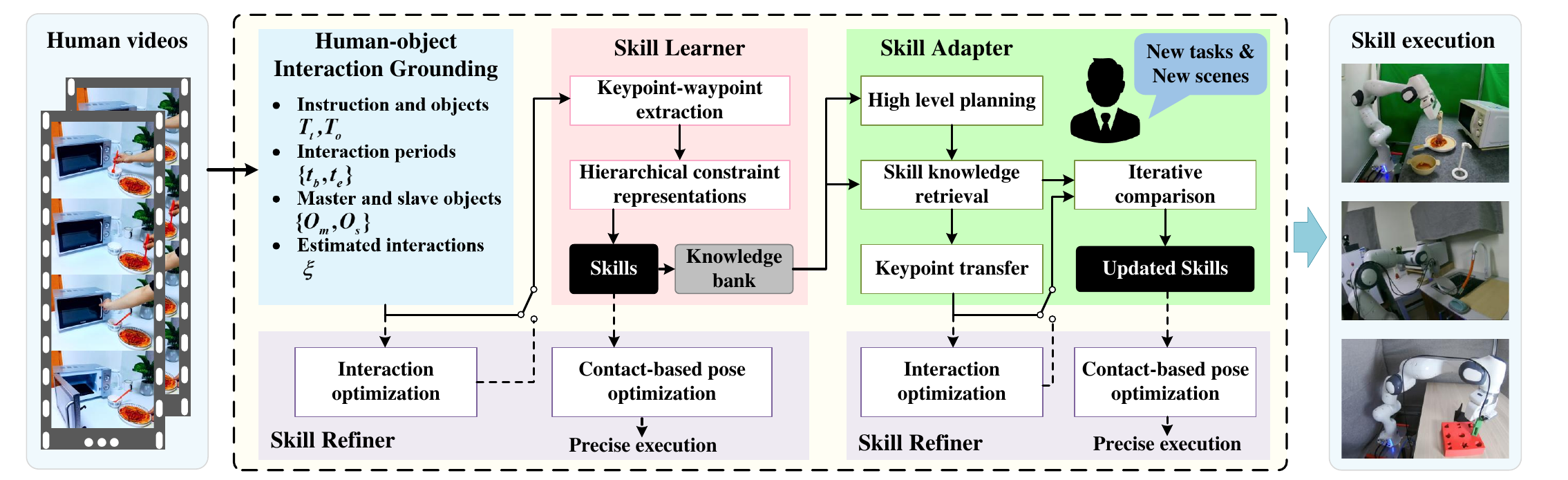}
\end{center}
\caption{\textcolor{black}{Illustration of our \nickname. The human-object interaction grounding module captures interaction motions. These interactions are utilized to derive fine-grained action skills through the skill learner. The skill adapter updates the learned skills to facilitate adaptation to novel scenes. For high-precision tasks with stringent constraints, the skill refiner optimizes interactions and pose estimation results for enhanced precision.}  }
\label{Overview_line}
\end{figure*}

\subsection{Visual Imitation Learning with  VLMs}
{Inspired by the remarkable achievements of Vision Language Models (VLMs) across diverse domains, an emerging body of research  \citep{chen2023human, wake2023gpt,patel2023pretrained, weng2024longvlm, li2023videochat, wang2024Demo2code} investigate their potential in VIL}. \textcolor{\mycolor}{GPT-4V for Robotics \citep{wake2023gpt} analyzes videos of humans performing tasks and synthesizes robot programs that leverage affordance insights}. \textcolor{black}{Digknow \citep{chen2023human} distills generalizable knowledge with a hierarchical structure, facilitating the effective generalization to novel scenes. }
\textcolor{\mycolor}{Demo2code \citep{wang2024Demo2code} generates robot task code from demonstrations via an enhanced chain-of-thought and defines a common latent specification to bridge human demonstrations and robot execution.} 
{VLaMP \citep{patel2023pretrained} predicts planning from videos through action segmentation and forecasting, effectively managing extended video sequences and intricate action dependencies.}
{While these approaches represent significant progress, they predominantly operate at a high level of abstraction, relying on predefined movement primitives or pre-trained skills for low-level execution, thereby only partially solving the control stack}. \textcolor{black}{In contrast, our investigation endeavors to push these boundaries and learn all lower-level actions for the robot, eliminating the reliance on predefined primitives and consequently broadening the applicability.}

\subsection{LLM/ VLMs for robotics} \label{LLM_works}
{The remarkable advancement of Large Language Models (LLMs) and Vision Language Models (VLMs) in recent years \citep{devlin2019bert,black2022gptneox20b,llama} has catalyzed significant innovations in robot learning. A growing body of research
\citep{wake2023chatgpt,huang2022language,xu2023creative}
have explored the application in the robot learning region}, enhancing robots' high-level environmental awareness and task comprehension abilities. Such models exhibit emergent conceptual understanding, open-ended knowledge, and vision-language reasoning abilities 
\citep{li2023blip, achiam2023gpt, kojima2022large, zhang2022opt}.
Additionally, their capacity to generate code offers a compact, transferable, generalizable strategy for representing skills \citep{chen2024roboscript, mu2024robocodex, chen2021evaluating}. Code as Policies \citep{liang2023code} and ChatGPT for Robotics \citep{vemprala2024chatgpt}, have predominantly employed LLMs to address the high-level planning aspect of robotic control. VoxPoser \citep{huang2023voxposer}, Language to Rewards \citep{yu2023language}, and MOKA \citep{liu2024moka} have explored the use of LLMs/ VLMs to generate desired regions for robot movement, significantly contributing to trajectory planning. 
{KALM \citep{fang2024keypoint} harnesses the capabilities of large pre-trained vision-language models to generate task-relevant and cross-instance consistent keypoints. ReKep \citep{huang2024rekep} presents a novel framework of Relational Keypoint Constraints, establishing a visually-grounded representational paradigm for robotic manipulation constraints.}
Nonetheless, these methods primarily concentrate on text-based task planning, facing challenges related to either insufficient provided information or complex input text content requirements, which pose obstacles for end-users to instruct robots effectively.



\section{\nickname}
In this section, we will elaborate on the technical details of \nickname, {beginning with an introduction to its framework for skill learning and generalization. We then elucidate the process of employing the acquired skills to accomplish high-precision tasks.}

\subsection{Overall Architecture}
{Considering video demonstrations $\bm{\mathcal{V}}$ of a human performing manipulation tasks, captured through an RGB-D camera.} The overall pipeline of \nickname is illustrated in Figure \ref{Overview_line}. {\nickname first ground human videos, segmenting them into subtask intervals $\{\bm{\tau}_i\}_{i=1}^{V} $ and capturing human-object interactions $\bm{I}$ (Sec. \nameref{Sec::Grounding})}. \textcolor{\mycolor}{The skill learner then distills these interactions into keypoints $\bm{\mathcal{F}}$ and waypoints $\bm{\chi}$}, and infers knowledge from distilled interactions with hierarchical representations, deriving skills with fine-grained actions (Sec. \nameref{learner}). Furthermore, {the skill adapter facilitates the transfer of extracted keypoints to novel environments}, and employs an iterative comparison strategy to revise and update the learned skills based on observations and task instructions (Sec. \nameref{Sec::Adapter}).
\textcolor{black}{For high-precision tasks with stringent constraints,
\textcolor{\mycolor}{the skill refiner optimizes the interactions and the pose estimation results, improving the ability of \nickname in even high-precision tasks} (Sec. \nameref{Sec::High}).}  
{The overall process is intuitively displayed on \href{https://fmimic-page.github.io/}{\textcolor{blue}{our homepage}}.}

\subsection{Human-object Interaction Grounding} \label{Sec::Grounding}
\textcolor{\mycolor}{While VLMs demonstrate proficiency across diverse vision tasks, they continue to struggle with fine-grained action recognition within videos}. To mitigate this limitation, {a four-stage process, illustrated in Figure \ref{Grounding}, is utilized to identify discrete subtask segments and extract corresponding human-object movements critical for skill learning}. 
\textcolor{\mycolor}{This approach transforms the complex action recognition problem into more tractable pattern reasoning problems}, leveraging the strengths of existing VLMs.

\subsubsection{Task recognition} 
\begin{figure*}[t]
\setlength{\abovecaptionskip}{+0.07cm}
\centering
\includegraphics[width=1.0\textwidth]{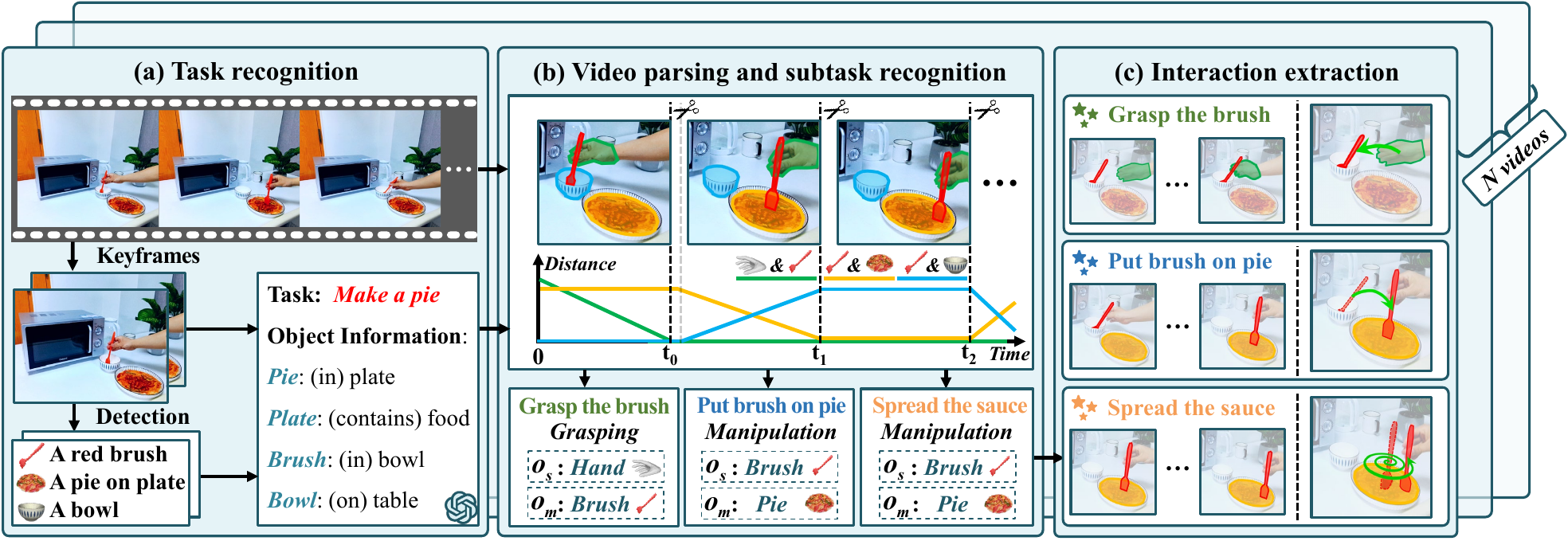}
\caption{{ Illustration of Human-object interaction grounding module.} (a) It recognizes tasks and related objects from human videos, (b) parses videos into multiple segments and identifies discrete subtasks, then (c) captures interactions within each segment. }
\vspace{-8pt}
\label{Grounding}
\end{figure*}

As shown in Figure \ref{Grounding}(a), keyframes $\bm{\mathcal{K}}$ are intermittently extracted from videos $\bm{\mathcal{V}}$, and vision foundation models ${\rm VFM}$ \citep{zhang2023recognize, wang2023caption, pan2023tokenize} are employed to detect objects within these frames. {Through the integration of extracted keyframes $\bm{\mathcal{K}}$ and their corresponding textual detection results $\bm{T}_{d}$}, { VLMs are instructed to transcribe videos into task instructions $\bm{T}_t$ (e.g., "make a pie"). \textcolor{\mycolor}{Concurrently, these models extract task-related objects $\bm{T}_o$ as structured textual descriptions}. \textcolor{\mycolor}{Each object entry includes its name and spatial relationships, exemplified by entries such as "pie: (in plate)"}. The task recognition procedure is formulated as:}
\begin{equation} 
\setlength{\abovedisplayskip}{2pt}
\setlength{\belowdisplayskip}{2pt}
\begin{split}
\bm{T}_{d} = {\rm VFM}(\bm{\mathcal{K}}), \quad \bm{T}_t, \bm{T}_o = {\rm VLM}(\bm{T}_{d}, \bm{\mathcal{K}}).
\end{split}
\end{equation}



\begin{figure}[t!]
\setlength{\abovecaptionskip}{-0.13cm}
\begin{center}
\includegraphics[width=0.30\textwidth]{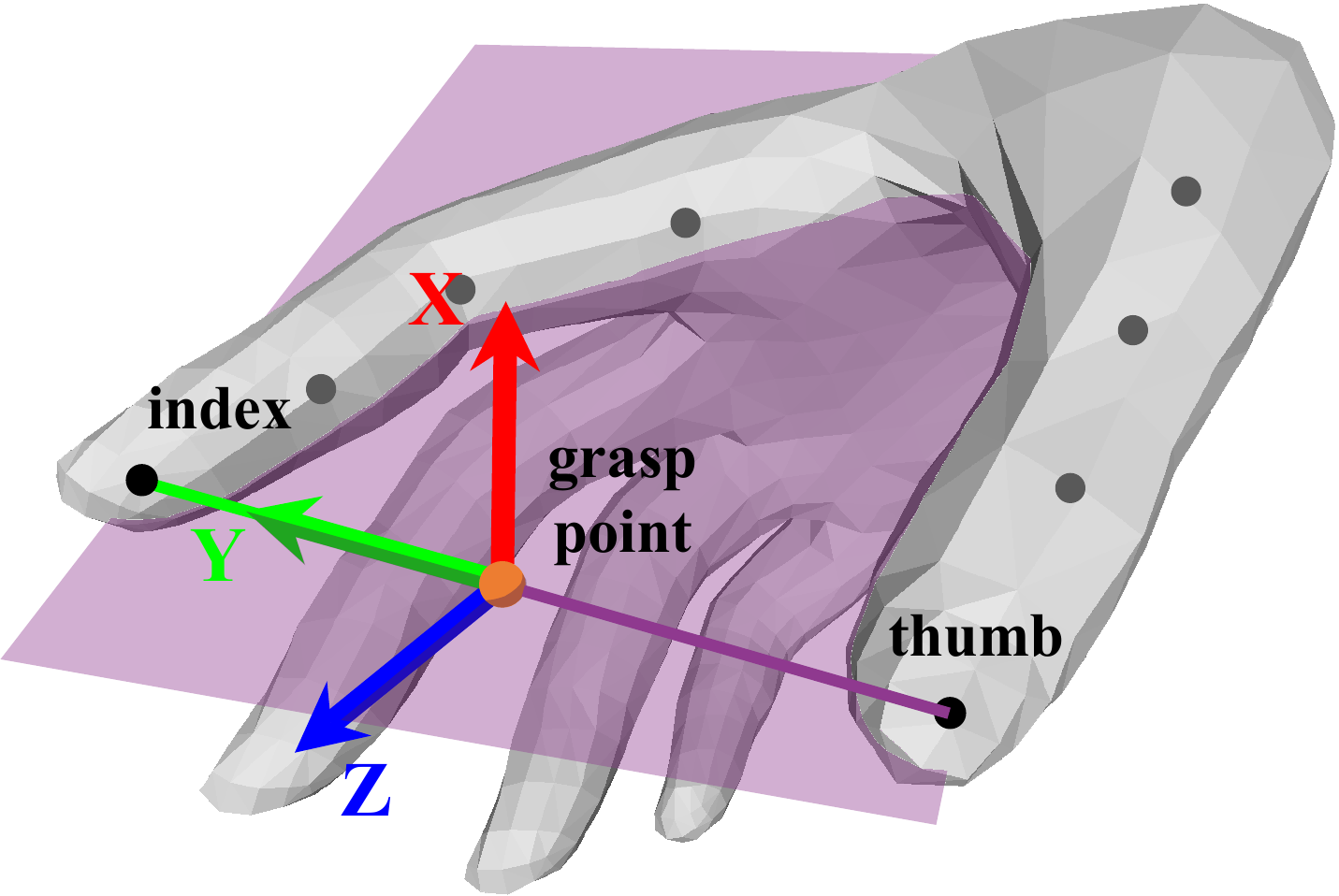}
\end{center}
\caption{Calculation of the 6D gripper pose from the estimated hand pose.  }
\label{handpose}
\vspace{-8pt}
\end{figure}

\subsubsection{Video parsing} 
{We segment videos into individual clips $\{\bm{\tau}_i\}_{i=1}^{V} $, where each clip encapsulates a distinct subtask. This segmentation process is predicated on the identification and utilization of interaction markers, which denote the onset of contact and the termination of contact.}
Specifically, SAM-Track \citep{cheng2023segment} predicts hand and task-related object masks for each frame, \textcolor{black}{utilizing object descriptions $\bm{T}_o$ as textual prompts for object mask generation, }
then the corresponding point clouds $\bm{\mathcal{P}}$ are generated through back-projection. \textcolor{\mycolor}{We identify markers by determining the interaction start time $\bm{t}_b$ and end time $\bm{t}_e$, effectively partitioning videos $\bm{\mathcal{V}}$ into multiple segments.} \textcolor{black}{Segments with hand motion trajectory lengths below than $\gamma$ are filtered out}, yielding final set of segments $\{\bm{\tau}_i\}_{i=1}^{V} $. 
Concretely, the interaction markers are obtained as follows:
\begin{equation} 
\setlength{\abovedisplayskip}{2pt}
\setlength{\belowdisplayskip}{2pt}
\begin{split}
\bm{d} &= {\rm dist}(\bm{\mathcal{P}}),  \quad \bm{t}_b = \{t|\bm{d}^{t-1}\! >\! \epsilon \wedge \bm{d}^{t} < \epsilon\}, \\ \bm{t}_e &= \{t|\bm{d}^{t-1}\! <\! \epsilon \wedge \bm{d}^{t} > \epsilon\},
\end{split}
\end{equation}
\textcolor{\mycolor}{where function ${\rm dist}$ calculates the distance between any two point clouds. $\bm{t}_b$ and $\bm{t}_e$ denote contact initiation and termination, respectively.}
As shown in Figure \ref{Grounding} (b), the hand initiates movement to grasp the brush, establishing hand-brush contact and demarcating the first video segment. \textcolor{\mycolor}{The brush then moves to the plate, creating brush-plate contact that marks the second segment. Notably, during the second subtask, the initial brush-bowl separation is filtered out based on hand trajectory length criteria, thereby maintaining the integrity of extracted segments. }

\subsubsection{Subtask recognition} \textcolor{\mycolor}{We instruct VLMs to analyze each segment $\bm{\tau}_i$, generating a subtask textual description $\bm{T}_{\tau_i}$, and categorizing the segment into grasping or manipulation phases based on the interacting entities and $\bm{T}_{\tau_i}$.} VLMs also identify master objects $\bm{O}_{m}$ and slave objects $\bm{O}_{s}$. \textcolor{\mycolor}{In the grasping phase, the agent executes a reach-and-grasp maneuver targeting $\bm{O}_{m}$, with the hand serving $\bm{O}_{s}$}. In the manipulation phase, {the agent performs a motion and makes contact with the master object $\bm{O}_{m}$ employing $\bm{O}_{s}$}.  {As illustrated in Figure \ref{Grounding} (b), the first video segment represents "grasping the brush," with the brush functioning as the master object $\bm{O}_{m}$ and the hand as the slave object $\bm{O}_{s}$. The second segment depicts "put brush on pie," wherein the brush transitions to a slave object role $\bm{O}_{s}$ while the pie serves as the master object $\bm{O}_{m}$.}



\subsubsection{Interaction estimation} \label{Sec::Motion}
{
Interactions $\bm{I}$ are characterized by trajectories $\bm{\bm{\xi}}_{O_s}$ and $\bm{\bm{\xi}}_{O_m}$ associated with slave objects $\bm{O}_s$ and master objects $\bm{O}_m$, respectively. 
The FrankMocap \citep{rong2020frankmocap} is employed to infer hand pose and shape parameters, facilitating the generation of a hand mesh model}. The Iterative Closest Point (ICP) \citep{besl1992method, rusinkiewicz2001efficient} is further implemented to align the hand mesh with the segmented hand point cloud, yielding precise hand pose trajectories $\bm{\bm{\xi}}_H =\{\bm{x}^{H^0}_{C},...,\bm{x}^{H^T}_{C}\}$ in the camera frame. These trajectories are then transformed into robot gripper pose trajectories $\bm{\bm{\xi}}_G =\{\bm{x}^{G^0}_{C},...,\bm{x}^{G^T}_{C}\}$, as shown in Figure \ref{handpose}. {The grasp location is determined as the 3D midpoint between the tips of the index finger and thumb}. The $X$-axis is established perpendicular to the plane defined by the spatial coordinates of all tracked points on the thumb and index finger. The $Y$-axis is oriented towards the index finger tip from the calculated grasp point. {The $Z$-axis is then derived through the cross product $\vec{z}=\vec{x} \times \vec{y}$.
Simultaneously, \textcolor{\mycolor}{we utilize FoundationPose \citep{wen2023foundationpose} to generate object pose trajectories $\bm{\bm{\xi}}_O =\{\bm{x}^{O^0}_{C},...,\bm{x}^{O^T}_{C}\}$}, based on object mesh models ${\bm O}$}. {These models can be either  reconstructed \citep{wen2023bundlesdf, barad2024object, sun2024l4d, fan2024hold} from videos or retrieved from the 3D mesh library}. This approach facilitates immediate application to novel objects at test time, thereby enhancing the capacity of \nickname for learning and executing robotic tasks in open-ended environments.
{In addition, ORB-SLAM3 \citep{campos2021orb} is integrated into the system to accurately capture 6DoF camera motion at the global scale}. The high-accuracy data facilitates the alignment of the estimated trajectories $\bm{\bm{\xi}}$ to the initial camera frame thus enabling the efficient human demonstration video recording using handheld cameras.

\begin{figure*}[t]
\setlength{\abovecaptionskip}{-0.13cm}
\begin{center}
\includegraphics[width=0.99\textwidth]{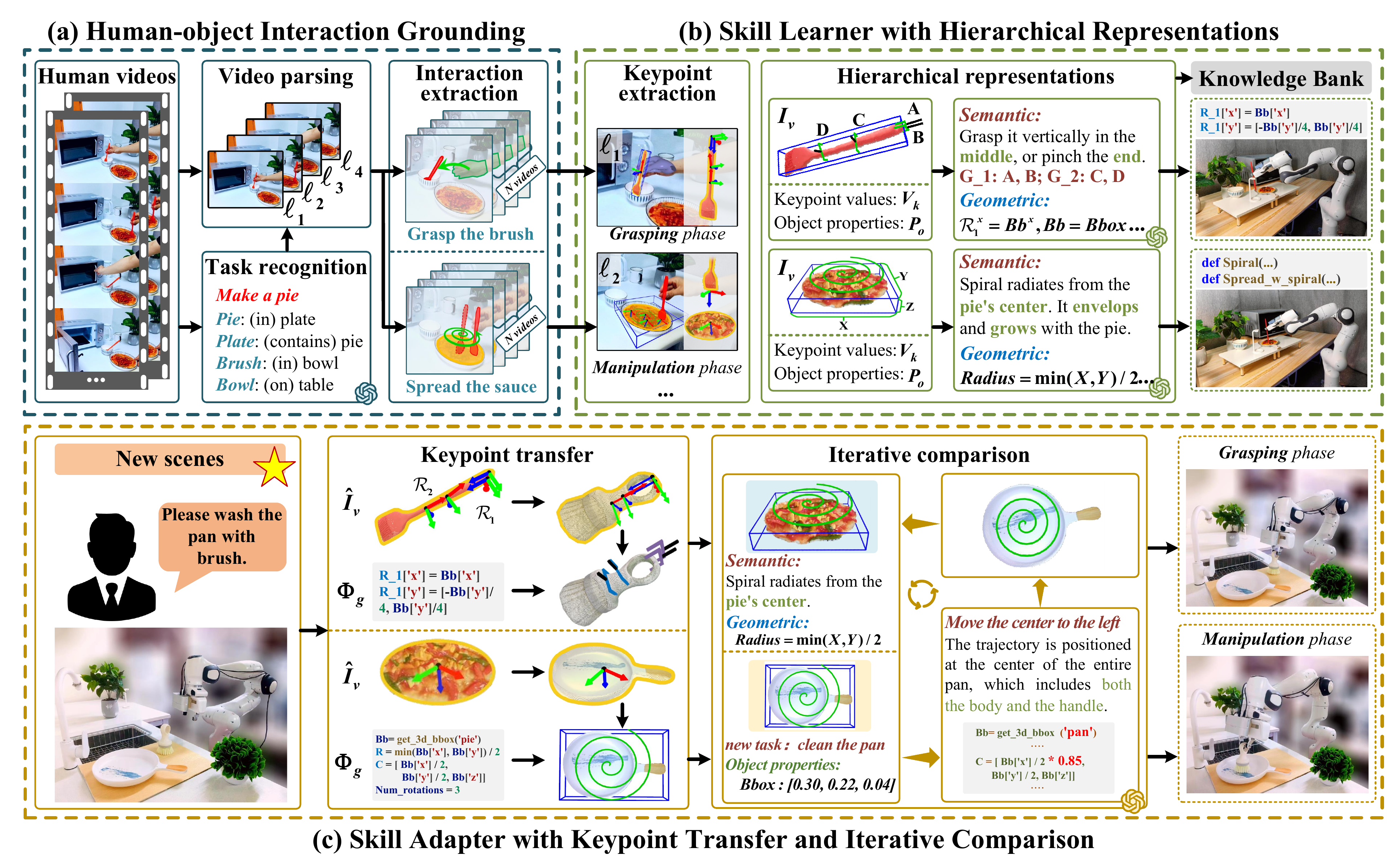}
\end{center}
\caption{{Illustration of our \nickname in skill learning and generalization. (a) The human-object interaction grounding module parses videos into multiple segments and captures interaction movements. Then, (b) the skill learner distills interactions into keypoints and waypoints, then extracts knowledge and derives skills. In novel scenes, (c) the skill adapter transfers keypoints and updates the learned skills to facilitate adaptation.}  }
\label{Overview}
\end{figure*}

\subsection{Skill Learner with Hierarchical Representations} \label{learner}

A straightforward approach for learning skills involves directly discerning the numerical trajectory patterns \citep{wang2023prompt, mirchandani2023large}. \textcolor{\mycolor}{However, VLMs struggle with reasoning about inherently redundant motion signals, limiting their capacity to extract meaningful information}. {To reduce redundancy and facilitate comprehensive comprehension, as illustrated in Figure \ref{Overview}(b), we distill the grounded interactions $\bm{I}$, and formulate them as keypoints $\bm{\mathcal{F}}$ and waypoints $\bm{\bm{\chi}}$, which compactly capture the important properties of motion signals. Subsequently, hierarchical constraint representations are proposed for analyzing these distilled interactions $\bm{I}=\{\bm{\mathcal{F}}, \bm{\chi}\}$}. \textcolor{\mycolor}{These representations express semantic constraints through visualized interactions $\bm{I}_V$ while specifying the fine-grained geometric constraints through numerical analysis of interaction values $\bm{I}$. }

\begin{figure}[t!]
\setlength{\abovecaptionskip}{-0.13cm}
\begin{center}
\includegraphics[width=0.50\textwidth]{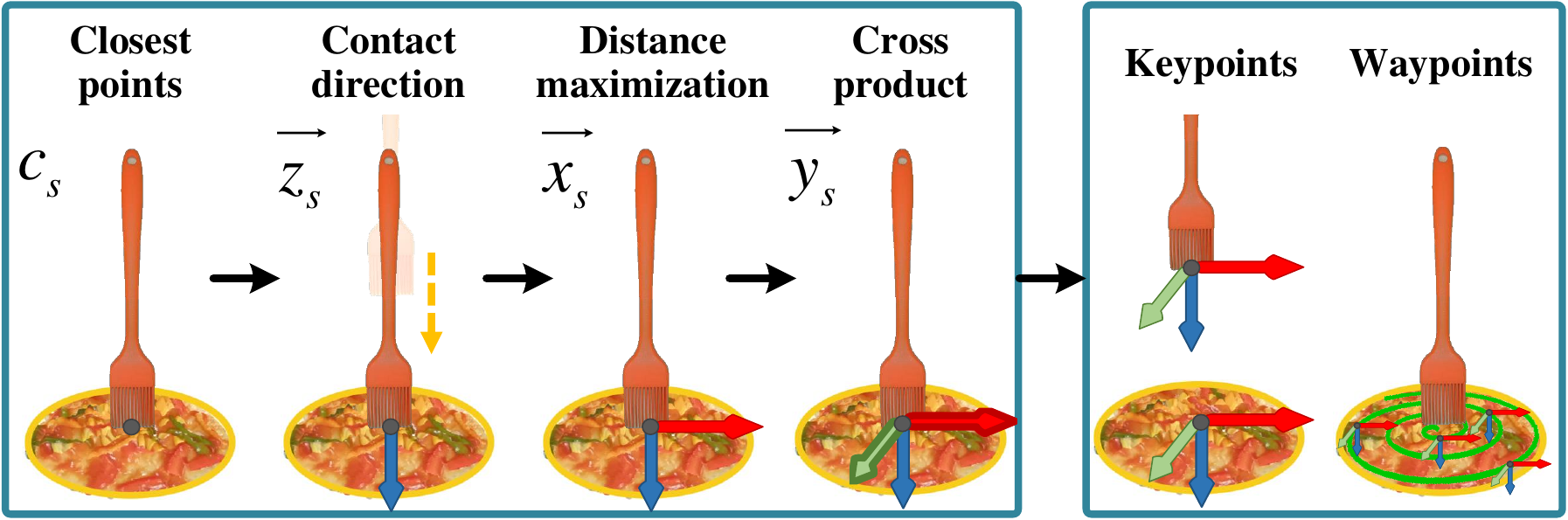}
\end{center}
\caption{{Illustration of keypoint-waypoint extraction}.  }
\label{keypoint_waypoint}
\vspace{-8pt}
\end{figure}
\subsubsection{Keypoint-waypoint extraction}
\textcolor{\mycolor}{
To reduce redundancy and foster a more comprehensive comprehension, interactions are further distilled and represented by keypoints and waypoints.
These keypoints, rigidly attached to objects, efficiently encapsulate task-relevant affordance properties}, while the waypoints delineate their relative motion trajectories.
This keypoint-waypoint paradigm facilitates efficient skill acquisition and enables \nickname to accommodate demonstrations across diverse viewpoints.

\textcolor{\mycolor}{The object point clouds in the camera frame $ \bm{O}^{t} = \bm{x}^{O^t}_{C}\bm{O} $ are derived by transforming objects $\bm{O}$ according to their poses $\bm{x}^{O^t}_{C}$,} then the position $\bm{c}$ and orientation $\{\vec{\bm{x}}, \vec{\bm{y}}, \vec{\bm{z}}\}$ of keypoints in the camera frame are determined based at the master-slave contacts $\bm{t}_b$, representing the motion affordance. 
In the grasping phase, master keypoints are specified as hand poses $\bm{x}^{G^{t_b}}_{C}$ on the master object $\bm{O}^{t_b}_m$ at hand-object contacts.
For manipulation phases, as exhibited in Figure \ref{keypoint_waypoint}, \textcolor{\mycolor}{the slave keypoint position $\bm{c}_s$ is defined as the point on the slave point clouds $ \bm{O}^{t_b}_s $ that lies nearest to the master point clouds $\bm{O}^{t_b}_m$ at master-slave contacts, which is crucial for contact maintenance or collision avoidance. }
The orientation $\vec{\bm{z}_s}$ indicates the pre-contact motion direction of the designated keypoint position. 
\textcolor{\mycolor}{To fully determine the slave object's pose, the $\vec{\bm{x}_s}$ is established perpendicularly to the $\vec{\bm{z}_s}$, extending towards the point within the object $\bm{O}_s^{t_b}$ that yields maximal distance from the origin $\bm{c}_s$:}
\begin{equation}
\begin{split}
\bm{c}_s &= {\rm argmin}_{\bm{p}\in \bm{O}_s^{t_{b}}}\|\bm{p} - \bm{O}_m^{t_{b}}\|, \\
\vec{\bm{z}_s} &= \frac{\bm{c}^{t_{b}}_s - \bm{c}^{(t_{b}-\Delta t)}_s}{\|\bm{c}^{t_{b}}_s - \bm{c}^{(t_{b}-\Delta t)}_s\|}, \, \bm{c}^{(t)}_s = \bm{c}_s(\bm{x}^{O^{t_b}}_{C})^{-1}\bm{x}^{O^t}_{C}, \\
\vec{\bm{x}_s} &= \bm{x}_s-\bm{c}_s, \\ 
&\text{where} \, \bm{x}_s \!=\! {\rm argmax}_{\bm{p}\in \bm{O}_s^{t_{b}}} \{\|\bm{p} \!-\! \bm{c}_s\| \mid (\bm{p} \!-\! \bm{c}_s) \perp \vec{\bm{z}_s}\}, \\
\vec{\bm{y}_s} &= \vec{\bm{z}_s} \times \vec{\bm{x}_s}, \\
\end{split}
\end{equation}
where $\bm{c}^{(t)}_s$ denotes the position of $\bm{c}_s$ at time $t$. {The master keypoint position $\bm{c}_m$ is selected as the nearest point to the slave keypoint position $\bm{c}_s$}, maintaining orientation $\{\vec{\bm{x}_m}, \vec{\bm{y}_m}, \vec{\bm{z}_m}\}$ alignment with those $\{\vec{\bm{x}_s}, \vec{\bm{y}_s}, \vec{\bm{z}_s}\}$ of slave keypoints.
These keypoints $\{\bm{c}, \vec{\bm{x}}, \vec{\bm{y}}, \vec{\bm{z}}\}$ for both slave and master objects are transformed into the object frames, yielding object-based keypoint frames $\bm{\mathcal{F}}$.

\textcolor{\mycolor}{To effectively capture the keypoint motion in the manipulation phase, as shown in Figure \ref{keypoint_waypoint}(b)}, the keypoint trajectories $\bm{\xi}_{\mathcal{F}} =\{\bm{x}^{\mathcal{F}^0}_{C},...,\bm{x}^{\mathcal{F}^T}_{C}\}$ in the camera frame are extracted, and relative trajectories $\bm{\xi}_{\mathcal{F}}^{'} =\{\bm{x}^{\mathcal{F}_s^0}_{M},...,\bm{x}^{\mathcal{F}_s^T}_{M}\}$ are derived by transforming the slave keypoint trajectories into the master keypoint frames. {Waypoints $\bm{\chi}$ are generated through the compression of  $\bm{\xi}_{\mathcal{F}}^{'}$ using Spatial Quality Simplification Heuristic - Extended (SQUISHE) \citep{muckell2014compression}, yielding $\bm{\chi} =\{\bm{x}^{\mathcal{F}_s^0}_{M},...,\bm{x}^{\mathcal{F}_s^N}_{M}\}$}. To accommodate keypoint motion from different demonstrations, \textcolor{\mycolor}{we employ a uniform set of keypoint frames $\bm{\mathcal{F}}_{s}$ and $\bm{\mathcal{F}}_{m}$ for the manipulation phase, established from the first demonstration. }

\subsubsection{Learning with hierarchical constraint representations}
{We render interactions $\bm{I}=\{\bm{\mathcal{F}}, \bm{\chi}\}$, and textual notations $\bm{T}_n$ on objects ${\bm O}$ to derive visualized interactions $\bm{I}_{V}=\{\bm{\mathcal{F}}_V, \bm{\chi}_V\}$. This approach enhances reasoning capabilities to analyze semantic constraints $\bm{\Phi}_{s}$ by encouraging VLMs to attend to objects and their related actions}. Furthermore, interaction values $\bm{I}$ and object properties $\bm{P}_o$ (e.g., 3-D bounding boxes) are integrated to derive geometric constraints $\bm{\Phi}_{g}$. Formally, constraints are learned as follows: 
\begin{equation} 
\setlength{\abovedisplayskip}{3pt}
\setlength{\belowdisplayskip}{3pt}
\begin{split}
\bm{I}_V &= {\rm Render}([\bm{I}, \bm{T}_n],\bm{O}), \, \bm{I}=\{\bm{\mathcal{F}}, \bm{\chi}\}\\ \bm{\Phi}_{s} &= {\rm {S}_l}(\bm{I}_V), \,  \bm{\Phi}_{g} = {\rm {G}_l}(\bm{\Phi}_{s}, \bm{I}, \bm{P}_o),
\end{split}
\end{equation}
where ${\rm {S}_l}$, and ${\rm {G}_l}$ are functions to learn semantic, and geometric constraints, respectively.


(\uppercase\expandafter{\romannumeral1}) Grasping constraints. Inspired by task space regions (TSRs) \citep{berenson2011task}, the grasping constraints $\bm{\Phi}_{g}$ are approximated as a series of bounded regions $\{\bm{\mathcal{R}}_i\}_{i=1}^{N_C}$.
\textcolor{black}{Interactive grasp poses}, represented as keypoint frames $\bm{\mathcal{F}}$, are exhibited on objects, each associated with an index notation $\bm{T}_n$. \textcolor{\mycolor}{We present these visualized keypoint frames $\bm{\mathcal{F}}_V$ to VLMs}, {which leverage their inherent knowledge and visual understanding ability to summarize semantic constraints $\bm{\Phi}_{s}$ and categorize these poses}. Geometric constraints $\bm{\Phi}_{g}$, represented as bounded regions, {are established by calculating ranges of object keypoint frames $\bm{\mathcal{F}}$ within the identical groups}, and associating them with object properties $\bm{P}_{o}$.
{This approach simplifies the complex task of constraint region generation into a series of visual understanding-based multiple-choice question answering problems. }
\textcolor{\mycolor}{Moreover, representing constraints through object properties enhances the generalization capabilities across diverse objects.} 
{Taking the brush grasping in Figure \ref{Overview}(b) as an example: keypoints A and B interact with the brush tail while keypoints C and D engage the brush handle. \nickname thus semantically categorizes A and B into one class and C and D into another, establishing semantic constraints. Subsequently, \nickname calculates the bounded regions for each keypoint group, thereby deriving the corresponding geometric constraints.}




(\uppercase\expandafter{\romannumeral2}) Manipulation constraints. Waypoint trajectories $\bm{\chi}$ are delineated on the master object ${\bm{O}_m}$, incorporating keypoints $\bm{\mathcal{F}}$ and waypoints $\bm{\bm{\chi}}$ in the textual prompt. {The VLMs identify the semantic constraints $\bm{\Phi}_{s}$ based on the visualized waypoint trajectories $\bm{\chi}_V$ and the description of the subtask $\bm{T}_{\tau_i}$.} {Geometric constraints $\bm{\Phi}_{g}$ are then formulated from semantic constraints $\bm{\Phi}_{s}$, keypoints $\bm{\mathcal{F}}$ and waypoints $\bm{\bm{\chi}}$, as well as object properties $\bm{P}_o$, expressing these constraints via the trajectory code}. {The code comprises two components: (1) parameter estimation functions $\bm{f}_p$, which derive trajectory parameters from object properties; and (2) trajectory generation functions $\bm{f}_s$, which employ estimated parameters to generate a sequence of slave waypoints relative to the master keypoint frame}, \textcolor{black}{thereby promoting effective generalization across various objects and spatial configurations}.
\textcolor{\mycolor}{As demonstrated in Figure \ref{Overview}(b), \nickname identifies the spiral trajectory pattern and determines its proportional scaling with pie dimensions. \nickname then formulates mathematical constraints and generates the implementation code, obtaining the geometric constraints.}


\begin{figure*}[t!]
\setlength{\abovecaptionskip}{-0.13cm}
\begin{center}
\includegraphics[width=0.98\textwidth]{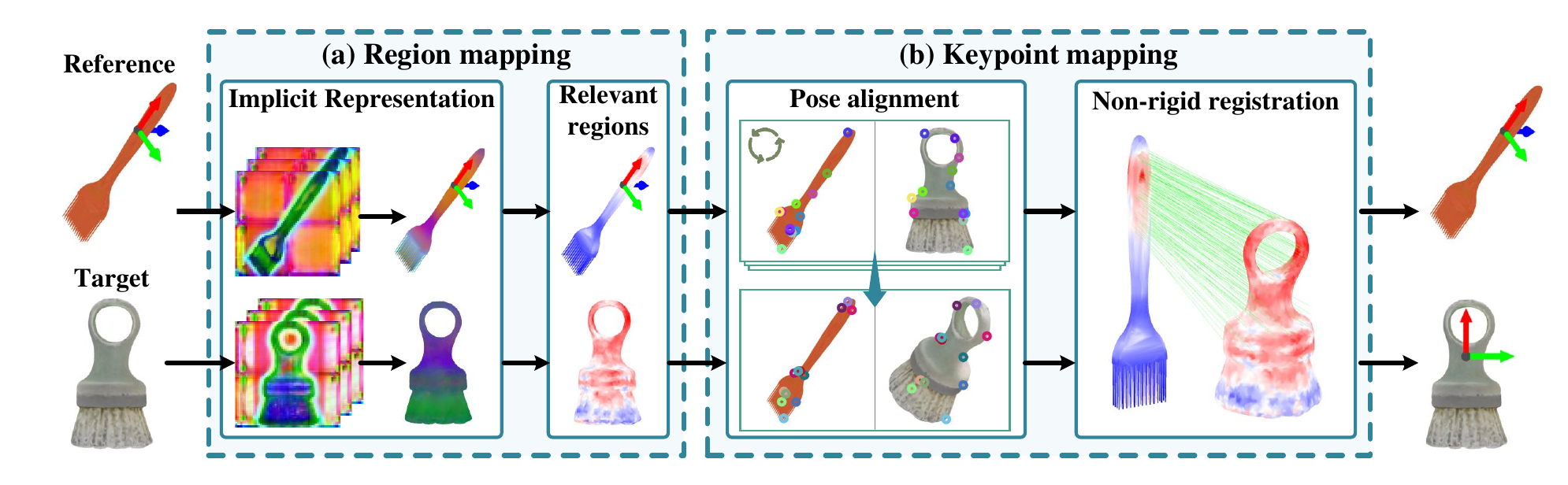}
\end{center}
\caption{\textcolor{black}{Illustration of keypoint transfer. Regional correspondences between the target and reference objects are established, and point set correspondences within these regions are identified to calculate the mapped 6D keypoints.}  }
\label{Fig::Keypoint_transfer}
\end{figure*}

\textcolor{\mycolor}{During execution, we uniformly sample grasp pose candidates within the learned grasping constraints and derive keypoint frame-based trajectory candidates from established manipulation constraints}. These trajectories are subsequently transformed into slave object pose trajectories in the world frame, {using the perceived object poses $\bm{x}^{O}_W$ \citep{vlpart2023, segrec2023} and established keypoint frames $\bm{\mathcal{F}}$}. {Consequently, end-effector trajectory candidates can be derived from the grasp candidates and the converted slave object pose trajectory candidates}. The resultant end-effector trajectory candidates are then evaluated using a motion planner, such as OMPL \citep{sucan2012open}, with the trajectories exhibiting the highest feasibility fraction being selected for implementation.

\subsubsection{Knowledge bank construction} A knowledge bank $\bm{B}$ is established to archive both high-level planning and low-level skill insights. 
{High-level planning knowledge is indexed utilizing task description $\bm{T}_t$ as keys, paired with the consequent action sequence $\bm{T}_{\tau}$ as values}. For low-level skill knowledge, the knowledge bank encompasses subtask description $\bm{T}_{\tau_i}$, interactions ${\bm{I}}=\{\bm{\mathcal{F}},\bm{\chi}\}$, reconstructed objects $\bm{O}_m$ and $\bm{O}_s$, {as well as semantic constraints $\bm{\Phi}_s$ and geometric constraints $\bm{\Phi}_g$ that represent learned skills. }
\subsection{Skill Adapter with Keypoint Transfer and Iterative Comparison} \label{Sec::Adapter}
Upon encountering the novel scenes, \nickname decomposes the instructions into multiple discrete subtasks, and then executes these specified subtasks sequentially, leveraging retrieved fine-grained action knowledge. However, the demonstration and execution scenes may differ in objects and tasks, thereby impeding direct skill transfer to unseen environments. To mitigate these challenges, as depicted in Figure \ref{Overview}(c), \textcolor{\mycolor}{we propose a region-to-keypoint mapping approach that precisely transfers keypoints to target objects, generalizing acquired affordances to previously unseen environments}. Furthermore, VLMs are instructed to adapt skills via an iterative comparison strategy, which updates learned skills by iteratively contrasting with the demonstrated knowledge. \textcolor{\mycolor}{This enables the effective adaptation of retrieved skills to targeted, unseen scenes. }

\subsubsection{High-level planning} High-level planning knowledge $\bm{T}_{\tau}$ is retrieved from knowledge bank $\bm{B}$ based on the task instruction, {which functions as an in-context example for VLMs, along with the scene observation.  VLMs serve as a physically-grounded task planners \citep{skreta2024replan, hu2023look}, generating both a sequence of actionable steps $\bm{T}_{\tau_i}$ and detailed descriptions of task-related objects $\bm{T}_{o}$.}

\subsubsection{Skill knowledge retrieval}
{Low-level skill knowledge is effectively retrieved from knowledge bank $\bm{B}$ through a three-step retrieval pipeline:}
(\uppercase\expandafter{\romannumeral1}) \textcolor{\mycolor}{Text-based retrieval. Based on the queried subtask description, we leverage the text encoder \citep{wang2022text} to retrieve $N_{t}$ demonstrations with the highest similarity between their associated subtasks $\bm{\hat{T}}_{\tau_i}$ and the query.} (\uppercase\expandafter{\romannumeral2}) Semantic constraint filtering. Then, we instruct VLMs to analyze the semantic constraints required to complete the task and select the $N_{\Phi}$ most relevant ones from the extracted semantic constraints $\bm{\hat{\Phi}}_s$, preserving the corresponding demonstrations. (\uppercase\expandafter{\romannumeral3}) \textcolor{\mycolor}{Visual matching. Finally, the demonstrations displaying the highest degree of object similarity are identified.} The image similarities \citep{radford2021learning} are calculated for both master and slave objects between the current scenarios $\{\bm{O}_m, \bm{O}_s\}$, and demonstrations $\{\bm{\hat{O}}_m, \bm{\hat{O}}_s\}$. 
{Consequently, the retrieved constraints $\{\bm{\hat{\Phi}}_g, \bm{\hat{\Phi}}_s\}$, objects $\{\bm{\hat{O}}_m, \bm{\hat{O}}_s\}$, and interactions ${\bm{\hat{I}}} = \{\bm{\hat{\mathcal{F}}}, \bm{\hat{\bm{\chi}}}\}$ serve as the reference knowledge for the current task.}

\subsubsection{Keypoint transfer} \label{Sec::keypoint_mapping}
\textcolor{\mycolor}{We present a novel region-to-keypoint mapping framework for precise and zero-shot 6D keypoint transfer from retrieved knowledge to unseen environments}. {As illustrated in Figure \ref{Fig::Keypoint_transfer}, regional correspondences between the reference and target objects are initially established using semantic features extracted from pre-trained models \citep{radford2021learning, oquab2023dinov2, caron2021emerging, rombach2022high}}. Subsequently, point set correspondences within these regions are identified, which are then utilized to calculate the mapped 6D keypoints.  


(\uppercase\expandafter{\romannumeral1}) {Region mapping}. 
Figure \ref{Fig::Keypoint_transfer}(a) illustrates the region mapping procedure. We initially construct the implicit descriptor fields representation \citep{wang2023d, tsagkas2024click} for reference object $\bm{\hat{O}}$ and target object $\bm{O}$. \textcolor{\mycolor}{These representations integrate RGB-D observations and their corresponding feature maps into single, multi-view consistent, and differentiable 3D representations $\phi(\bm{\hat{O}})$ and $\phi(\bm{O})$.}
In addition to the signed distance to the surface $\bm{s}$, the representation incorporates additional features  $\bm{f}^{\text{SD}} \in \mathbb{R}^{D_{\text{SD}}}$ extracted from stable diffusion models (SD) \citep{rombach2022high}. {Specifically, the representation for the target object is obtained as follows: }
\begin{equation}
\setlength{\abovedisplayskip}{3pt}
\setlength{\belowdisplayskip}{3pt}
\bm{s} = \phi_{s}(\bm{O}), \bm{f}^{\text{SD}} = \phi_{f}^{\text{SD}}(\bm{O}).
\end{equation}
{Then, we voxelize these representations and obtain voxel feature grids $ \bm{\hat{\mathcal{V}}}^{\text{SD}}$ and $ \bm{\mathcal{V}}^{\text{SD}}$, respectively.} 
We separately identify the keypoint-relevant areas in the reference and target objects by
computing the cosine similarity between the voxel features \{$\bm{\hat{\mathcal{V}}}^{\text{SD}}$, $\bm{\mathcal{V}}^{\text{SD}}$\} and the features $\bm{\hat{\mathcal{V}}}^{\text{SD}}_{\hat{\mathcal{F}}}$ located in the reference keypoint frame $\bm{\hat{\mathcal{F}}}$. {The similarity $SIM_{\mathcal{V}}$ between the $\bm{\mathcal{V}}^{\text{SD}}$ and  $\bm{\hat{\mathcal{V}}}^{\text{SD}}_{\hat{\mathcal{F}}}$ is computed as:}
\begin{equation}
\setlength{\abovedisplayskip}{3pt}
\setlength{\belowdisplayskip}{3pt}
SIM_{\mathcal{V}} = \frac{\bm{\mathcal{V}}^{\text{SD}} \cdot \bm{\hat{\mathcal{V}}}^{\text{SD}}_{\hat{\mathcal{F}}}}{\| \bm{\mathcal{V}}^{\text{SD}}\| \cdot \| \bm{\hat{\mathcal{V}}}^{\text{SD}}_{\hat{\mathcal{F}}}\|}.
\end{equation}
\textcolor{\mycolor}{Subsequently, the OTSU \citep{otsu1975threshold} thresholding method is applied to adaptively identify the relevant regions for both reference and target objects, according to similarity $SIM_{\hat{\mathcal{V}}}$ and $SIM_{\mathcal{V}}$, respectively.}

(\uppercase\expandafter{\romannumeral2}) {Keypoint mapping}. 
{We transform the relevant regions of both reference and target objects into the point sets $\bm{\hat{\mathcal{P}}}=\{\hat{{p}}_i\in \mathbb{R}^3\}_{i=1}^{N_{\hat{\mathcal{P}}}}$ and $\bm{\mathcal{P}}=\{{p}_j\in \mathbb{R}^3\}_{j=1}^{N_{\mathcal{P}}}$, respectively.
The keypoint frames $\bm{\hat{\mathcal{F}}}$ and $\bm{{\mathcal{F}}}$ are associated with their corresponding point sets $\bm{\hat{\mathcal{P}}}$ and $\bm{{\mathcal{P}}}$, parameterizing $\bm{\hat{\mathcal{F}}}$ and $\bm{{\mathcal{F}}}$ as $\{\bm{\hat{\mathcal{F}}}(\hat{p}_i)\}_{i=1}^{N_{\hat{\mathcal{P}}}}$ and $\{\bm{{\mathcal{F}}}({p}_j)\}_{j=1}^{N_{{\mathcal{P}}}}$, respectively}, where $\bm{\hat{\mathcal{F}}}(\hat{p}_i)$ denotes position of point $\hat{p}_i$ in keypoint frame $\bm{\hat{\mathcal{F}}}$. 
\textcolor{black}{Therefore,
keypoint frame alignment from reference objects to target objects is formulated as a problem of point set correspondence establishment within relevant region pairs and can be solved through non-rigid point cloud registration. }

{However, the discrepancies in spatial orientation and positioning between reference and target objects may introduce significant challenges in non-rigid registration processes}. To address this, we initially perform a pose alignment procedure using salient point correspondences extracted from the semantic feature maps. 
{Specifically, we leverage the established feature grids \{$\bm{\hat{\mathcal{V}}}^{\text{SD}}$, $\bm{\mathcal{V}}^{\text{SD}}$\} to identify correspondences through the implementation of the best-buddies nearest neighbor matching algorithm \citep{dekel2015best}}, which establishes correspondences between point pairs that exhibit mutual nearest neighbor relationships across the two feature point sets. {Since this stringent mutual nearest-neighbor criterion may result in a paucity of established correspondences and lead to imprecise transformation solutions}, \textcolor{\mycolor}{we relax this strict nearest-neighbor requirement by allowing matches to be considered valid when they are mutually located within a specified feature space neighborhood.}
Formally, the relaxed nearest-neighbor criteria are defined as follows:
\begin{equation}
bb_{r} = \left\{\begin{array}{ll}
1 & \mathrm{d}\left(\mathrm{NN}\left(\bm{{\mathcal{V}}}^{\text{SD}}, \bm{\hat{\mathcal{V}}}^{\text{SD}}_{\hat{C}_{i}}\right), \bm{{\mathcal{V}}}^{\text{SD}}_{{C}_{i}}\right) \le d_t  \\ & \wedge\mathrm{d}\left(\mathrm{NN}\left(\bm{{\mathcal{V}}}^{\text{SD}}_{{C}_{i}}, \bm{\hat{\mathcal{V}}}^{\text{SD}}\right), \bm{\hat{\mathcal{V}}}^{\text{SD}}_{\hat{C}_{i}}\right) \le d_t \\
0 & \mathrm{otherwise},
\end{array}\right.
\end{equation}
where $\mathrm{NN}$ denotes the nearest neighbor operation, $\mathrm{d}$ represents the feature space distance, and $d_t$ denotes the neighborhood threshold.
These correspondences are processed by the RANSAC \citep{fischler1981random} algorithm to filter out mismatches. {Following this, we derive the least-squares rigid transformation $\mathcal{T}$ to achieve pose alignment between target and reference objects.}

{Subsequently, we utilize non-rigid registration \citep{zhao2024correspondence} to identify the optimal deformation map ${\varphi}$ that minimizes the shape deviation between ${\varphi}(\bm{\hat{\mathcal{P}}})\triangleq\bm{\hat{\mathcal{P}}}+\mathbf{\nu}(\bm{\hat{\mathcal{P}}})$ and $\mathcal{T}(\bm{\mathcal{P}})$}, where $\mathbf{\nu}$ represents the displacement filed acting on each reference point ${\hat{p}}_i$. Consequently, the keypoint frames $\bm{\mathcal{F}}'$ of transformed target objects $\mathcal{T}(\bm{O})$ are obtained by minimizing the mean squared displacement of the observed coordinates. 
\begin{equation}
\setlength{\abovedisplayskip}{3pt}
\setlength{\belowdisplayskip}{3pt}
\begin{split}
\bm{\mathcal{F}}' = {\rm argmin}_{\mathcal{F}'} \sum_{i=1}^{N_{\hat{\mathcal{P}}}} \|\bm{\mathcal{F}}'({\varphi}({\hat{p}}_i)) - \bm{\hat{\mathcal{F}}}({\hat{p}}_i) \|^{2}.
\end{split}
\end{equation}
{Finally, the keypoint frames $\bm{\mathcal{F}}$ of target objects $\bm{O}$ are obtained by applying the inverse transformation $\mathcal{T}^{-1}$ to $\bm{\mathcal{F}}'$, therefore facilitating the generalization of learned keypoint frames to the unseen objects.} 

\subsubsection{Iterative comparison} 
\textcolor{\mycolor}{Skill adapting is further advanced through an iterative comparison process designed to continually refine skills. }
At each iteration, the updated interactions $\bm{I}=\{\bm{\mathcal{F}}, \bm{\chi}\}$ are derived by applying the geometric constraints $\bm{{\Phi}}_{g}$, predicated on the perception results, e.g., object properties $\bm{P}_o$ and estimated object poses $\bm{x}^{O}_{W}$ in the world frame. {Then, interactions $\bm{I}$, along with textual notations $\bm{T}_n$ and objects ${\bm O}$ are rendered to synthesize visualized interactions $\bm{I}_{V}=\{\bm{\mathcal{F}}_V, \bm{\chi}_V\}$.} VLMs perform a comparative analysis between the adapted interactions $\bm{I}=\{\bm{\mathcal{F}, \bm{\chi}}\}$ and retrieved interactions $\bm{\hat{I}}=\{\bm{\hat{\mathcal{F}}, \bm{\hat{\chi}}}\}$, \textcolor{black}{subsequently modifying skill constraints $\bm{{\Phi}}_{s}$ and $\bm{{\Phi}}_{g}$}. \textcolor{\mycolor}{This iterative process continues until either convergence or the maximum number of iterations $N_I$ is reached.} This approach facilitates reasoning in  VLMs by directing their attention to discrepancies, thus enabling them to pinpoint the best available solution through an iterative process. The adaptation procedure at the $i$-th iteration can be formally represented as:
\begin{equation} 
\setlength{\abovedisplayskip}{2pt}
\setlength{\belowdisplayskip}{2pt}
\begin{split}
\bm{{I}}^{i} &= \bm{{\Phi}}_{g}^{i}(\bm{P}_o, \bm{x}^{O}_{W}), \\
\bm{{I}}_V^{i} &= {\rm Render}([\bm{{I}}^{i}, \bm{T}_n], \bm{O}), \\
\bm{{\Phi}}_{s}^{i\!+\!1} &= {\rm {S}_a}(\bm{\hat{I}}_V, \bm{{I}}_V^{i},\bm{{\hat{\Phi}}}_{s}, \bm{{\Phi}}_{s}^{i}), \\  \bm{{\Phi}}_{g}^{i\!+\!1} &= {\rm {G}_a}( \bm{{\hat{\Phi}}}_{g}, \bm{{\Phi}}_{g}^{i}, \bm{{\Phi}}_{s}^{i\!+\!1}, \bm{I}^{i}, \bm{P}_o),
\end{split}
\end{equation}
where $\bm{{\hat{\Phi}}}_{g}$ and $\bm{{\hat{\Phi}}}_{s}$ denote referential constraints, extracted from the knowledge base. The functions ${\rm {S}_a}$ and ${\rm {G}_a}$ adapt semantic and geometric constraints, respectively.

As the grasping constraints are directly determined by the keypoint frames, {the iterative comparison adaptation process predominantly focuses on adjusting the manipulation constraints.}
Specifically, \textcolor{black}{VLMs are instructed to iteratively self-summarize and update manipulation constraints based on the task instruction and scene variations}. \nickname generates waypoints $\bm{\chi}$ adhering to geometric constraints $\bm{\Phi}_g$,
which are exhibited on master objects based on master keypoint frames $\bm{\mathcal{F}}_m$. {The VLMs analyze discrepancies  between the adapted interactions $\bm{\chi}_V$ and the referential interactions $\bm{\hat{\chi}}_V$ to revise semantic constraints $\bm{\Phi}_s$. Subsequently, the geometric constraints $\bm{\Phi}_g$ are modified according to the updated semantic constraints $\bm{\Phi}_s$}, along with generated waypoint values $\bm{\chi}$ and perceived object properties $\bm{P}_o$.
\textcolor{\mycolor}{As illustrated in Figure \ref{Overview}(c), \nickname endeavors to transfer the acquired 'spread the sauce' skill to accomplish the 'brush the pan' task. However, a direct application proves problematic since the original skill maximizes coverage of the entire bounding box, whereas the pan's handle presents a structural impediment that would result in task failure.} Consequently, \nickname implements modifications to the skill parameters, specifically offsetting the spiral's central coordinates and reducing its radius, thereby enabling successful skill transfer and task execution.

\begin{figure}[t!]
\setlength{\abovecaptionskip}{-0.13cm}
\begin{center}
\includegraphics[width=0.48\textwidth]{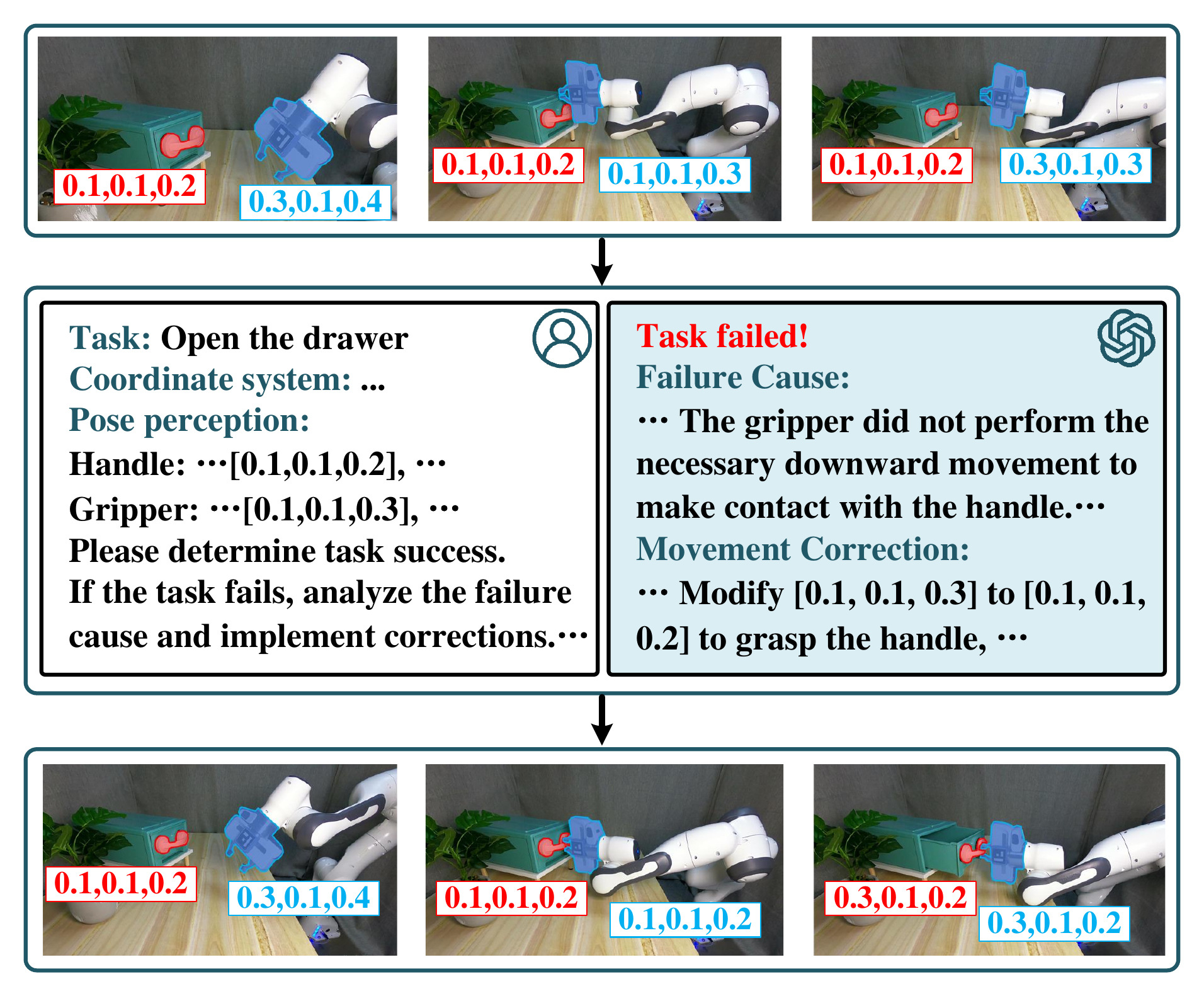}
\end{center}
\caption{{Illustration of the process of fine-grained action correction. Our \nickname exhibits autonomous failure identification and reasoning for rectification. }  }
\label{FD}
\end{figure}

\subsubsection{Fine-grained action correction} {Despite the capacity  of  VLMs to generate effective constraints, environmental noise, such as trajectory estimation errors, impedes successful task execution}. Thus, we leverage  VLMs to detect and address failures during execution.
\textcolor{\mycolor}{As illustrated in Figure \ref{FD}, the status of task completion is evaluated through real-time object pose data analysis.} In the event of the detected failure, \nickname transmits both object and end-effector pose information to VLMs for subsequent inference and generation of fine-grained corrective actions. Object tracking is implemented via FoundationPose \citep{wen2023foundationpose}, \textcolor{\mycolor}{while end-effector pose estimation is achieved through AprilTag tracking, followed by transformation to the end-effector frame}. The AprilTag is rigidly attached to the gripper, enabling precise localization. 

\begin{figure*}[t!]
\setlength{\abovecaptionskip}{-0.13cm}
\begin{center}
\includegraphics[width=1.0\textwidth]{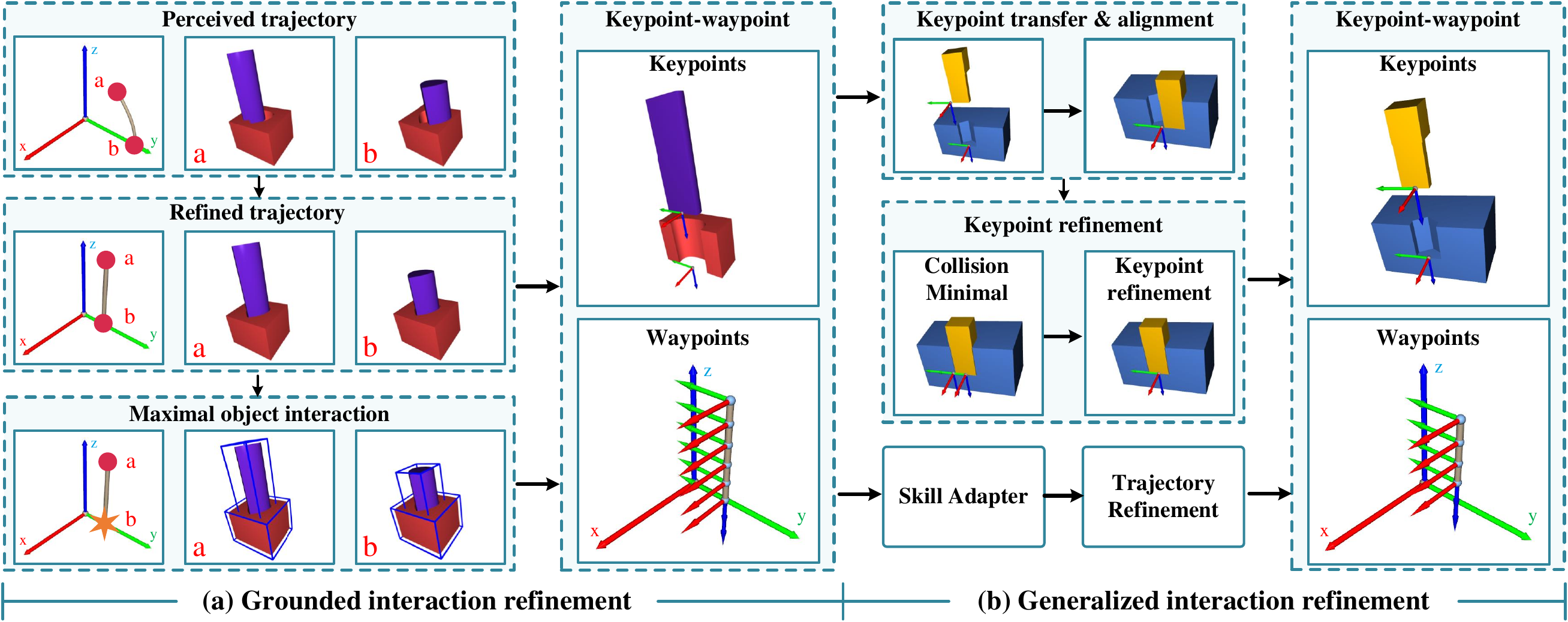}
\end{center}
\caption{{Illustration of collision-minimal interaction optimization. (a) FMimic extracts the perceived trajectory via human-object interaction grounding and subsequently optimizes it for collision minimization. Interaction points are determined by identifying minimal bounding box unions. Keypoints and waypoints are extracted based on the refined trajectory and interaction points.  (b) For novel scenes, we first transfer and align keypoints, then adjust object positions to minimize collisions. Keypoints are subsequently realigned, yielding the final transferred keypoints and waypoints.}  }
\label{Int_opt}
\end{figure*}

\subsection{Skill Refiner with Contact-aided Optimization}\label{Sec::High}
\textcolor{black}{High-precision manipulation tasks characterized by stringent constraints persist as formidable challenges in the domain of autonomous robotics, such as assembly tasks. \textcolor{\mycolor}{These tasks are typically considered beyond the reach of visual imitation learning, primarily due to their stringent precision requirements}. To resolve this and showcase the effectiveness of \nickname in such high-precision tasks, we propose a skill refiner with contact-aided optimization. Concretely, the grounded and transferred interactions are refined based on the collision-minimal condition, \textcolor{\mycolor}{facilitating the acquisition and generalization of skills applicable to highly constrained tasks}. During execution, the perceived relative poses are optimized through iterative master-slave contact, achieving effective pose optimization and precise skill execution. Consequently, {our \nickname effectively learns highly constrained manipulation tasks from a limited corpus of human videos}. Furthermore, it exhibits robust task completion capabilities even in novel environmental contexts.
}

\begin{figure*}[t!]
\setlength{\abovecaptionskip}{-0.13cm}
\begin{center}
\includegraphics[width=1.0\textwidth]{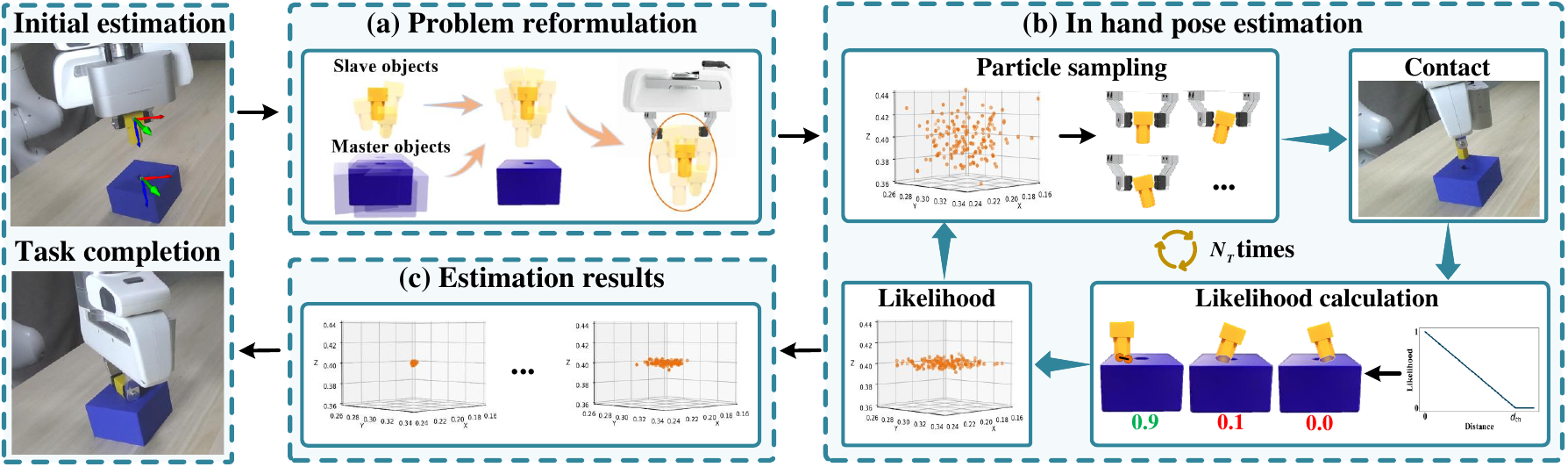}
\end{center}
\caption{\textcolor{black}{Illustration of contact-based pose optimization. (a) The pose estimation uncertainty associated with both objects is integrated and approximated through the in-hand pose estimation procedure. (b) This process iteratively refines the initial pose estimates via master-slave contact interactions. (c) The refinement significantly reduces uncertainty, enabling the robust execution of high-precision manipulation tasks.}}
\label{Reform}
\end{figure*}

\subsubsection{Collision-minimal interaction optimization} Leveraging the highly constrained nature of the tasks, the grounded and generalized interactions undergo refinement through collision avoidance mechanisms.

(\uppercase\expandafter{\romannumeral1}) Grounded interaction refinement. {As shown in Figure \ref{Int_opt} (a)}, the grounded trajectories $\bm{\xi}_O$, {extracted from human videos, are initially converted into the corresponding object point cloud trajectories $O^{t}$ by transforming objects $\bm{O}$ according to their poses $x^{O^{t}}_C$}. \textcolor{\mycolor}{Subsequently, a search is conducted within the 6D neighborhood of perceived slave object poses to identify the proximal collision-minimal poses, thereby facilitating the refinement and optimization of the estimated trajectories. }
Afterwards, the slave and master keypoint frames are defined at the temporal point of maximal object interaction, {which corresponds to the minimal volume encapsulating the combined bounding boxes master objects $O^{t}_m$ and slave objects $O^{t}_s$}. Finally, the slave keypoint frames are determined as the approach described in \nameref{learner}, and master keypoint frames are aligned with their slave counterparts. The waypoints are extracted based on the refined trajectories and established keypoint frames. The refined interactions $\bm{I} = \{\bm{\mathcal{F}}, \bm{\chi}\}$ enable the \nickname to learn skills with enhanced precision.

(\uppercase\expandafter{\romannumeral2})
Generalized interaction refinement. 
Upon encountering the novel scenes, \textcolor{\mycolor}{as presented in Figure \ref{Int_opt} (b), keypoint frames are transferred into previously unseen environments. Then we determine the master and slave object poses that minimize the distance between their respective keypoint frames while avoiding object collisions, and master keypoint frames are recalibrated to align with slave keypoint frames based on the identified poses}, yielding optimized keypoint frames. {Utilizing these refined keypoint frames,
the skills are adapted through the iterative comparison based on the optimized keypoint frames},
and the waypoints $\bm{\chi}$ predicted from updated skills are refined by identifying the proximal collision-minimal configurations. This approach facilitates robust and effective skill adaptation, even for tasks characterized by high degrees of constraint.


\subsubsection{Efficient contact-based pose optimization} \textcolor{\mycolor}{Pose estimation via visual recognition often exhibits limited precision due to factors such as noise, occlusions, and calibration errors.} Current methods mainly introduce contact  sensing to enhance pose estimation precision \citep{von2021precise, wirnshofer2019state, wirnshofer2020controlling, hartley2020contact}. However, contact  sensing-based pose estimation methods estimate poses sequentially, resulting in prolonged execution times. To enhance execution efficiency, we reformulate the problem and directly optimize their relative poses through a contact-based estimation procedure.  

(\uppercase\expandafter{\romannumeral1}) Problem reformulation. 
Visual recognition methods provide initial estimates of keypoint poses $\{\bm{x}^{\mathcal{F}_m}_W, \bm{x}^{\mathcal{F}_s}_W\}$ in the world frame, abbreviated as $\{\bm{x}^{M}_W, \bm{x}^{S}_W\}$. {These initial estimates are then refined using contact  sensing to approximate the ground truth poses $\{\bm{x}^{M*}_W, \bm{x}^{S*}_W\}$, which are formulated as:}
\begin{equation}
\setlength{\abovedisplayskip}{3pt}
\setlength{\belowdisplayskip}{3pt}
\bm{x}_W^{M*} = \bm{e}^{M}\bm{x}_W^{M}, \quad \bm{x}_W^{S*} = \bm{e}^{S}\bm{x}_W^{S},
\end{equation}
where $e$ denotes the pose estimation error.
Then, we define the relative pose between master and slave keypoints, and integrate these two error terms $\bm{e}^{M}$ and $\bm{e}^{S}$ into an overall error term $\bm{e}^o$. {The overall error is determined through the slave keypoint pose estimation process as follows:}
\begin{equation}
\setlength{\abovedisplayskip}{3pt}
\setlength{\belowdisplayskip}{3pt}
\begin{split}
\bm{x}_M^{S*} &= {(\bm{e}^{M}\bm{x}_W^{M})}^{-1}\bm{e}^{S}\bm{x}_W^{S} \\
&= \bm{x}_M^{W}{(\bm{e}^M)}^{-1}\bm{e}^{S}\bm{x}_W^{S} \\
&= \bm{x}_M^{W}(\bm{e}^{O}\bm{x}_W^{S}), \quad \bm{e}^{o}={(\bm{e}^M)}^{-1}\bm{e}^{S}.
\end{split} 
\end{equation}
The pose of the slave keypoint can be estimated through the in-hand pose $\bm{x}_G^{S}$ and the end-effector pose $\bm{x}_W^{G}$. {Specifically, the relative pose can be further written as:}
\begin{equation}
\setlength{\abovedisplayskip}{3pt}
\setlength{\belowdisplayskip}{3pt}
\begin{split}
\bm{x}_M^{S*} &= \bm{x}_M^{W}(\bm{x}_W^{G}\bm{e}^{O'}\bm{x}_G^{S}) \\
&= \bm{x}_M^{W}\bm{x}_W^{G}(\bm{e}^{O'}\bm{x}_G^{S}),
\end{split} 
\end{equation}
\textcolor{black}{the error term $\bm{e}^{O'}$ is considered in the in-hand pose estimation procedure to approximate the ground truth relative poses. Figure \ref{Reform} (a) presents the illustration of this reformulation. }





(\uppercase\expandafter{\romannumeral2}) {Contact-based pose optimization}.
{
The reformulation of the problem into an in-hand pose estimation paradigm enables our method to leverage the assumption of pose stability within the gripper \citep{von2021precise}. This reformulation facilitates the establishment of a compact state-space model:
}
\begin{align}
\setlength{\abovedisplayskip}{3pt}
\setlength{\belowdisplayskip}{3pt}
\text{System:}& \ \boldsymbol{z}_{n} = \boldsymbol{z}_{n-1} + \bm{\varepsilon}, \\
\text{Contact:}& \ \boldsymbol{y}_n = h(\boldsymbol{z}_{n};\bm{O}_m,\boldsymbol{O}_{s},\boldsymbol{x}_W^{G_n}) + \bm{e}^{C}, \label{touch}
\end{align}
\textcolor{\mycolor}{where $\boldsymbol{z}_{n}$ denotes the estimated values for $\bm{e}^{O'}\bm{x}_G^{S_n}$ at the step $n$, $\bm{\varepsilon}$ and $\bm{e}^{C}$ are the system error and contact estimation error, respectively.
The objective is to iteratively estimate $\boldsymbol{z}_{{N_T}}$ based on a set of measurements acquired through environmental interaction $\left\{\boldsymbol{y}_n,\boldsymbol{x}_W^{G_n}\right\}_{n=1}^{N_T}$. 
This estimation problem can be formulated within a Bayesian framework and addressed using a particle filter.} The estimation procedure is illustrated in Figure \ref{Reform} (b). Let $M$ particles $\left\{\boldsymbol{z}_{n-1}^{(j)}\right\}_{j=1}^M$ of pose estimates be sampled from the posterior, {which are predicted from the previous estimation step. At step $n$, the posterior distribution conditioned on the contact information up to step $n-1$ is expressed as follows:}
\begin{equation}
\setlength{\abovedisplayskip}{3pt}
\setlength{\belowdisplayskip}{3pt}
\begin{split}
\frac1M\sum_{j=1}^M&p\left(\boldsymbol{y}_n|\boldsymbol{z}_{n|n-1}^{(j)},\bm{O}_m, \bm{O}_s , \bm{x}_W^{M}, \bm{x}_W^{G_n}\right), \\
\boldsymbol{z}_{n|n-1}^{(j)}&=\boldsymbol{z}_{n-1}^{(j)}+\bm{\varepsilon}.
\end{split}
\end{equation}

\textcolor{\mycolor}{During the contact phase, the robot is incrementally advanced toward the designated position of the master object until contact is confirmed by force detection.} The robot end-effector pose $\boldsymbol{x}_W^{G_n}$ and master object pose $\bm{x}_W^{M_n}$ are recorded. {These pose measurements are subsequently leveraged to estimate the simulated contact results for each particle $\left\{\boldsymbol{z}_{n}^{(j)}\right\}_{j=1}^M$ through the transform $h$ (Eq. \ref{touch})}. The transform $h$ computes a geographical distance $d_n^{(j)}$ to the master object, where $d_n^{(j)}=d(\boldsymbol{z}_{n}^{(j)};{\boldsymbol{z}}_{n-1},\boldsymbol{x}_W^{M},\boldsymbol{x}_W^{G_n})$, $\boldsymbol{z}_n$ is given by Eq. \ref{likelihood}. \textcolor{\mycolor}{The likelihood of each particle $\boldsymbol{z}_{n}^{(j)}$ is then calculated as follows:} 
\begin{equation}
\setlength{\abovedisplayskip}{3pt}
\setlength{\belowdisplayskip}{3pt}
\begin{split}
p(\boldsymbol{y}_n|& \boldsymbol{z}_{n|n-1}^{(j)}, \bm{O}_m, \bm{O}_s , \bm{x}_W^{M}, \bm{x}_W^{G_n}) \\
&=\begin{cases}1 - \frac{|d_n^{(j)}|
}{d_{th}}, \quad \text{when}\,|d_n^{(j)}|\leq d_{th}\\0, \quad \text{otherwise}\end{cases}.
\end{split}
\end{equation}
{The particle likelihood is inversely proportional to the distance $|d_n^{(j)}|$, and the particle with a distance $d$ that is greater than a threshold value $d_{th}$ is discarded.} 
The weight for each particle is given as the normalized likelihood:
\begin{equation}
\bm{w}(\boldsymbol{z}_n^{(j)})=\frac{p\left(\boldsymbol{y}_n|\boldsymbol{z}_{n|n-1}^{(j)},\bm{O}_m, \bm{O}_s , \bm{x}_W^{M}, \bm{x}_W^{G_n}\right)}{\sum_{j=1}^Mp\left(\boldsymbol{y}_n|\boldsymbol{z}_{n|n-1}^{(j)},\bm{O}_m, \bm{O}_s , \bm{x}_W^{M}, \bm{x}_W^{G_n}\right)}.
\end{equation}

Given this, the current estimate $\bm{z}_n$ can be given by
\begin{equation}\label{likelihood}
\begin{aligned}{\boldsymbol{z}}_n=\sum \bm{z}_n^{(j)}\bm{w}(\boldsymbol{z}_n^{(j)}).\end{aligned}
\end{equation}

\textcolor{black}{After the pose optimization, given the slave waypoint $\bm{x}^{P}_{m}$ in the master keypoint frame, then the end-effector pose $\bm{x}_W^{G}$ corresponding to this waypoint can be derived as}
\begin{equation}
\setlength{\abovedisplayskip}{3pt}
\setlength{\belowdisplayskip}{3pt}
\begin{split}
\bm{x}_W^{G} = \bm{x}^{M}_{W} \bm{x}^{P}_{M} (\boldsymbol{z}_{{N_T}})^{-1}.
\end{split} 
\end{equation}

\begin{figure}[t!]
\setlength{\abovecaptionskip}{-0.13cm}
\begin{center}
\includegraphics[width=0.50\textwidth]{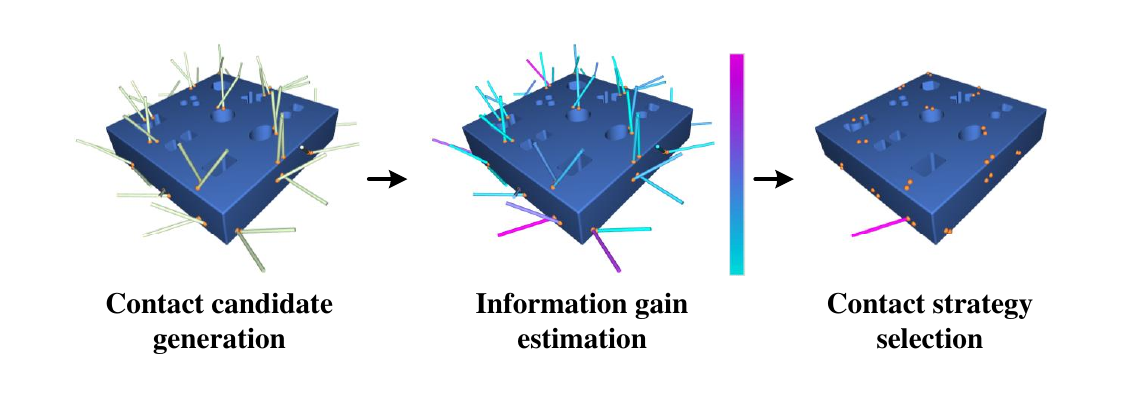}
\end{center}
\caption{{Illustration of the information gain maximization-based contact strategy selection. We initially sample contact strategies on the master objects, followed by the calculation of $IG$ for each strategy. Ultimately, we select the strategy that yields the highest $IG$ for execution.}  }
\label{Information_gain}
\end{figure}

\begin{table*}[t]
\caption{Success rates on RLbench with single-task training paradigm. "Obs-act", "Template", and "Video" indicate paired observation-action sequences, code templates, and videos performing subtasks.}
\label{RLBench}
\begin{minipage}{\textwidth}

\makeatletter\def\@captype{table}
\begin{subtable}[t]{\textwidth}
\resizebox{\linewidth}{!}{
    {\fontsize{8}{10}\selectfont
        \begin{tabular}{m{2.00cm}<{\centering} | *{7}{m{1.50cm}<{\centering}}}
            \toprule[1.5pt]
            \footnotesize
            
            Methods& {R3M-DP} & {DP}   & {GraphIRL} & {CaP}                      & {Demo2Code} & Ours$_{1v}$ & Ours$_{5v}$                     \\
            \midrule
            Overall & $0.65 (\pm 0.29)$ & $0.66 (\pm 0.28)$ & $0.12 (\pm 0.12)$ & $0.44 (\pm 0.33)$ & $0.49 (\pm 0.32)$ & $\bm{0.75 (\pm 0.13)}$ & $\bm{0.87 (\pm 0.11)}$ \\
            \bottomrule
        \end{tabular}
    }
}
\end{subtable}

\makeatletter\def\@captype{table}
\begin{subtable}[t]{\textwidth}
\centering
\newcommand{\myfontsize}{\scriptsize}
\resizebox{\linewidth}{!}{
    {\myfontsize
        \begin{tabular}{@{}lcccccccc@{}}
                \toprule
                Methods & \thead{\myfontsize Type of\\demos } & \thead{\myfontsize Num of\\demos }&   \thead{\myfontsize Reach\\ target}  &   \thead{\myfontsize Take lid off\\ saucepan} & 
                \thead{\myfontsize Pick\\ up cup} & 
                \thead{\myfontsize Toilet\\seat up} & 
                \thead{\myfontsize Open\\box} & 
                \thead{\myfontsize Open\\door} \\
                \midrule
                R3M-DP & Obs-act & 100 & 1.00  & 1.00  & 0.87  & 0.75  & 0.61  & 0.70\\ 
DP & Obs-act & 100 & 1.00  & 1.00  & 0.85  & 0.77  & 0.57  & 0.67 \\  
GraphIRL & Video & 100 &0.39 & 0.14 & 0.23 & 0.03 & 0.03 & 0.21\\ 
CaP & Template & 5 &   {0.95} &  {0.90} & 0.58 & 0.05 & 0.12 & 0.65\\ 
Demo2Code & Video & 5 &0.94 & 0.86 &  {0.65} & 0.06 &  {0.19} &  {0.83} \\ 
\rowcolor{gray!25} \textbf{Ours$_{1v}$} & \textbf{Video} & \textbf{1} & \textbf{0.94} & \textbf{0.92} & \textbf{0.76} & \textbf{0.77} & \textbf{0.73} & \textbf{0.87}\\ 
\rowcolor{gray!25} \textbf{Ours$_{5v}$} & \textbf{Video} & \textbf{5} & \textbf{1.00} & \textbf{1.00} & \textbf{0.92} & \textbf{0.87} & \textbf{0.85} & \textbf{0.95}\\
                
                \toprule
                
                {Methods} & \thead{\myfontsize Type of\\demos } & \thead{\myfontsize Num of\\demos}& 
                \thead{\myfontsize Meat off \\ grill} &\thead{\myfontsize Open\\drawer} & 
                \thead{\myfontsize Open\\grill} & 
                \thead{\myfontsize Open\\microwave} & 
                \thead{\myfontsize Open\\oven} & 
                \thead{\myfontsize Knife on\\board}\\
                \midrule
                R3M-DP & Obs-act & 100 & 0.74  & 0.81  & 0.75  & 0.17  & 0.14  & 0.33\\ 
DP & Obs-act & 100 & 0.83  & 0.79  & 0.73  & 0.21  & 0.10  & 0.41 \\ 
GraphIRL & Video & 100 &0.16 & 0.18 & 0.04 & 0.04 & 0.02 & 0.00 \\ 
CaP & Template & 5 &0.35 & 0.17 & 0.46 & 0.12 & 0.16 & 0.78 \\ 
Demo2Code & Video & 5& {0.57} &  {0.22} &  {0.40} &  {0.14} &  {0.21} &  {0.79}\\  
\rowcolor{gray!25} \textbf{Ours$_{\bm{1v}}$} & \textbf{Video} & \textbf{1} & \textbf{0.77} & \textbf{0.69} & \textbf{0.69} & \textbf{0.51} & \textbf{0.56} & \textbf{0.84} \\ 
\rowcolor{gray!25} \textbf{Ours$_{\bm{5v}}$} & \textbf{Video} & \textbf{5} & \textbf{0.89} & \textbf{0.90} & \textbf{0.91} & \textbf{0.63} & \textbf{0.67} & \textbf{0.89} \\
                \bottomrule
            \end{tabular}
        }
    }
\end{subtable}
\end{minipage}
\end{table*}

\subsubsection{Information gain maximization-based contact strategy selection} 
\textcolor{\mycolor}{Current contact-based pose estimation methods depend on predefined contact sequences, which restrict their applicability to novel objects, which constrain their applicability to unfamiliar objects and limit their generalization capabilities}. Moreover, these predefined sequences restrict their ability to adaptively optimize the contact strategy based on estimation results, thus failing to fully exploit the effectiveness of each contact, necessitating either additional contact iterations or compromising estimation accuracy.

To address these limitations, we propose a contact selection method based on information gain maximization. 
This approach selects the effective contact strategy $\psi$ from a candidate set $\Psi$ by optimizing the information gain (${\rm IG}$). {${\rm IG}$ serves as a quantitative measure of the reduction in the uncertainty of estimation}, calculated as the differential in information entropy $H$ between prior and posterior estimation results. {The ${\rm IG}$ achieved by leveraging the contact strategy $\psi$ is expressed as:}
\begin{equation}
\setlength{\abovedisplayskip}{3pt}
\setlength{\belowdisplayskip}{3pt}
\begin{split}
{\rm IG}(Z|\Psi=\psi) &= H(Z) - H(Z|\Psi=\psi) \\
&= -\sum_{j} \boldsymbol{p}(\boldsymbol{z}^{(j)}) \log \boldsymbol{p}(\boldsymbol{z}^{(j)})  \\
&+ \sum_{j} \boldsymbol{p}(\boldsymbol{z}^{(j)}|\psi) \log \boldsymbol{p}(\boldsymbol{z}^{(j)}|\psi). \\
\end{split} 
\end{equation}
{The probability $\boldsymbol{p}(\boldsymbol{z}^{(j)})$ of each particle $\boldsymbol{z}^{(j)}$ is uniformly distributed, each with a value of $\frac{1}{M}$, where $M$ denotes the total number of particles. The posterior probability $\boldsymbol{p}(\boldsymbol{z}^{(j)}|\psi)$ of each particle is represented by the updated weight, derived from the normalized likelihood. The information gain ${\rm IG}$ is expressed as:} 
\begin{equation}
\setlength{\abovedisplayskip}{3pt}
\setlength{\belowdisplayskip}{3pt}
\begin{split}
{\rm IG}(Z|\Psi=\psi) &= - \log \boldsymbol{p}(\frac{1}{M})  \\
&+ \sum_{j} \bm{w}(\boldsymbol{z}^{(j)}|\psi) \log \bm{w}(\boldsymbol{z}^{(j)}|\psi). \\
\end{split} 
\label{IG}
\end{equation}
{Consequently, the optimization of the ${\rm IG}$ can be reformulated as a minimization problem of the entropy $H(Z|\Psi=\psi)$ of the posterior particle weight distribution post-contact, with the objective of reducing uncertainty in the pose estimation.}

\textcolor{\mycolor}{Our approach first samples representative contact candidates, followed by selecting the contact strategy that minimizes the entropy of the particle weight distribution. Figure \ref{Information_gain} exhibits the process of our contact strategy selection.} To generate contact candidates, we sample $N_p$ contact positions on the master object's surface.
{A local spherical coordinate system is established at each of these positions, centering the origin at the contact points and aligning the $Z$-axis with the normal vector of the surrounding surface. An arbitrary vector orthogonal to the normal is designated as the $X$-axis. Within this framework}, the contacting orientation candidates of the slave object are generated by sampling its $\vec{\bm{z}_s}$-axis direction using azimuth and elevation angles, and the $\vec{\bm{x}_s}$-axis direction is sampled based on the $\vec{\bm{z}_s}$-axis. $N_o$ distinct orientations are generated for each contact position.



\textcolor{\mycolor}{For each contact candidate, we randomly sample $N_c$ particles and calculate the corresponding contact conditions, which serve as potential contact scenarios.} 
To efficiently estimate the ${\rm IG}$, the original particle set $\left\{\boldsymbol{z}^{(j)}\right\}_{j=1}^M$ is uniformly downsampled to a subset $\left\{\boldsymbol{\tilde{z}}^{(j)}\right\}_{j=1}^{N_d}$, where $N_d \ll M$. \textcolor{\mycolor}{The entropy of this downsampled subset is then calculated under each of the $N_c$ contact scenarios, and the averaged entropy across these contact scenarios is computed to characterize the entropy for the given contact candidate}. Finally, the contact candidate with the minimum mean entropy is selected for execution.

\begin{table*}[t]
\caption{Success rates on RLbench with multi-task training paradigm. "Obs-act", "Template", and "Video" indicate paired observation-action sequences, code templates, and videos performing subtasks.}
\label{RLBench_multi}
\begin{minipage}{\textwidth}

\makeatletter\def\@captype{table}
\begin{subtable}[t]{\textwidth}
    \resizebox{\linewidth}{!}{
        {\fontsize{8}{10}\selectfont
            \begin{tabular}{m{2.00cm}<{\centering} | *{7}{m{1.80cm}<{\centering}}}
                \toprule[1.5pt]
                \footnotesize
                
                Methods& {R3M-DP} & {DP}   & {CaP}                      & {Demo2Code} & Ours$_{5v}$                     \\
                \midrule
                Overall & $0.13 (\pm 0.12)$ & $0.15 (\pm 0.13)$ & $0.31 (\pm 0.33)$ & $0.34 (\pm 0.32)$ & $\bm{0.83 (\pm 0.12)}$ \\
                \bottomrule
            \end{tabular}
        }
    }
\end{subtable}

\makeatletter\def\@captype{table}
\begin{subtable}[t]{\textwidth}
    \centering
    \newcommand{\myfontsize}{\scriptsize}
    \resizebox{\linewidth}{!}{
        {\myfontsize
            \begin{tabular}{@{}lcccccccccccc@{}}
                    \toprule
                    Methods & \thead{\myfontsize Type of\\demos } & \thead{\myfontsize Num of\\demos}&   \thead{\myfontsize Reach\\ target}  &   \thead{\myfontsize Take lid off\\ saucepan} & 
                    \thead{\myfontsize Pick\\ up cup} & 
                    \thead{\myfontsize Toilet\\seat up} & 
                    \thead{\myfontsize Open\\box} & 
                    \thead{\myfontsize Open\\door} \\
                    \midrule
                    R3M-DP & Obs-act & 100 & 0.37 & 0.20 & 0.20 & 0.07 & 0.02 & 0.25\\ 
DP & Obs-act & 100 & 0.43 & 0.25 & 0.24 & 0.05 & 0.04 & 0.22 \\  
CaP & Template & 5 &   {0.85} &  {0.78} & 0.39 & 0.01 & 0.05 & 0.37\\ 
Demo2Code & Video & 5 &0.83 & 0.77 &  {0.45} & 0.02 &  {0.06} &  {0.43} \\ 
\rowcolor{gray!25} \textbf{Ours$_{5v}$} & \textbf{Video} & \textbf{5} & \textbf{1.00} & \textbf{0.98} & \textbf{0.85} & \textbf{0.83} & \textbf{0.79} & \textbf{0.92}\\
                    
                    \toprule
                    
                    {Methods} & \thead{\myfontsize Type of\\demos } & \thead{\myfontsize Num of\\demos}& 
                    \thead{\myfontsize Meat off \\ grill} &\thead{\myfontsize Open\\drawer} & 
                    \thead{\myfontsize Open\\grill} & 
                    \thead{\myfontsize Open\\microwave} & 
                    \thead{\myfontsize Open\\oven} & 
                    \thead{\myfontsize Knife on\\board}\\
                    \midrule
                    R3M-DP & Obs-act & 100 & 0.15 & 0.25 & 0.07 & 0.03 & 0.00 & 0.00\\ 
DP & Obs-act & 100 & 0.17 & 0.28 & 0.09 & 0.07 & 0.00 & 0.00 \\ 
CaP & Template & 5 &0.18 & 0.10 & 0.31 & 0.07 & 0.08 & 0.54 \\ 
Demo2Code & Video & 5& {0.28} &  {0.15} &  {0.31} &  {0.09} &  {0.14} &  {0.59}\\  
\rowcolor{gray!25} \textbf{Ours$_{\bm{5v}}$} & \textbf{Video} & \textbf{5} & \textbf{0.85} & \textbf{0.84} & \textbf{0.83} & \textbf{0.59} & \textbf{0.62} & \textbf{0.83} \\
                    \bottomrule
                \end{tabular}
            }
        }
    \end{subtable}
\end{minipage}
\end{table*}

\section{Experiments} \label{Exp.}

\textcolor{black}{In this section, we aim to answer the following questions:}

\textcolor{black}{(\uppercase\expandafter{\romannumeral1}) Can \nickname learn fine-grained action levels from human videos and complete tasks with a high success rate?}

\textcolor{black}{(\uppercase\expandafter{\romannumeral2}) Can \nickname generalize learned skills across diverse environments to distinct tasks?}

\textcolor{black}{(\uppercase\expandafter{\romannumeral3}) Can \nickname accomplish complex tasks, particularly those with long-horizon or high-precision demands?}

\textcolor{black}{(\uppercase\expandafter{\romannumeral4}) Can \nickname learn skills efficiently and robustly?}

\textcolor{black}{(\uppercase\expandafter{\romannumeral5}) Which design decisions in \nickname matter most for effectively learning skills from human videos?}

\textcolor{black}{The first research question is addressed in Sec. \nameref{Sec::Manipulation Task Learning}. Subsequently, generalization experiments are presented in Sec. \nameref{Sec::Real_Manip} and Sec. \nameref{Task_generalization}. To validate the capability of our proposed \nickname in executing complex tasks, experimental studies are conducted and detailed in Sec. \nameref{Sec::Long-Horizon Tasks} and Sec. \nameref{Sec::Real-world High Precision Tasks}. The efficiency of the system is evaluated in Sec. \nameref{Efficiency}, while comprehensive robustness assessments are performed in Sec. \nameref{Sec::viewpoint}, Sec. \nameref{Sec::cumulative errors}, and Sec. \nameref{Robustness}. Lastly, Sec. \nameref{Sec::Ablation} presents extensive ablation studies to substantiate the design choices of our \nickname framework.}

\begin{figure*}[h]
    \setlength{\abovecaptionskip}{-0.13cm}
    \begin{center}
        \includegraphics[width=1.0\textwidth]{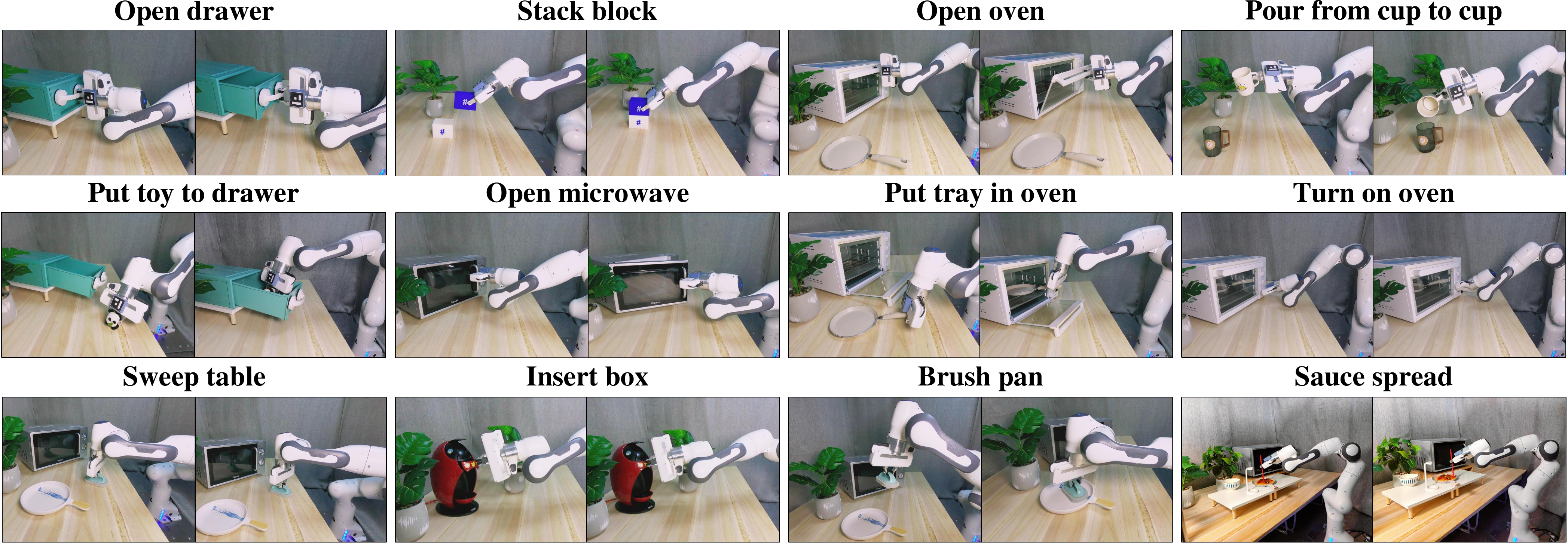}
    \end{center}
    \caption{{Example qualitative results for real-world manipulation task in seen environments.}  }
    \label{seen_result_fig}
\end{figure*}

\begin{figure*}[h]
    \setlength{\abovecaptionskip}{-0.13cm}
    \begin{center}
        \includegraphics[width=1.0\textwidth]{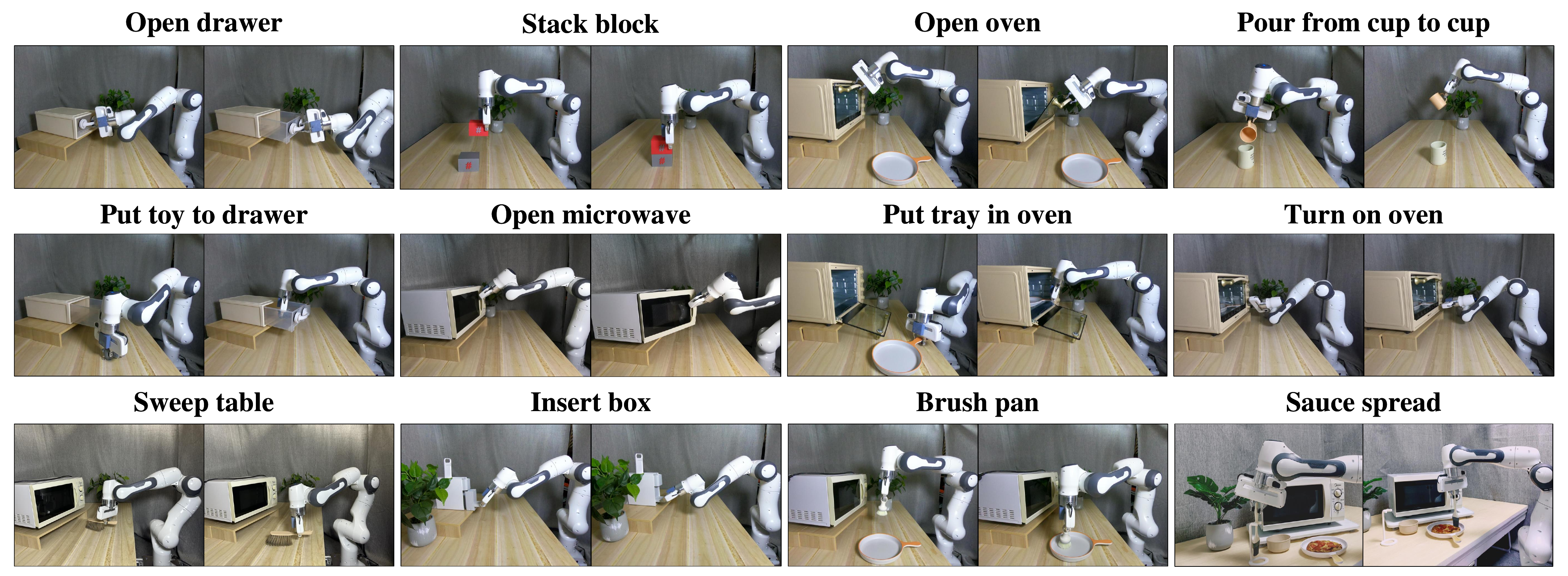}
    \end{center}
    \caption{\textcolor{black}{Example qualitative results for real-world manipulation task in unseen environments.}  }
    \label{unseen_result_fig}
\end{figure*}

\subsection{Experimental setup}
\subsubsection{Implementation details} \label{Implementation}
{The keyframes are obtained by sampling video frames every three seconds. In instances where the number of extracted keyframes falls below five}, we uniformly sample five keyframes. Then, the videos are divided into multiple segments using a threshold $\epsilon$ of $2 cm$. Segments with hand motion trajectory lengths below $\gamma=5 cm$ are discarded. During the grasping constraint learning phase, the number of regions $N_c$ is determined by the VLMs.
In the context of the skill adapter, the maximum number of iterations is set to $N_I=4$. \textcolor{\mycolor}{For pose optimization within the skill refiner, we employ a gripper with a $100 N$ grasping force and $1mm$ silicone-tipped fingers, thus assuming negligible in-hand pose variation. We introduce Gaussian noise with $\sigma = 1.0mm$ for translation, $\sigma = 0.01 rad$ for rotation, and utilize a particle filter with $M = 500$ samples for robust pose estimation.} To generate the contact candidates, we sample $N_p=4$ contact positions, and for each position, we further sample $N_o=12$ orientations. To improve the entropy evaluation efficiency, we utilize $N_c=4$ particles to calculate potential collision conditions. The entropy assessment for each condition is then conducted using a downsampled subset $\left\{\boldsymbol{\tilde{z}}^{(j)}\right\}_{j=1}^{N_d}$, where $N_d=10$ points.

\subsubsection{Real-world experimental setup}
Experiments are conducted on Franka Emika \citep{haddadin2022franka}, utilizing three RGB-D cameras (ORBBEC Femto Bolt) for environment observation: one at the top right of the table, one at the top left, and one mounted on the robot's wrist. {All cameras simultaneously capture and transmit real-time RGB-D observations at a frequency of 30 Hz.}

\subsubsection{Baselines} \textcolor{black}{\nickname is compared with five representative methods: (1) {R3M-DP, which utilizes the pre-trained R3M visual representation \citep {nair2022r3m} with the SOTA diffusion policy \citep{chi2023diffusion}}; (2) Diffusion Policy (DP) \citep{chi2023diffusion}, a SOTA end-to-end policy method. R3M-DP \citep {nair2022r3m} and DP \citep {chi2023diffusion} employ a CNN-based network architecture for its robustness across diverse tasks. These methods are trained on robot demonstrations using default hyper-parameters, robot demonstrations consist of paired observation and action sequences.  (3) GraphIRL \citep{kumar2023graph}, a method that employs graph abstraction and learns reward functions for reinforcement learning (RL), GraphIRL is trained in simulators with paired robot videos; (4) Code as Policy (CaP) \citep{liang2023code}, an LLM-driven method that re-composes API calls to generate new policy code, it employs natural language instruction directly for reasoning; and (5) Demo2code \citep{wang2024Demo2code}, an LLM-driven planner method that translates demonstrations into task code. We modify it to generate code from textual video analysis results provided by GPT-4V for Robotics \citep{wake2023gpt}, a video analysis approach for robotics, enabling Demo2code to learn from human videos. Specifically, the detailed task analysis results and affordance analysis outcomes from GPT-4V for Robotics are incorporated as contextual information within the textual prompt for Demo2code.
\textcolor{\mycolor}{In alignment with CaP, both CaP and Demo2code utilize a set of primitives: move to pos, rotate by quat, set vel, open gripper, close gripper, pick obj, and place at pos. }
Demo2code and our method learn skills through human videos in real-world experiments and robot videos in simulation experiments. 
To facilitate an intuitive comparison, our skill refiner is only included in the high-precision tasks.}

\subsection{Manipulation Task Learning} \label{Sec::Manipulation Task Learning}
\subsubsection{Experimental setup} To assess our approach on challenging robotic manipulation tasks, the RLBench \citep{james2020rlbench} benchmark is utilized for simulation tasks.
\textcolor{\mycolor}{The positions and orientations of objects are randomly initialized within specified constraints.} Specifically, the spatial coordinates of objects are randomly sampled within the operational range of the robotic manipulator on the task workspace. {Concurrently, the rotational orientation of objects is randomly selected from a predetermined set of viable angular configurations}, such as $[-\pi/2,\pi/2]$ for the $Z$-axis of the microwave. {Additionally, the color attributes of the objects are sampled probabilistically}. 
Due to the unavailability of human videos in simulations, Demo2code and our method utilize robot videos captured from a single-camera perspective during demonstrations, incorporating robot gripper trajectories.

\begin{table*}[t]
\caption{Success rates on real-world manipulation experiments. "Obs-act", "Template", and "Video" indicate paired observation-action sequences, code templates, and videos performing subtasks. "SE" and "UE" denote seen and unseen environments. }
\label{Real_mani}
\begin{minipage}{\textwidth}
    
    \makeatletter\def\@captype{table}
    \begin{subtable}[t]{\textwidth}
        \resizebox{\linewidth}{!}{
            {\fontsize{8}{10}\selectfont
                \begin{tabular}{>{\centering\arraybackslash}m{2.00cm} | *{7}{>{\centering\arraybackslash}m{1.50cm}}}
                    \toprule[1.5pt]
                    
                    Methods& R3M-DP & DP   & GraphIRL & CaP                      & Demo2Code & Ours$_{1v}$ & Ours$_{5v}$                                   \\
                    \midrule
                    Overall (SE) & $0.49 (\pm 0.20)$ & $0.55 (\pm 0.21)$ & $0.25 (\pm 0.21)$ & $0.39 (\pm 0.22)$ & $0.43 (\pm 0.21)$ & $\bm{0.73 (\pm 0.07)}$ & $\bm{0.84 (\pm 0.06)}$                    \\
                    Overall (UE) & $0.09 (\pm 0.10)$ & $0.10 (\pm 0.10)$ & $0.07 (\pm 0.08)$ & $0.37 (\pm 0.24)$ & $0.37 (\pm 0.26)$ & $\bm{0.63 (\pm 0.08)}$ & $\bm{0.75 (\pm 0.12)}$ \\
                    
                    \bottomrule
                \end{tabular}
            }
        }
    \end{subtable}
    
    \makeatletter\def\@captype{table}
    \begin{subtable}[t]{\textwidth}
        \centering
        \resizebox{\linewidth}{!}{
            {\fontsize{8}{10}\selectfont
                \begin{tabular}{lcccccccccccccccccccc}
                        \toprule
                        \multirow{2}{*}{Methods} & \multirow{2}{*}{\thead{Type of\\demos }} & \multirow{2}{*}{\thead{Num of\\demos}} & \multicolumn{2}{c}{\thead{Open\\drawer}} & \multicolumn{2}{c}{\thead{Stack\\block}} & \multicolumn{2}{c}{\thead{Open\\oven}} &\multicolumn{2}{c}{\thead{Put fruit \\ on plate}} & \multicolumn{2}{c}{\thead{Press\\button}} & \multicolumn{2}{c}{\thead{Close\\microwave}} & \multicolumn{2}{c}{\thead{Put tray\\in oven}}\\
                        \cmidrule(lr){4-5} \cmidrule(lr){6-7} \cmidrule(lr){8-9} \cmidrule(lr){10-11} \cmidrule(lr){12-13} \cmidrule(lr){14-15}
                        \cmidrule(lr){16-17} 
                        & & & SE & UE & SE & UE  & SE & UE &SE & UE  & SE & UE  & SE & UE  & SE & UE \\
                        
                        \midrule
                        R3M-DP & Obs-act & 100 & 0.2 & 0.1 &  {0.6} & 0.2 & 0.3 & 0.0 &  {0.8} & 0.3 &  {0.7} & 0.2 & 0.2 & 0.0 &  {0.4} & 0.0 \\
DP & Obs-act & 100 & 0.3 & 0.1 &  {0.6} & 0.2 &  {0.4} & 0.1 &  {0.9} & 0.4 &  {0.7} & 0.1 & 0.3 & 0.0 &  {0.4} & 0.0 \\
GraphIRL & Video & 100 & 0.2 & 0.0 & 0.4 & 0.1 & 0.0 & 0.0 & {0.7} & 0.2 & 0.4 & 0.2 &  {0.0} &  {0.0} & 0.2 & 0.0 \\
CaP & Template & 5 & 0.3 &  {0.3} & 0.5 &  {0.5} & 0.3 &  {0.2} & 0.8 &  {0.8} &  {0.7} & 0.7 & 0.1 &  {0.1} & 0.2 & 0.1 \\
Demo2Code & Video & 5 &  {0.3} & 0.3 & 0.5 & 0.4 & 0.3 & 0.1 & 0.8 &  {0.9} & {0.8} &  {0.8} & 0.2 &  {0.1} & 0.3 &  {0.2} \\
\rowcolor{gray!25} \textbf{Ours$_{1v}$} & \textbf{Video} & \textbf{1} & \textbf{0.8} & \textbf{0.8} & \textbf{0.7} & \textbf{0.6} & \textbf{0.7} & \textbf{0.6} & \textbf{0.8} & \textbf{0.7} & \textbf{0.8} & \textbf{0.6} & \textbf{0.7} & \textbf{0.6} & \textbf{0.7} & \textbf{0.6} \\
\rowcolor{gray!25} \textbf{Ours$_{5v}$} & \textbf{Video} & \textbf{5} & \textbf{0.9} & \textbf{0.9} & \textbf{0.9} & \textbf{0.8} & \textbf{0.8} & \textbf{0.7} & \textbf{0.9} & \textbf{0.9} & \textbf{0.9} & \textbf{0.9} & \textbf{0.8} & \textbf{0.7} & \textbf{0.8} & \textbf{0.7} \\
                        \toprule
                        \multirow{2}{*}{Methods} & \multirow{2}{*}{\thead{Type of\\demos }} & \multirow{2}{*}{\thead{Num of\\demos}}  & \multicolumn{2}{c}{\thead{Turn on\\oven}} & \multicolumn{2}{c}{\thead{Sweep\\table}} & \multicolumn{2}{c}{\thead{Insert\\box}} & \multicolumn{2}{c}{\thead{Brush\\pan}} & \multicolumn{2}{c}{\thead{Sauce\\spread}} & \multicolumn{2}{c}{\thead{Put toy\\to drawer}} & \multicolumn{2}{c}{\thead{Pour from\\cup to cup}}\\
                        \cmidrule(lr){4-5} \cmidrule(lr){6-7} \cmidrule(lr){8-9} \cmidrule(lr){10-11} \cmidrule(lr){12-13} \cmidrule(lr){14-15}
                        \cmidrule(lr){16-17}
                        &&&SE&UE&SE&UE&SE&UE&SE&UE&SE&UE&SE&UE&SE&UE \\
                        \midrule
                        R3M-DP & Obs-act & 100 & 0.2 & 0.0 & 0.7 & 0.2 &  {0.4} & 0.0 & 0.6 & 0.1 & 0.6 & 0.1 & 0.6 & 0.1 &  {0.5} & 0.0 \\
DP & Obs-act & 100 &  {0.3} & 0.0 &  {0.8} & 0.1 & 0.3 & 0.1 &  {0.7} & 0.1 &  {0.7} & 0.0 &  {0.7} & 0.1 & {0.6} & 0.1 \\
GraphIRL & Video & 100 &  {0.2} & 0.1 & 0.5 & 0.2 & 0.0 & 0.0 & 0.2 & 0.0 & 0.2 & 0.1 & 0.4 & 0.1 & 0.1 & 0.0 \\
CaP & Template & 5 &  {0.3} &  {0.3} & 0.6 & 0.5 & 0.1 & 0.1 & 0.3 &  {0.4} & 0.3 & 0.3 & 0.6 &  {0.7} & 0.4 &  {0.2} \\
Demo2Code & Video & 5 & 0.2 & 0.1 & 0.6 &  {0.6} & 0.3 &  {0.2} & 0.4 & 0.3 & 0.3 &  {0.4} &  {0.7} & 0.6 & 0.3 &  {0.2} \\
\rowcolor{gray!25}\textbf{Ours$_{1v}$} & \textbf{Video} & \textbf{1} & \textbf{0.7} & \textbf{0.6} & \textbf{0.8} & \textbf{0.8} & \textbf{0.6} & \textbf{0.5} & \textbf{0.7} & \textbf{0.6} & \textbf{0.7} & \textbf{0.6} & \textbf{0.8} & \textbf{0.6} & \textbf{0.7} & \textbf{0.6} \\
\rowcolor{gray!25}\textbf{Ours$_{5v}$} & \textbf{Video} & \textbf{5} & \textbf{0.8} & \textbf{0.8} & \textbf{0.9} & \textbf{0.9} & \textbf{0.7} & \textbf{0.5} & \textbf{0.8} & \textbf{0.7} & \textbf{0.8} & \textbf{0.6} & \textbf{0.9} & \textbf{0.8} & \textbf{0.9} & \textbf{0.7} \\
                        \bottomrule
                    \end{tabular}
                }
            }
        \end{subtable}
    \end{minipage}
\end{table*}

\subsubsection{Single-task results} \textcolor{\mycolor}{Comparison methods are learned and evaluated on discrete tasks independently. We investigate the capacity of \nickname to acquire skills from a limited collection of video demonstrations}. Our evaluation encompasses a diverse set of 12 manipulation tasks, as detailed in Table \ref{RLBench}. {The empirical results indicate that our method, utilizing only a single human demonstration video, achieves a 75\% success rate, which is comparable to DP and R3M-DP trained on 100 robot demonstrations and significantly outperforms other approaches. Our \nickname learned on five human demonstration videos, exhibits improved performance, and significantly surpasses both R3M-DP and DP by a margin of over 22\% in overall success rates}. Compared to CaP and Demo2code, our method demonstrates an improvement exceeding 39\%. \textcolor{\mycolor}{These approaches, which rely on VLMs/LLMs as planners, typically yield substantially lower success rates on tasks requiring an understanding of intrinsic physical constraints}, exemplified by 'open drawer', indicating their inability to infer physical constraints from observations, resulting in diminished performance when object configurations deviate from demonstrated scenarios. Our method is able to learn the object-dependent semantic and geometric constraints from human demonstration videos. thus demonstrating robust task completion despite variations in object placement and configuration. Furthermore, the provided templates of CaP would require the manual development of low-level skills, which demands significant human efforts. In contrast, our method autonomously learns generalizable skills through a small number of low-cost human demonstration videos, and can subsequently complete tasks with a high success rate.

\subsubsection{Multi-task results} \textcolor{\mycolor}{Multi-task capability is a fundamental criterion for evaluating robotic policies. To assess the multi-task proficiency of the comparative methods, we train both DP and R3M-DP on the complete set of task training data, conditioning them on task descriptions.} For CaP, we provide templates for all tasks in the prompt, while for Demo2code, we initially process each video individually, generating code for each task, and subsequently incorporating the aggregated task codes as exemplars within the prompt during testing. 
    {Experimental results, as provided in Table \ref{RLBench_multi}, reveal that DP and R3M-DP fail to effectively manage different skills and generalize to test examples}, and our method significantly surpasses both R3M-DP and DP by a margin of over 68\% in overall success rates, despite these methods being trained on 100 robot demonstrations. {CaP and Demo2Code also exhibit a marked decline in multi-task performance compared to single-task scenarios, excessive examples across diverse tasks can overwhelm LLMs/VLMs}, potentially resulting in the misapplication of task exemplars or promoting rote repetition rather than adaptive reasoning for specific scenarios. Our method demonstrates an improvement exceeding 49\%, highlighting the significant multi-task performance enhancements facilitated by the \nickname framework.
\textcolor{\mycolor}{Our method demonstrates robust performance in multi-task settings by precisely extracting task-relevant knowledge from the knowledge base, thereby facilitating skill adaptation based on the retrieved contextual information.}


\begin{table*}[t]
    \caption{Success rates on real-world unseen manipulation tasks. Methods develop skills from manipulation tasks (Table \ref{Real_mani}) within the SE settings. "Obs-act", "Template", and "Video" indicate observation-action sequences, code templates, and videos performing tasks. }
    \label{table::unseen_tasks}
    \resizebox{\linewidth}{!}{
        {\footnotesize
                \begin{tabular}{>{\centering\arraybackslash}m{2cm}  *{6}{>{\centering\arraybackslash}m{1.3cm}}*{1}{>{\centering\arraybackslash}m{1.8cm}}}
                    \toprule
                    Methods &\thead{  Open \\ fridge} & \thead{ Press \\ switch} & \thead{ Close \\ oven} & \thead{ Put rubbish \\ in bin} & \thead{ Put bottle \\in cabinet} & \thead{ Get peach\\ from fridge} & \thead{Overall } \\
                    \midrule
                    R3M-DP & 0.20 &  {0.00} &  {0.20} &  {0.10} &  {0.10} &  {0.10} & $0.13(\pm0.05)$ \\
                    DP & 0.20 & 0.00 & 0.20 & 0.00 & 0.10 & 0.00 & $0.07(\pm0.09)$ \\
                    CaP& 0.30 & 0.30 & 0.20 & 0.70 & 0.20  & 0.10 & $0.30(\pm0.27)$  \\
                    Demo2Code& 0.30 & 0.30 & 0.30& 0.70& 0.20&  0.20  & $0.35(\pm0.23)$ \\
                    \rowcolor{gray!25} \textbf{Ours$_{5v}$}& \textbf{0.80} & \textbf{0.70} & \textbf{0.60} & \textbf{0.70} & \textbf{0.70} & \textbf{0.60} & $\bm{0.67(\pm0.06)}$   \\
                    \bottomrule
                \end{tabular}
            }
        }
    \end{table*}

    \begin{table*}[t]
\caption{Success rates on long-horizon tasks. "Obs-act", "Template", and "Video" indicate observation-action sequences, code templates, and videos performing tasks. }
\label{Long}
\resizebox{\linewidth}{!}{
{\footnotesize
\begin{tabular}{lccccccccc}
\toprule
Methods & \thead{ Type of \\ demos} & \thead{ Num of \\ demos} &\thead{  Make \\ coffee} & \thead{ Clean \\ table} & \thead{ Make \\ a pie} & \thead{ Wash \\ pan} & \thead{ Make \\slices} & \thead{ Chem.\\ exp.} & \thead{Overall } \\
\midrule
R3M-DP & Obs-act & 100 & 0.10 &  {0.30} &  {0.20} &  {0.10} &  {0.00} &  {0.10} &  $0.13 (\pm 0.09)$ \\
DP & Obs-act & 100 & 0.00 & 0.20 & 0.10 & 0.00 & 0.10 & 0.00 & $0.07 (\pm 0.07)$ \\
GraphIRL& Video & 100 & 0.00 & 0.10 & 0.00 & 0.00 & 0.00 & 0.00 & $0.02 (\pm 0.04)$  \\
CaP& Template & 5 & 0.00 & 0.10 & 0.00 & 0.00 & 0.00  & 0.00 & $0.02 (\pm 0.04)$ \\
Demo2Code& Video & 5 &0.00 &0.10 & 0.00& 0.00& 0.00&  0.00  & $0.02 (\pm 0.04)$\\
\rowcolor{gray!25} \textbf{Ours$_{5v}$}& \textbf{Video} & \textbf{5} & \textbf{0.50} & \textbf{0.70} & \textbf{0.70} & \textbf{0.50} & \textbf{0.60} & \textbf{0.60} & $\bm{0.60 (\pm 0.09)}$  \\
\bottomrule
\end{tabular}
}
}
\end{table*}

\subsection{Generalization across environments} \label{Sec::Real_Manip}
\subsubsection{Experimental setup} 
To validate the real-world performance and cross-environment generalization capability of our \nickname,
\textcolor{\mycolor}{we categorize real-world testing into seen environment (SE) and unseen environment (UE) scenarios}. The SE scenario enables testing in the environment where demonstrations were collected, whereas the UE scenario challenges the system in a distinct environment characterized by different objects and layouts. The comparative methods conduct learning for each task and perform testing separately under SE and UE settings. \textcolor{\mycolor}{Success criteria are determined through human evaluation, and the success rate is computed based on 10 randomized object positions and orientations. }

\subsubsection{Results} \textcolor{\mycolor}{We conduct a comprehensive experimental evaluation spanning 14 challenging real-world manipulation tasks, carefully selected from recent robotics research \citep{ahn2022can, xiao2022robotic, brohan2022rt, yu2023scaling}}.
The quantitative analysis, as elucidated in Table \ref{Real_mani}, demonstrates the superior performance of \nickname. Our method exhibits  superior performance in all 14 tasks even with a single human demonstration video as input, outperforming comparative methods by at least 18\% in overall success rate. {Moreover, when trained on a corpus of five human demonstration videos, it consistently surpasses baseline approaches across the entire evaluated task spectrum.}
Specifically, \nickname exhibits substantial improvements in performance metrics, with an increase exceeding 29\% in the SE setting and more than 38\% in the UE setting. {These results highlight the robustness and generalizability of \nickname in diverse real-world manipulation scenarios.}
{Learning-based methods, such as DP and R3M-DP, exhibit a significant performance decline when transitioning from seen to unseen environments}, while our method fully utilizes the generalization capabilities of VLMs and vision foundation models to effectively identify task-relevant objects and infer task constraint adaptation in unseen environments.
    Moreover, the learned skills are represented using object properties, such as bounding boxes, {which facilitate skill adaptation to novel objects based on their characteristics}, as exemplified in the 'Brush pan' task, where the brush's height relative to the pan is dynamically calibrated according to the brush's length.
    Furthermore, the skill adapter iteratively revises and updates learned skills. \textcolor{\mycolor}{Our approach demonstrates robust generalization, maintaining consistent performance when transferred to new environments.}
Results underscore the outstanding ability of \nickname to acquire skills from human videos and adapt them to unseen environments. 
The qualitative results are presented in Figure \ref{seen_result_fig} and Figure \ref{unseen_result_fig}.


        \begin{figure*}[h]
            \setlength{\abovecaptionskip}{-0.13cm}
            \begin{center}
                \includegraphics[width=0.98\textwidth]{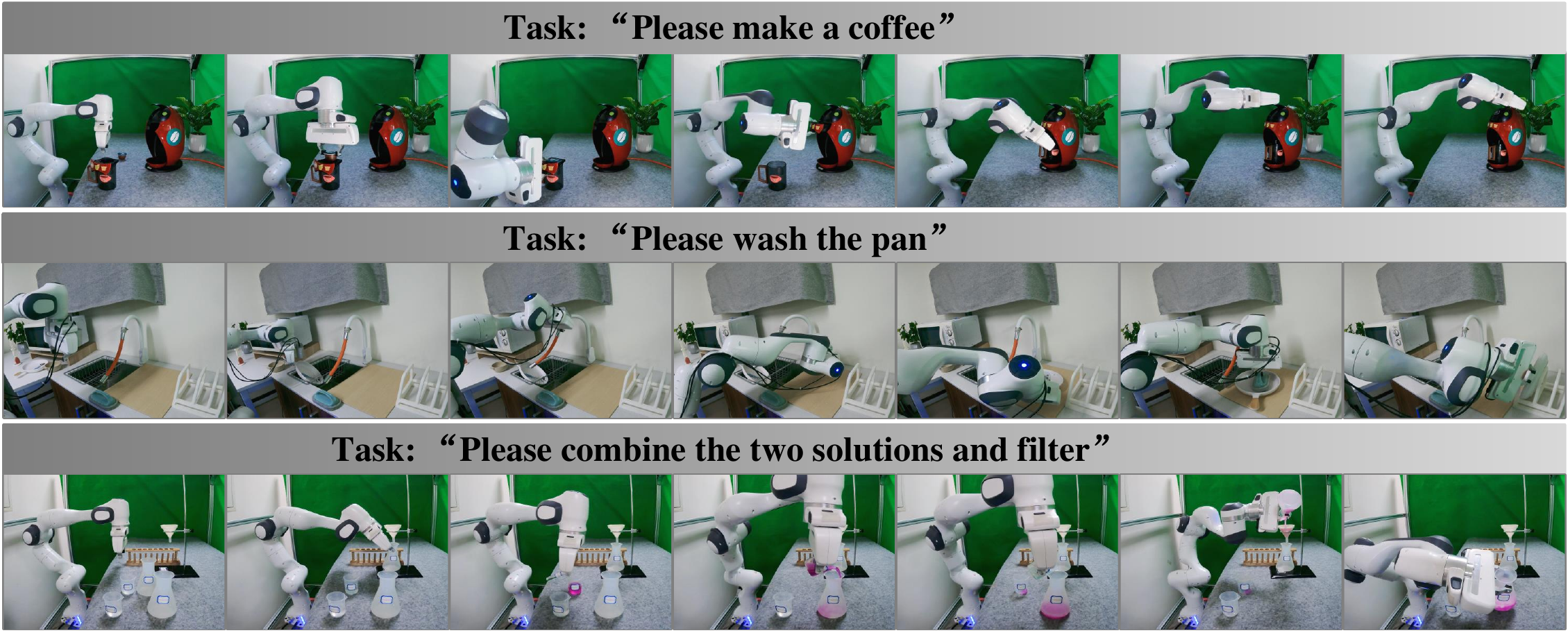}
            \end{center}
            \caption{\textcolor{black}{Example qualitative results for long horizon task. Our \nickname exhibits robust performance in executing long-horizon tasks.}  }
            \label{long_horizon_task}
        \end{figure*}
        
        \subsection{Generalization across Tasks}  \label{Task_generalization}
        \subsubsection{Experimental setup} The demonstrations from seen environments in \nameref{Sec::Real_Manip} are utilized for policy learning, and comparison methods are evaluated on unseen tasks. Learning-based approaches (R3M-DP, DP, and GraphIRL) incorporate instructions as additional conditions and are trained on data from all seen tasks. {LLM-based methods (CaP and Demo2Code) process demonstrations by integrating instructions and corresponding templates into their prompts. All other experimental settings remain consistent with those used in the real-world manipulation task.}

\begin{table*}[ht]
\caption{\textcolor{black}{Success rates for high-precision tasks with seen environments. "Obs-act", "Template", and "Video" indicate paired observation-action sequences, code templates, and videos performing subtasks. \textcolor[RGB]{1, 122, 177}{$\bullet$}  denote the utilization of \nameref{Sec::High}.}}
\label{table::HiP}
\begin{minipage}{\textwidth}
\small

\makeatletter\def\@captype{table}

\makeatletter\def\@captype{table}
\begin{subtable}[t]{\textwidth}
    \centering
    \resizebox{\linewidth}{!}{
        {\fontsize{8}{10}\selectfont
            \begin{tabular}{@{}lcccccccccc@{}}
                    \toprule
                    Methods & \thead{Type of\\demos } & \thead{Num of\\demos} & Rectangle & Round &
                    Oval & 
                    Hexagon & 
                    Arch & Star & \thead{Square \\ Circle} & Overall \\
                    \midrule 
                    R3M-DP & Obs-act & 100 &  0.50 &  {0.60} &  {0.40} &  {0.50} &  {0.30} &  {0.60} & 0.40 & $0.47 (\pm 0.11)$ \\
                    DP & Obs-act & 100 & 0.60 & 0.70 & 0.40 & 0.40 & 0.30 & 0.70 & 0.30 & $0.48 (\pm 0.17)$ \\
                    GraphIRL& Video & 100 & 0.30  & 0.40 & 0.10 & 0.20 & 0.00 & 0.30 & 0.10 & $0.20 (\pm 0.14)$  \\
                    CaP& Template & 5 & 0.10 &  0.60 & 0.10 & 0.00 & 0.20 & 0.10  & 0.00 & $0.16 (\pm 0.20)$ \\
                    Demo2Code& Video & 5 & 0.10 &  0.70 & 0.20 & 0.30 & 0.10 & 0.00  & 0.10 & $0.22 (\pm 0.23)$ \\
                    \rowcolor{gray!25} \textbf{Ours$_{5v}$} & \textbf{Video} & \textbf{5} & \textbf{0.60} & \textbf{0.80} & \textbf{0.50} & \textbf{0.40} & \textbf{0.30} & \textbf{0.60} & \textbf{0.40} & $\bm{0.51 (\pm 0.17)}$  \\
                    \rowcolor{gray!25} \textbf{Ours$_{5v}$} \textcolor[RGB]{1, 122, 177}{$\bullet$}& \textbf{Video} & \textbf{5} & \textbf{0.90} & \textbf{1.00} & \textbf{0.80} & \textbf{0.80} & \textbf{0.70} & \textbf{1.00} & \textbf{0.60} & $\bm{0.82 (\pm 0.15)}$  \\
                    \bottomrule
                \end{tabular}
            }
        }
    \end{subtable}
\end{minipage}
\end{table*}

        \subsubsection{Results} The execution environments for robotic systems often exhibit substantial disparities from human demonstration scenarios, particularly in terms of object characteristics and task specifications. \textcolor{\mycolor}{To evaluate the generalization capabilities across tasks of our proposed method, a series of experiments are conducted to assess its performance on previously unseen tasks.}
            The experimental results, as presented in Table \ref{table::unseen_tasks}, indicate that R3M-DP and DP achieve success rates of only 13\% and 7\% respectively. CaP and Demo2Code attain success rates of 30\% and 35\% respectively. {While CaP and Demo2Code leverage the extensive prior knowledge and strong reasoning capabilities of LLMs and VLMs to update high-level plans, these methods suffer from limitations in generalizing fine-grained actions}, thus failing to achieve robust generalization across diverse tasks.
            Our proposed method, \nickname, employs a skill adapter that initially maps keypoints using a region-to-keypoint approach, {thereby facilitating the transfer of skill affordance to target objects.} Subsequently, it adapts skills to novel tasks through an iterative comparison method, {significantly enhancing performance on out-of-distribution tasks while demonstrating robust generalization capabilities}. {Consequently, \nickname achieves a 67\% success rate on even unseen tasks, representing improvements of over 54\% relative to learning-based methods and 32\% compared to LLM-based approaches.}
            \textcolor{\mycolor}{These findings provide compelling evidence of \nickname's ability to generalize acquired skills to novel tasks.}
        

\begin{table*}[ht]
\caption{\textcolor{black}{Success rates for high-precision tasks with unseen environments. "Obs-act", "Template", and "Video" indicate paired observation-action sequences, code templates, and videos performing subtasks. \textcolor[RGB]{1, 122, 177}{$\bullet$}  denote the utilization of \nameref{Sec::High}.}}
\label{table::HiP_Un}
\begin{minipage}{\textwidth}
\small

\makeatletter\def\@captype{table}

\makeatletter\def\@captype{table}
\begin{subtable}[t]{\textwidth}
\centering
\resizebox{\linewidth}{!}{
    {\fontsize{8}{10}\selectfont
        \begin{tabular}{@{}lccccccccc@{}}
                \toprule
                Methods & \thead{Type of\\demos } & \thead{Num of\\demos} & Triangle & Pear &
                Heart & 
                Pentagon & \thead{Double \\ Square} & \thead{Triple \\ Prong} & Overall \\
                \midrule 
                R3M-DP & Obs-act & 100 &  0.30 &  {0.30} &  {0.20} &  {0.20} &  {0.30} & 0.20 & $0.25 (\pm 0.05)$ \\
                DP & Obs-act & 100  & 0.20 & 0.10 & 0.20 & 0.30 & 0.10 & 0.20 & $0.19 (\pm 0.07)$ \\
                CaP& Template & 5  &  0.20 & 0.20 & 0.20 & 0.10 & 0.00  & 0.00 & $0.12 (\pm 0.09)$ \\
                Demo2Code& Video & 5  &  0.10 & 0.20 & 0.00 & 0.30 & 0.10  & 0.00 & $0.12 (\pm 0.12)$ \\
                \rowcolor{gray!25} \textbf{Ours$_{5v}$} & \textbf{Video} & \textbf{5} & \textbf{0.50}  & \textbf{0.50} & \textbf{0.40} & \textbf{0.50} & \textbf{0.30} & \textbf{0.50} & $\bm{0.45 (\pm 0.08)}$  \\
                \rowcolor{gray!25} \textbf{Ours$_{5v}$} \textcolor[RGB]{1, 122, 177}{$\bullet$}& \textbf{Video} & \textbf{5}  & \textbf{0.90} & \textbf{0.80} & \textbf{0.70} & \textbf{0.80} & \textbf{0.60} & \textbf{0.60} & $\bm{0.73 (\pm 0.12)}$  \\
                \bottomrule
            \end{tabular}
        }
    }
\end{subtable}
\end{minipage}
\end{table*}

                \subsection{Real-world Long-Horizon Tasks}\label{Sec::Long-Horizon Tasks}
                \subsubsection{Experimental setup} {Comparison methods are evaluated on a selected set of 6 long-horizon tasks with different goals across three environments: kitchen (3 tasks), desktop (2 tasks), and chemical laboratory (1 task).} These tasks constitute a representative benchmark of real-world challenges, with success criteria human-evaluated and designed to align with those in the original papers. {Since baseline methods encounter difficulties in completing long-horizon tasks in the UE setting, experiments are conducted in the SE setting. All other experimental settings are consistent with those in the subgoal manipulation task experiments}.
                
                \subsubsection{Results} {The performance of \nickname on long-horizon tasks is quantitatively assessed through its successful execution of six distinct tasks, each comprising at least seven subtasks}, across three diverse environmental contexts: kitchen, desktop, and chemical laboratory. 
                    The empirical results, as illustrated in Table \ref{Long}, demonstrate a statistically significant improvement achieved by our method in comparison to baseline approaches. These findings strongly indicate that the proposed \nickname is capable of developing robust and transferable skills, thereby achieving superior performance even in complex, long-horizon tasks. \textcolor{\mycolor}{Learning-based methods for long-horizon tasks often rely on direct mappings from visual observations to actions and lack intermediate guidance, which leads to a pronounced decline in success rates as task duration increases.} Our approach effectively leverages human demonstration videos to learn high-level plans, strategically segmenting the video into multiple subgoals. This segmentation enables the sequential execution of long-horizon tasks through a series of discrete subtask skills, {substantially  enhancing the overall success rate of long-horizon task completion.
                    While approaches such as CaP and Demo2Code similarly harness the reasoning capabilities of LLMs and VLMs for planning in long-horizon tasks,} they exhibit relatively low success rates in subtask completion. This limitation leads to challenges in robustly executing long-horizon tasks entirely.  \textcolor{\mycolor}{Our method overcomes these challenges by acquiring fine-grained action representations alongside high-level plan knowledge, offering a more robust and effective solution to the demands of long-horizon task execution.}
                {For additional details regarding long-horizon task designs, please refer to the Appendix.}

        \begin{table*}[t!]
            \caption{\textcolor{black}{Processing time of individual modules and sub-modules within \nickname, utilizing video sequences with a mean duration of 9 seconds. Computational efficiency can be enhanced through the implementation of multi-threading techniques and frame rate reduction.}}
            \centering
            \label{effieciency}
            \resizebox{1.0\linewidth}{!}{
                \footnotesize
                \begin{tabular}{>{\centering\arraybackslash}m{2.60cm}>{\centering\arraybackslash}m{3.60cm}>{\centering\arraybackslash}m{2.60cm}>{\centering\arraybackslash}m{2.60cm}>{\centering\arraybackslash}m{2.60cm}}
                    \toprule
                    \textbf{Module}                                                                    & \textbf{Sub-module}                    & \textbf{Sub-module duration} & \textbf{Module duration} & \textbf{Accelerated duration (single GPU)} \\
                    \midrule
                    \multirow{4}{*}{\begin{tabular}[c]{@{}c@{}}Interaction\\ Grounding\end{tabular}} & Task recognition         & 41s                            & \multirow{4}{*}{2.2min}  & \multirow{4}{*}{5.8 s}     \\ 
                    & Video parsing         & 14s                         &                          \\
                    & Subtask recognition       & 44s                         &                          \\
                    & Interaction estimation   & 32s                        &                          \\ \midrule
                    \multirow{4}{*}{Skill Learner}       
                    & Keypoint-waypoint extraction & 10s \\
                    & Grasping constraints      & 56s                            & \multirow{3}{*}{2.3min}   & \multirow{3}{*}{5.3 s}       \\ 
                    & Manipulation constraints     & 69s                            &                          \\ 
                    & Knowledge bank construction     & 3s                            &                          \\ \midrule
                    \multirow{4}{*}{Skill Adapter}                                                   & High-level planning                & 94s                            &\multirow{4}{*}{5.3min}  &\multirow{4}{*}{33s}       \\
                    & Keypoint transfer                & 36s                            & \\
                    
                    & Iterative comparison            & 130s                            &                          \\
                    & Fine-grained action correction            & 57s                            &                          \\
                \bottomrule
            \end{tabular}
        }
    \end{table*}
                
                \subsection{Real-world High Precision Tasks} \label{Sec::Real-world High Precision Tasks}
                \subsubsection{Experimental setup}  
\begin{figure}[t!]
                    \setlength{\abovecaptionskip}{-0.13cm}
                    \begin{center}
                        \includegraphics[width=0.47\textwidth]{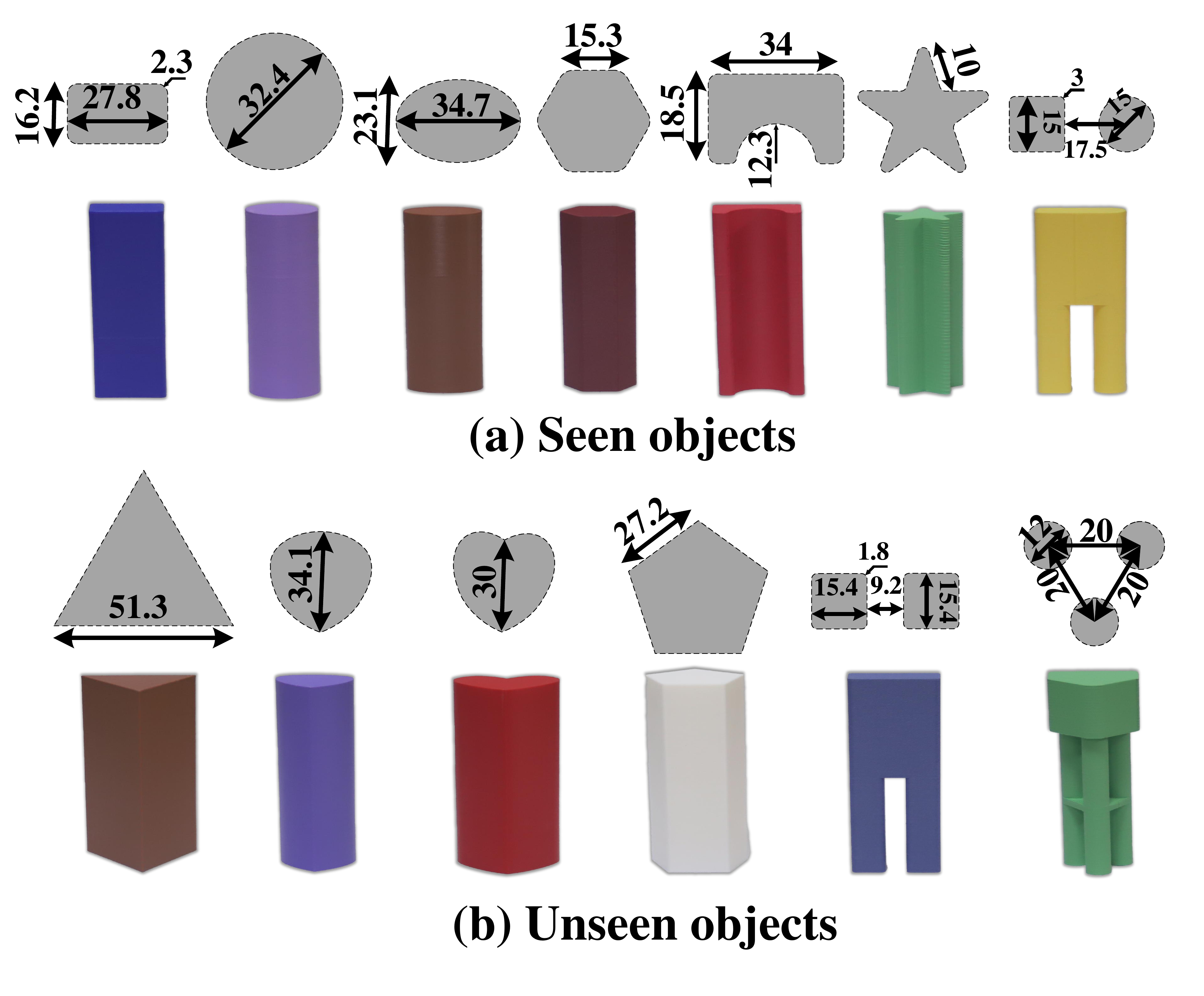}
                    \end{center}
                    \caption{\textcolor{black}{Objects used in high-precision tasks across (a) seen environments and (b) unseen environments.}  }
                    \label{Insert_objects}
                \end{figure}

{
                    To thoroughly assess our performance in high-precision manipulation tasks, experiments are conducted in both seen environment (SE) and unseen environment (UE) settings as shown in Figure \ref{Insert_objects}. The success criteria for these tasks are evaluated by human assessors, with the success rate calculated  based on 10 randomized object positions and orientations. All other settings remain the same as in the real-world manipulation task.}

                \subsubsection{Seen environment task results} \textcolor{\mycolor}{The Functional Manipulation Benchmark (FMB) \citep{luo2024fmb} offers a comprehensive framework for evaluating robotic manipulation capabilities}, encompassing steps such as grasping, reorientation, and assembly of various 3D-printed objects. Our evaluation protocol in the seen environment setting focuses on a subset of peg insertion tasks, specifically targeting seven distinct object geometries. {We provide demonstrations for each object to facilitate policy learning, and then deploy the learned policy for testing on the same object.} The experimental results, as shown in Table \ref{table::HiP}, demonstrate that our approach exhibits competitive performance relative to learning-based methods and substantially surpasses LLM-based approaches, even when relying exclusively on visual perception. {Furthermore, the integration of the skill refiner module into \nickname yields a statistically significant performance enhancement, surpassing all comparative methods.} \textcolor{\mycolor}{These findings collectively indicate that \nickname exhibits an impressive ability to learn and perform highly constrained manipulation tasks, despite utilizing only a limited set of human demonstration videos. Moreover, our method demonstrates robust task completion across a variety of scenarios.}

\begin{table*}[t]
\caption{\textcolor{black}{Robustness against viewpoint variance.} }
\label{viewpoint}
\resizebox{\linewidth}{!}{
{\footnotesize

\begin{tabular}{>{\centering\arraybackslash}m{2.5cm}  *{4}{>{\centering\arraybackslash}m{2.5cm}}}
\toprule
Methods & Viewpoint 1 & Viewpoint 2 & Viewpoint 3 & Viewpoint 4 \\ \hline
Ours & $0.75(\pm0.12)$ & $0.70 (\pm 0.14)$ & $0.73 (\pm 0.13)$ & $0.69 (\pm 0.15)$ \\
\bottomrule
\end{tabular}
}
}
\end{table*}

\begin{figure*}[th!]
    \setlength{\abovecaptionskip}{-0.13cm}
    \begin{center}
        \includegraphics[width=0.98\textwidth]{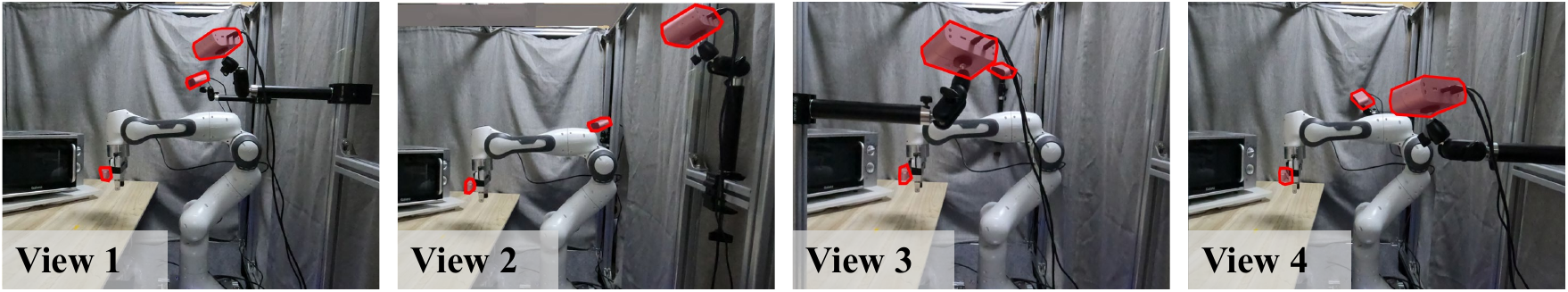}
    \end{center}
    \caption{\textcolor{black}{Configuration of various viewpoints.}  }
    \label{fig::viewpoint}
\end{figure*}
                
                \subsubsection{Unseen environment task results} {
                    To assess the generalization capabilities of our method in high-precision tasks, we conduct a series of experiments involving previously unobserved objects.
                    The policies learned in the seen environments are tested in the unseen environments.
                    DP and R3M-DP employ models trained exclusively on seen environment tasks for evaluation purposes, while CaP and Demo2Code leverage demonstrations provided within these familiar environments}. Our method develops a knowledge base derived from the skill acquisition process in seen environment tasks. 
                    {These unseen environment scenarios enable the assessment of the method's adaptability to previously unencountered objects,} providing insights into its robustness and flexibility in novel scenarios.
                    {The results, as shown in Table \ref{table::HiP_Un}, reveal significant limitations in the generalization capabilities of learning-based approaches when confronted with novel objects. Specifically, DP and R3M-DP exhibit suboptimal performance, completing tasks with just 25\% and 19\% success rates respectively.} Similarly, approaches employing VLMs or LLMs for planning also exhibit inadequate performance in high-precision tasks, attributable to their lack of fine-grained action adaptation ability, resulting in a mere 12\% success rate.
                    {Our proposed approach demonstrates superior generalization, maintaining comparatively high success rates even when presented with previously unseen environments.} Our method, relying exclusively on visual perception, demonstrates a 45\% success rate. \textcolor{\mycolor}{Moreover, the integration of the skill refiner module significantly enhances performance, elevating the success rate to 73\% and outperforming comparable  methods by at least 48\%.} \textcolor{\mycolor}{These empirical results compellingly demonstrate the superior generalization and adaptability of our approach in high-precision manipulation tasks, even when faced with novel objects and scenarios.}

\begin{table*}[t]
\caption{\textcolor{black}{Robustness against cumulative errors.} }
\label{Cumulative}
\resizebox{\linewidth}{!}{
    {\footnotesize
        
        \begin{tabular}{>{\centering\arraybackslash}m{2.5cm}  *{3}{>{\centering\arraybackslash}m{3.5cm}}}
            \toprule
            \nickname & Interaction grounding errors & Knowledge retrieval errors & Execution errors \\ \hline
            $0.83 (\pm 0.12)$ & $0.74 (\pm 0.17)$ & $0.70 (\pm 0.16)$ & $0.79 (\pm 0.14)$ \\
            \bottomrule
        \end{tabular}
    }
}
\end{table*}

\subsection{Real-world Practical High Precision Tasks}
\begin{figure*}[h]
    \setlength{\abovecaptionskip}{-0.13cm}
    \begin{center}
        \includegraphics[width=0.98\textwidth]{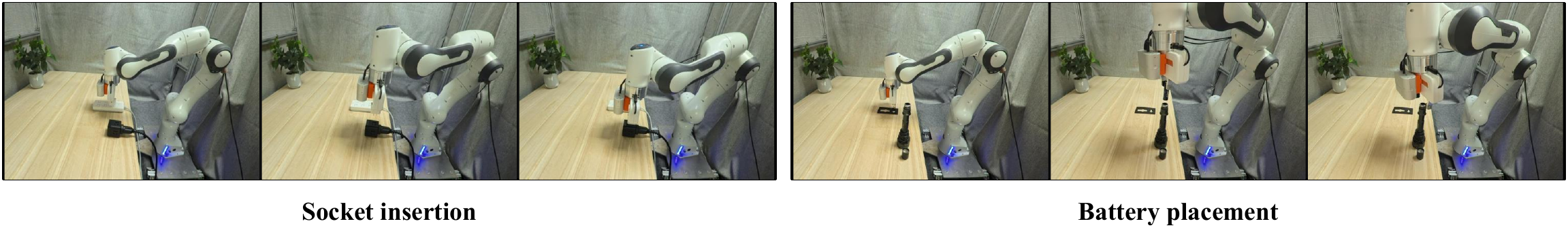}
    \end{center}
    \caption{{\textcolor{\mycolor}{Example qualitative results for practical high-precision tasks}.}  }
    \label{Fig:Pratical_High_precision}
\end{figure*}

\textcolor{\mycolor}{To validate the broad applicability of our approach to high-precision manipulation tasks, we conduct performance evaluations on representative practical scenarios, specifically socket insertion and battery placement tasks. The experimental configuration remains consistent with the high-precision task protocols (Sec. \nameref{Sec::Real-world High Precision Tasks}). Qualitative assessments are visualized in Figure \ref{Fig:Pratical_High_precision}, while quantitative performance metrics are summarized in Table \ref{Pratical_high_precision}. The experimental evidence demonstrates that our method sustains promising success rates across practical high-precision tasks, thereby corroborating the universal applicability of our proposed \nickname.}

\begin{table}[t]
\caption{\textcolor{\mycolor}{Success rates for practical high-precision tasks.} }
\label{Pratical_high_precision}
\resizebox{\linewidth}{!}{
    {\footnotesize
        
        \begin{tabular}{>{\centering\arraybackslash}m{2.9cm}  *{1}{>{\centering\arraybackslash}m{2.9cm}}}
            \toprule
            Socket insertion & Battery placement \\ \hline
            $0.70$ & $0.80$  \\
            \bottomrule
        \end{tabular}
    }
}
\end{table}

\subsection{Visualization of Skill Learning}
\begin{figure*}[h]
    \setlength{\abovecaptionskip}{-0.13cm}
    \begin{center}
        \includegraphics[width=0.88\textwidth]{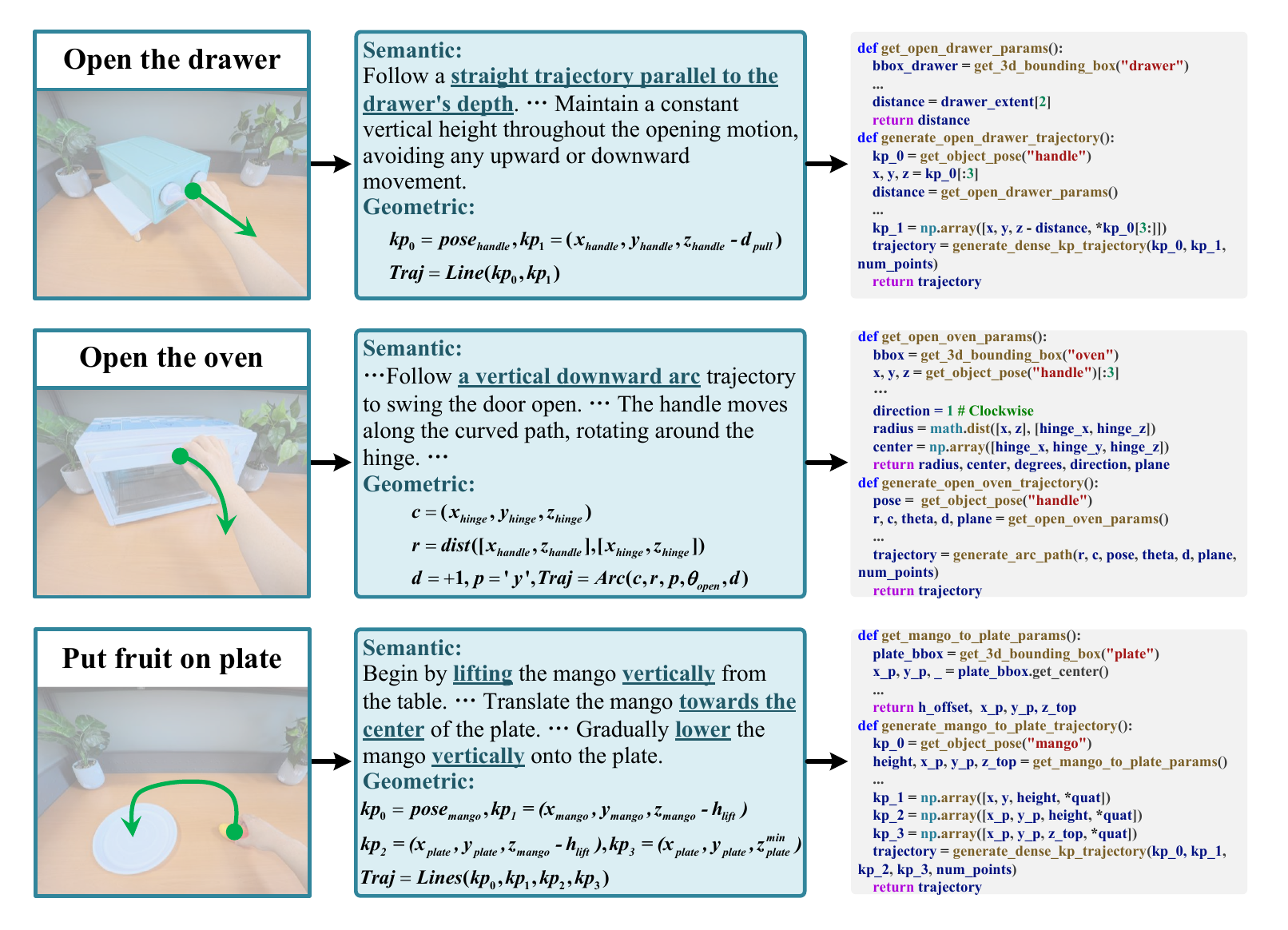}
    \end{center}
    \caption{{\textcolor{\mycolor}{Visualization of skill learning in our \nickname. Our FMimic can generate corresponding functions for different tasks}.}  }
    \label{Functions_examples}
\end{figure*}

\textcolor{\mycolor}{To provide empirical evidence of FMimic's skill learning proficiency, we showcase three representative skill acquisition scenarios, as visualized in the Figure \ref{Functions_examples}. These demonstrative cases establish that FMimic possesses the capability to systematically deduce task-specific semantic and geometric constraints, subsequently facilitating the generation of corresponding executable skill code. Additionally, the synthesized skill code encompasses parameter estimation methodologies and trajectory generation algorithms, thereby promoting skill reusability across diverse task domains.}

                        \subsection{Efficiency Evaluation} \label{Efficiency}
                        {The computational efficiency of each module and sub-module is assessed through temporal profiling on a workstation configured with an Intel I7-10700 CPU and an Nvidia RTX 3090 GPU}. The efficiency of our proposed framework is empirically validated through a comprehensive analysis of processing times across a diverse array of real-world manipulation tasks. This evaluation utilizes video sequences with an average duration of 9 seconds, captured at a frame rate of 30 Hz. \textcolor{\mycolor}{Experimental results, as illustrated in Table \ref{effieciency},  indicate that our method is capable of acquiring robust robotic skills within just 5 minutes}, and adapt the acquired skills to novel scenarios in less than 6 minutes. These findings underscore the potential of our \nickname for rapid skill acquisition and transfer. 
                        
                        {The computational efficiency of our \nickname can be further optimized through the implementation of the following simple but efficient strategies}, without compromising the effectiveness of skill acquisition and adaptation:
                            (\uppercase\expandafter{\romannumeral1}) The VLMs are accessed via online APIs, obviating the need for local computational resources. We utilize multi-threading to process subsequent videos concurrently with VLM operations. Consequently, we deliver a significant reduction in processing time. \textcolor{\mycolor}{The duration of the learning phase, encompassing interaction grounding and skill acquisition, is reduced from $269 s$ to $67 s$, Concurrently}, the adaptation phase exhibits a contraction from $317 s$ to $93 s$.
                            (\uppercase\expandafter{\romannumeral2}) {Video redundancy is mitigated through frame rate reduction, specifically from 30 frames per second (fps) to 5 fps. This optimization results in a substantial decrease of processing time}, with the learning phase duration further decreasing from $67s$ to $11s$, and the adaptation phase from $93s$ to $33s$.
                            (\uppercase\expandafter{\romannumeral3}) {The implementation of distributed processing methods}, wherein distinct video sequences are allocated to separate GPUs. \textcolor{\mycolor}{This approach yields significant  acceleration.} The concurrent utilization of 10 GPUs achieves a 9-fold acceleration in the learning phase and a 5-fold acceleration in the adaptation phase. {Consequently, the average learning time is reduced to $1.2s$, while the adaptation time decreases from $33s$ to $6.2s$.}
                        
                        
\subsection{Robustness against viewpoint variance} \label{Sec::viewpoint}
{The keypoint-centric representation approach enables our method to accommodate different observational perspectives}. To demonstrate the robustness of our method to varying viewpoints. Experiments are conducted in real-world unseen environments, utilizing distinct viewpoints, as shown in Figure \ref{fig::viewpoint}, where the first angle serves as the default perspective used in our experiments. \textcolor{\mycolor}{Experimental results prove that our method exhibits only a 6\% fluctuation in performance under varying viewpoints, highlighting the resilience of \nickname to viewpoint changes.}

\begin{figure}[t!]
    \setlength{\abovecaptionskip}{-0.13cm}
    \begin{center}
        \includegraphics[width=0.48\textwidth]{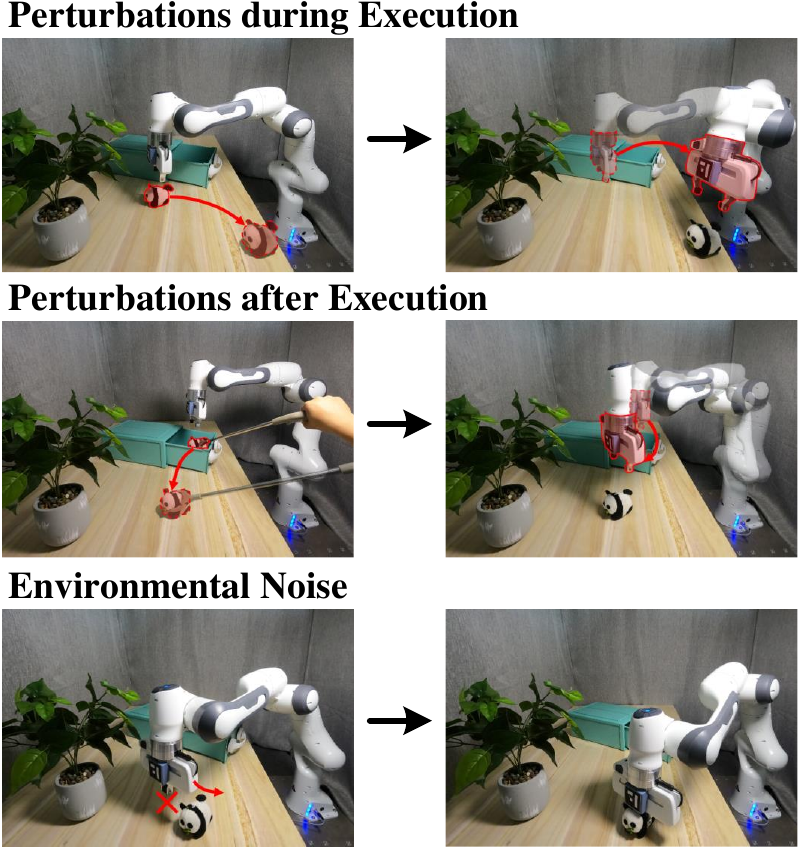}
    \end{center}
    \caption{\textcolor{black}{Illustration of robustness against visual and physical perturbations.}  }
    \label{robustness_fig}
\end{figure}


\begin{figure*}[t!]
    \setlength{\abovecaptionskip}{-0.13cm}
    \begin{center}
        \includegraphics[width=0.98\textwidth]{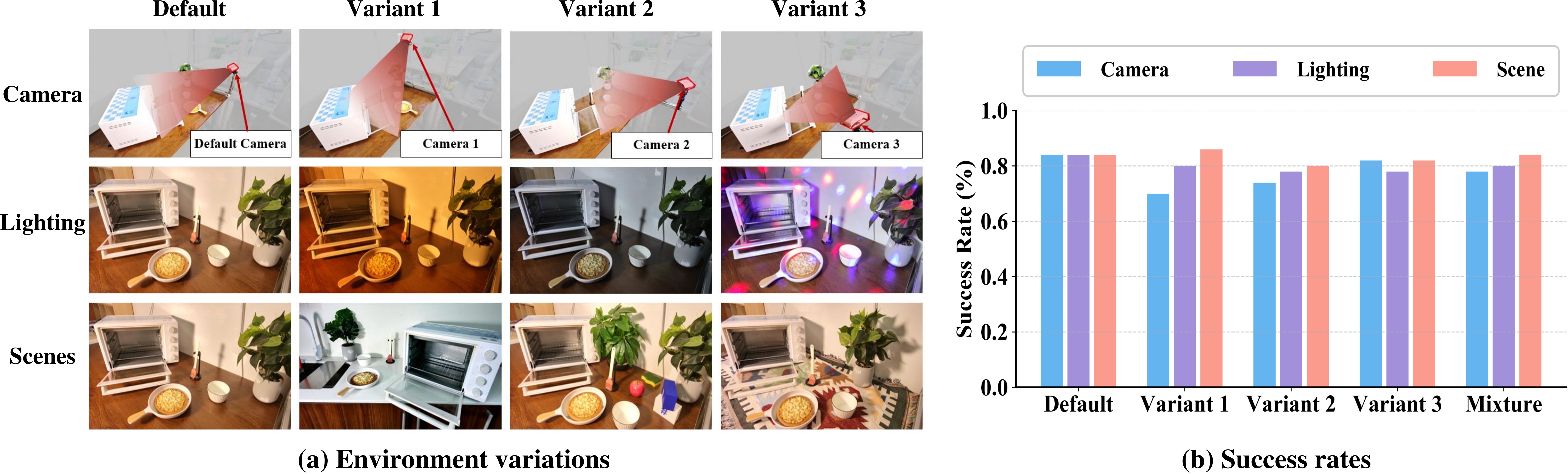}
    \end{center}
    \caption{{Environmental variations in terms of camera viewpoints, lighting, and scenes. Our \nickname exhibits strong robustness against these variations.}  }
    \label{Diversity}
\end{figure*}

\begin{table*}[ht]
\caption{{Success rates on real-world manipulation experiments with different visibility ratios}.}
\label{table::Visibility}
\resizebox{\linewidth}{!}{
{\footnotesize
\begin{tabular}{>{\centering\arraybackslash}m{3cm}  *{6}{>{\centering\arraybackslash}m{1.8cm}}*{1}{>{\centering\arraybackslash}m{1.8cm}}}
\toprule
 \thead{  Visibility \\ ratio} &\thead{  Stack \\ block} & \thead{ Put fruit \\ on plate} & \thead{ Put tray \\ in oven} & \thead{ Sweep \\ table} & \thead{ Sauce \\ spread} & \thead{Overall } \\
\midrule
1.0 & {0.90} & {0.90} & {0.80} & {0.90} & {0.70} & $\bm{0.84(\pm0.07)}$   \\
 0.8 & {0.90} & {0.80} & {0.80} & {0.80} & {0.60} & $\bm{0.78(\pm0.08)}$   \\
 0.6 & {0.90} & {0.70} & {0.70} & {0.60} & {0.60} & $\bm{0.70(\pm0.05)}$   \\
 0.4 & {0.70} & {0.60} & {0.50} & {0.50} & {0.50} & $\bm{0.56(\pm0.04)}$   \\
 0.2 & {0.50} & {0.40} & {0.30} & {0.30} & {0.30} & $\bm{0.36(\pm0.04)}$   \\
\bottomrule
\end{tabular}
}
}
\end{table*}

\subsection{Robustness against cumulative errors}\label{Sec::cumulative errors}
\textcolor{black}{Our \nickname framework exhibits robust performance in the presence of cumulative errors. \textcolor{\mycolor} We introduce perturbations at critical junctures of the skill acquisition and execution pipeline, specifically targeting the input stages of skill learning, adaptation, and execution. These perturbations correspond to interaction grounding errors, knowledge retrieval errors, and execution errors, respectively:}

\textcolor{black}{(\uppercase\expandafter{\romannumeral1}) Interaction grounding errors. \textcolor{\mycolor}{To simulate real-world uncertainty, Gaussian noise is intentionally applied to pose estimation results obtained from human demonstration videos}. The noise parameters are set at $\sigma = 5cm$ for positional and $\sigma = 5^{\circ}$ for rotational components, in alignment with the evaluation metrics employed in the Megapose study \citep{labbe2022megapose}. These metrics quantify prediction accuracy based on the percentage of estimates that fall within $5^{\circ}$ rotational and $5cm$ translational thresholds from the ground truth.}

\textcolor{black}{(\uppercase\expandafter{\romannumeral2}) Knowledge retrieval errors. A knowledge base is constructed utilizing experimental learning data from the RLBench environment. During the testing phase, relevant knowledge is retrieved from this corpus. \textcolor{\mycolor}{To emulate retrieval errors, we deliberately exclude the most similar entries from the set of 12 potential matches, thereby inducing suboptimal knowledge selection}.}

(\uppercase\expandafter{\romannumeral3}) Execution errors.  Gaussian noise is introduced to object pose estimation results during the execution phase. {The noise parameters remain consistent with those applied in the interaction grounding phase, with $\sigma = 5cm$ for positional and $\sigma = 5^{\circ}$ for rotational components. }

\textcolor{\mycolor}{Empirical evaluation through systematic error injection enables a comprehensive assessment of the system’s resilience to cumulative errors across different stages of skill acquisition and execution}. Experimental results are provided in Table \ref{Cumulative}. The proposed \nickname demonstrates remarkable resilience in the presence of interaction grounding errors, maintaining a robust success rate of 74\%. Our hierarchical constraint representation employs semantic-geometric constraints, enabling our method to leverage inferred semantic constraints to rectify geometric constraints, thereby mitigating the effects of interaction grounding errors.
    {Despite the introduction of knowledge retrieval errors, which cause the most substantial performance degradation, \nickname still maintains a commendable success rate of 70\%, demonstrating that our skill adapter module is able to dynamically update retrieved knowledge to align with the specific requirements of the current task.}
    Additionally, our \nickname exhibits remarkable resilience to execution errors, with their introduction resulting in a mere 4\% decrease in success rate. This minimal performance degradation strongly indicates that our fine-grained action correction mechanism possesses the capability to accurately identify and rectify execution errors, thereby facilitating successful task completion.
    \textcolor{\mycolor}{These empirical findings collectively underscore the robust performance of \nickname\ across a range of error conditions, highlighting its potential for reliable operation in complex scenarios characterized by uncertainty and error accumulation.}

\begin{figure*}[th!]
\setlength{\abovecaptionskip}{-0.13cm}
\begin{center}
\includegraphics[width=0.98\textwidth]{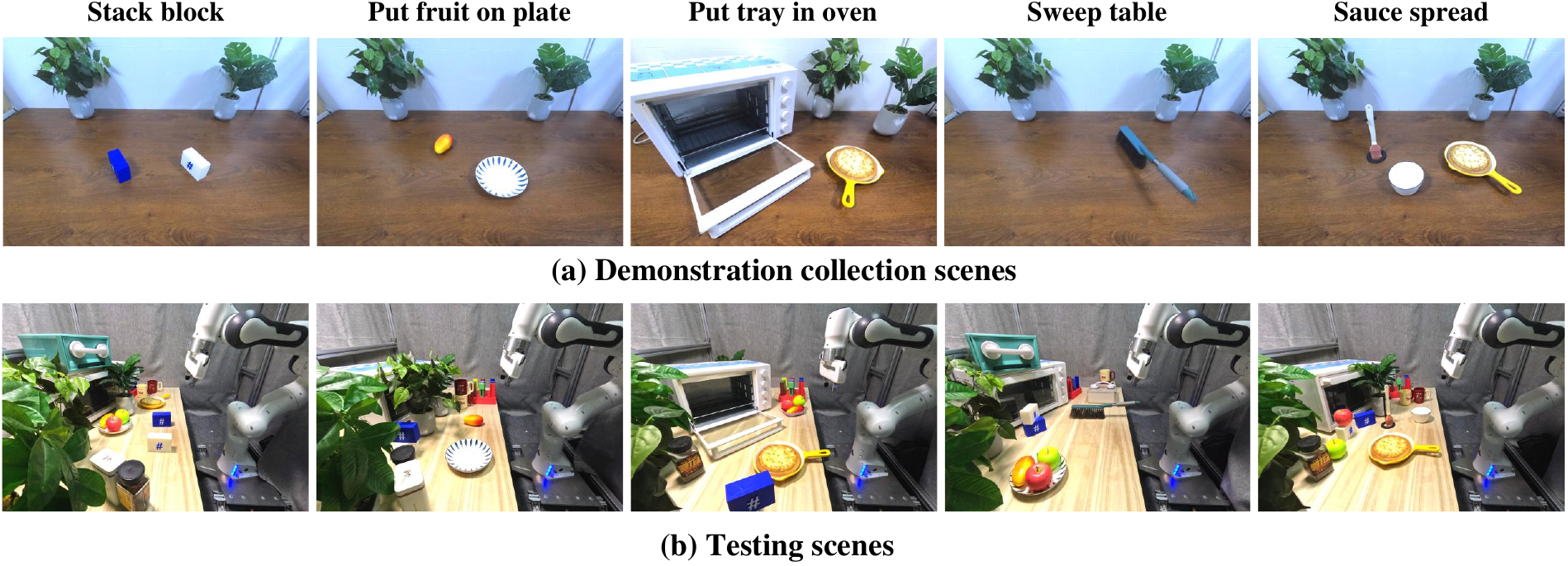}
\end{center}
\caption{{Ablation Study on object count and spatial distribution variations.}  }
\label{objects}
\end{figure*}

\begin{table*}[t]
\caption{{Success rates on environments with different object counts and spatial distribution. $vo$ denotes the variants evaluated in environments with different object quantities and spatial distribution.}}
\label{table::object_number}
\resizebox{\linewidth}{!}{
{\footnotesize
\begin{tabular}{>{\centering\arraybackslash}m{2cm}  *{6}{>{\centering\arraybackslash}m{1.8cm}}*{1}{>{\centering\arraybackslash}m{1.8cm}}}
\toprule
Methods &\thead{  Stack \\ block} & \thead{ Put fruit \\ on plate} & \thead{ Put tray \\ in oven} & \thead{ Sweep \\ table} & \thead{ Sauce \\ spread} & \thead{Overall } \\
\midrule
\textbf{Ours$\rm_{vo}$}& \textbf{0.90} & \textbf{0.80} & \textbf{0.80} & \textbf{0.80} & \textbf{0.70} & $\bm{0.80(\pm0.06)}$   \\
\rowcolor{gray!25} \textbf{Ours}& \textbf{0.90} & \textbf{0.90} & \textbf{0.80} & \textbf{0.90} & \textbf{0.80} & $\bm{0.86(\pm0.04)}$   \\
\bottomrule
\end{tabular}
}
}
\end{table*}

\subsection{Robustness against Perturbation}  \label{Robustness}
\textcolor{black}{The robustness of our \nickname against visual and physical perturbations is assessed in a separate episode, exemplified by a pick-and-place task. As shown in Figure \ref{robustness_fig}, three types of perturbation are introduced to validate the robustness of our method. 
    (\uppercase\expandafter{\romannumeral1}) Position perturbations during execution. Objects are subjected to spatial displacement during the picking or placing phases. {Our \nickname approach demonstrates adaptive capability by modifying its trajectory to complete the task, thus validating the efficacy of our keypoint-centric representation in mitigating object position disturbances. }
    (\uppercase\expandafter{\romannumeral2}) Position perturbations after execution. Objects are repositioned while the robotic system is in transit to the designated end-stage following action completion. {Our \nickname exhibits the ability to re-adjust the object to its intended target before proceeding to the end-stage}, thereby substantiating the effectiveness of our method in failure detection and correction.
    (\uppercase\expandafter{\romannumeral3}) Environmental noise. Extrinsic parameter noise is deliberately introduced into the camera system. Despite the initial impediment to task execution caused by introduced perturbations, {experiments demonstrate that \nickname possesses the capability to perform fine-grained action correction and perceived pose optimization, ultimately facilitating successful task completion.
    These findings collectively demonstrate the robustness and adaptability of \nickname when confronted with diverse  perturbations}, encompassing both physical object displacements and environmental noise. The empirically demonstrated capacity of our \nickname to adapt to and overcome these challenges highlights its potential for deployment in real-world scenarios characterized by dynamic and unpredictable environments.
}
                        
\begin{table*}
\caption{Ablation experiments with \nickname on real-world manipulation experiments. "SE" and "UE" are seen and unseen environments. Default settings are marked in \colorbox{gray!40}{gray}.}
\begin{minipage}{\textwidth}
\makeatletter\def\@captype{table}
\begin{subtable}[t]{0.32\textwidth}
\centering
\caption{Hierarchical representations.}
\vspace{-1.5mm}
\begin{tabular}{>{\centering\arraybackslash}m{3.60cm} *{1}{>{\centering\arraybackslash}m{0.90cm}}}
    \toprule
    Variants     & SE       \\
    \midrule
    Geometric constraints & 0.67       \\
    Semantic constraints   & 0.74       \\
    \rowcolor{gray!40} Hierarchical constraints     & 0.84         \\
    \bottomrule
\end{tabular}
\end{subtable}
\makeatletter\def\@captype{table}
\begin{subtable}[t]{0.32\textwidth}
\centering
\caption{Grasping learning.}
\vspace{-1.5mm}
\begin{tabular}{>{\centering\arraybackslash}m{3.60cm} *{1}{>{\centering\arraybackslash}m{0.90cm}}}
    \toprule
    Variants     & SE       \\
    \midrule
    Value prediction   & 0.56       \\
    Grouping (DBSCAN) & 0.66       \\
    \rowcolor{gray!40} Grouping with VLMs     & 0.84         \\
    \bottomrule
\end{tabular}
\end{subtable}
\makeatletter\def\@captype{table}
\begin{subtable}[t]{0.32\textwidth}
\centering
\caption{Number of videos.}
\vspace{-1.5mm}
\begin{tabular}{>{\centering\arraybackslash}m{1.25cm} *{1}{>{\centering\arraybackslash}m{0.60cm}}| >{\centering\arraybackslash}m{1.25cm} *{1}{>{\centering\arraybackslash}m{0.60cm}}}
    \toprule
    Number     & SE & Number     & SE      \\
    \midrule
    1 &  0.73 & 7 &   0.85    \\
    3     &  0.81   & 9 &   0.85    \\
    \rowcolor{gray!40}  5     &  0.84 &  \cellcolor{white}11     &  \cellcolor{white}0.86       \\
    \bottomrule
\end{tabular}
\end{subtable}

\vspace{6pt}

\makeatletter\def\@captype{table}
\begin{subtable}[t]{0.32\textwidth}
\caption{Comparison strategy.}
\vspace{-1.5mm}
\centering
\begin{tabular}{>{\centering\arraybackslash}m{3.60cm} *{1}{>{\centering\arraybackslash}m{0.90cm}}}
    \toprule
    Variants     & UE       \\
    \midrule
    Visual comparison   & 0.65         \\
    Values comparison      & 0.63           \\
    \rowcolor{gray!40} Visual with values     & 0.75  \\
    \bottomrule
\end{tabular}
\end{subtable}
\makeatletter\def\@captype{table}
\begin{subtable}[t]{0.32\textwidth}
\centering
\caption{Number of iterations.}
\vspace{-1.5mm}
\begin{tabular}{>{\centering\arraybackslash}m{1.25cm} *{1}{>{\centering\arraybackslash}m{0.60cm}}| >{\centering\arraybackslash}m{1.25cm} *{1}{>{\centering\arraybackslash}m{0.60cm}}}
    \toprule
    Number     & UE & Number     & UE      \\
    \midrule
    0 &  0.62 & 3 &   0.72    \\
    \rowcolor{gray!40} \cellcolor{white}1     &  \cellcolor{white}0.67   & 4 &   0.75    \\
    2     &  0.70 &  5     &  0.75       \\
    \bottomrule
\end{tabular}
\end{subtable}
\makeatletter\def\@captype{table}
\begin{subtable}[t]{0.32\textwidth}
\centering
\caption{Fine-grained action correction.}
\vspace{-1.5mm}
\begin{tabular}{>{\centering\arraybackslash}m{1.25cm} *{1}{>{\centering\arraybackslash}m{0.60cm}}| >{\centering\arraybackslash}m{1.25cm} *{1}{>{\centering\arraybackslash}m{0.60cm}}}
    \toprule
    Number     & UE & Number     & UE      \\
    \midrule
    0 &  0.69 & 3 &   0.76    \\
    1     &  0.73   & 4 &   0.75    \\
    \rowcolor{gray!40}  2     &  0.75 &  \cellcolor{white}5     &  \cellcolor{white}0.76       \\
    \bottomrule
\end{tabular}
\end{subtable}
\end{minipage}
\label{Ablation}
\end{table*}

\subsection{Robustness against Object Visibility}  
To evaluate the robustness of our method with respect to trajectory visibility, we conduct experimental validation across five representative manipulation tasks: stack block, put fruit on plate, put tray in oven, sweep table, and sauce spread.  We investigate the impact of varying visibility proportions on performance by evaluating corresponding success metrics. As demonstrated in Table \ref{table::Visibility}, our approach exhibits robust performance even when objects are not continuously visible throughout the demonstration sequence. Notably, the method maintains a substantial success rate of 70\% even when the visibility ratio decreases to 60\%. \textcolor{\mycolor}{This represents only a 14\% reduction in performance compared to scenarios with complete object visibility, thereby validating our method's resilience to partial occlusion conditions.}

\subsection{Robustness against Video Diversity}  
To evaluate environmental robustness, we assess our method's performance across diverse demonstration videos, incorporating variations in camera perspectives, lighting conditions, and environmental settings. For each setting, we design three variants, as well as a mixed variant that randomly samples from the demonstration videos of these three variants. We conduct experimental validation across five representative manipulation tasks: stack block, put fruit on plate, put tray in oven, sweep table, and sauce spread. As illustrated in Figure \ref{Diversity}, our method demonstrates remarkable stability under different lighting and environmental conditions, with negligible performance degradation. While variations in camera perspective exhibited a more pronounced impact on performance metrics, our approach maintains substantial resilience across viewpoint changes. \textcolor{\mycolor}{The experimental results collectively confirm our method's consistent performance across diverse conditions, highlighting its adaptability and generalizability.}

\subsection{Performance across varied object counts and spatial arrangements} 
To demonstrate the generalization capability of our method across scenarios with varying object quantities and spatial distributions, we conduct experiments on five representative tasks: stack block, put fruit on plate, put tray in oven, sweep table, and sauce spread. Data are collected in the scenarios shown in Figure \ref{objects} (a) and evaluation is conducted in the scenarios depicted in Figure \ref{objects} (b).  \textcolor{\mycolor}{The experimental results presented in Table \ref{table::object_number} substantiate the efficacy of our  \nickname, which demonstrates both elevated success rates and robust performance, even in the presence of variations in object quantities and spatial distributions.}

\subsection{Robustness of identifying objects within cluttered environments}
\begin{figure*}[h!]
\setlength{\abovecaptionskip}{-0.13cm}
\begin{center}
\includegraphics[width=0.98\textwidth]{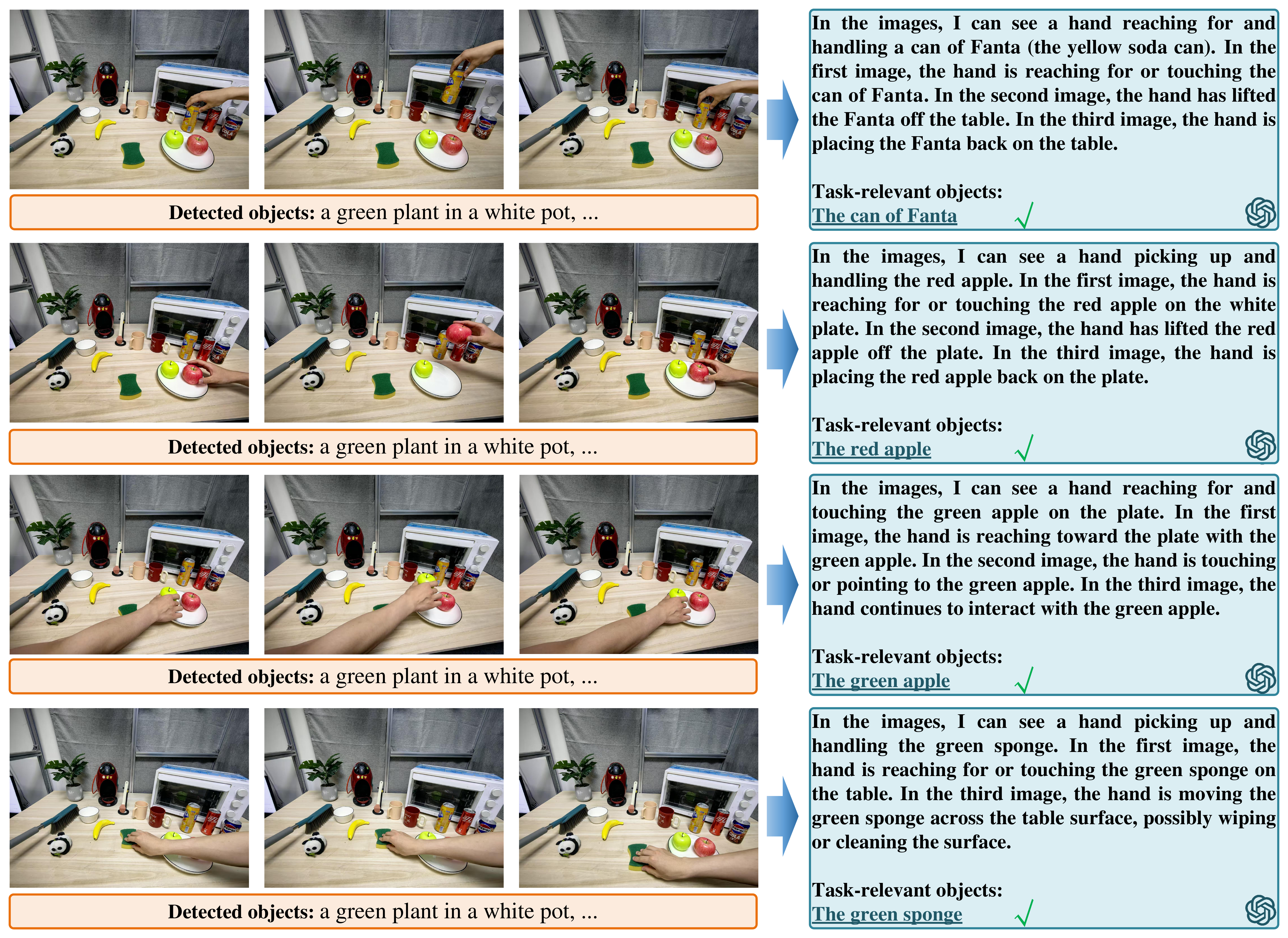}
\end{center}
\caption{\textcolor{\mycolor}{Demonstrations of task-relevant object reasoning in cluttered environments. We provide keyframes and their corresponding object detection results to VLMs, VLMs accurately reason about task-relevant objects, even in cluttered environments.}  }
\label{Fig::Tokenize_anything_VLMs}
\end{figure*}

\textcolor{\mycolor}{
To verify FMimic's ability to recognize task-relevant objects in cluttered environments, we performed experiments under cluttered environmental conditions.
Our FMimic, equipped with vision foundation models, generates detailed descriptions that distinguish between objects of the same category with different semantic meanings. FMimic transmits these processed textual descriptions to VLMs for analysis, effectively mitigating the recognition limitations that VLMs conventionally encounter in complex visual scenarios. Experimental results show that our FMimic accurately identifies task-relevant objects, even in cluttered environments, as demonstrated in Figure \ref{Fig::Tokenize_anything_VLMs}. Additionally, our \nickname effectively differentiates between objects of the same category that possess distinct semantic meanings.}

                        \subsection{Ablation Studies}\label{Sec::Ablation}
                        {Comprehensive ablation studies are performed to investigate the fundamental designs of our \nickname approach. The effects of these designs are evaluated by measuring the success rate on real-world manipulation tasks, which is computed across 10 randomized object positions and orientations.}
                        
                        \subsubsection{Hierarchical constraint representations} 
                        \textcolor{black}{An analysis of three distinct constraint representation approaches is presented in Table \ref{Ablation} (a). \textcolor{\mycolor}{Variants that rely solely on semantic constraints, or that directly extract geometric constraints without incorporating semantic analysis, demonstrate a marked decline in performance}. The results indicate that the proposed hierarchical constraint representations, which seamlessly integrate semantic and geometric constraints, significantly enhance skill acquisition capabilities. This finding underscores the pivotal role of such hierarchical representations in facilitating and augmenting the understanding and reasoning capabilities of VLMs in the context of robotic manipulation tasks.}
                        
                        \subsubsection{Grasping learning} \textcolor{\mycolor}{Table \ref{Ablation} (b) offers a comprehensive comparison of the performance across different variants. The first variant, which employs VLMs for direct prediction of constraint region values, exhibits a substantial decline in performance metrics.} This observation suggests the inadequacy of the direct prediction in capturing the nuanced complexities of constraint regions. The second variant implements the DBScan clustering algorithm to aggregate grasp poses and derive constraints as bounded regions. However, this approach exhibits limited efficacy due to its exclusive reliance on numerical distributions, neglecting the incorporation of grasping heuristics and domain-specific knowledge. 
                        {Empirical evaluation reveals that the proposed method achieves superior performance by effectively leveraging the inherent common sense knowledge and reasoning capabilities of VLMs.} Furthermore, we enhance the pattern reasoning abilities of these models through the introduction of hierarchical constraint representations. \textcolor{\mycolor}{The integration of these advanced approaches yields a substantial improvement in overall performance, emphasizing the efficacy of our novel method in addressing the complexities inherent in constraint region prediction.}

\begin{figure*}[th!]
\setlength{\abovecaptionskip}{-0.13cm}
\begin{center}
    \includegraphics[width=0.98\textwidth]{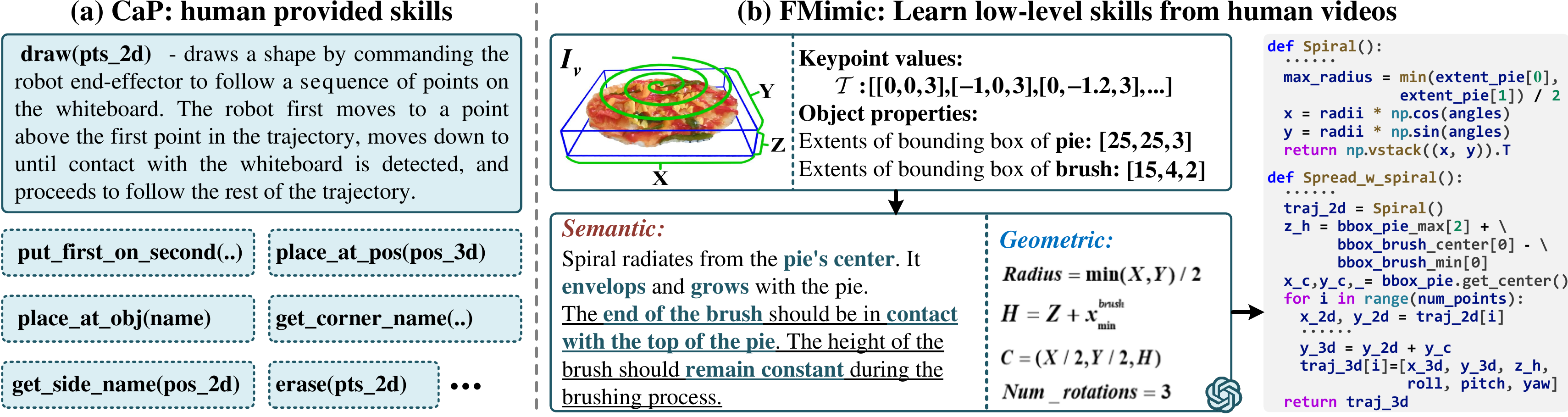}
\end{center}
\caption{\textcolor{black}{Comparative illustration of fine-grained motion skill acquisition between \nickname and planner-based approaches. (a) Methods utilize VLMs as planners rely on the provided motion primitives. (b) Our \nickname autonomously acquires fine-grained actions from human videos.}  }
\label{Comparison_with_Cap}
\end{figure*}
                        
                        \subsubsection{Number of human videos} Table \ref{Ablation}(c) presents a quantitative analysis of the relationship  between the number of human videos and system performance. The results demonstrate that our method achieves promising success rates on complex tasks even with a singular human video demonstration. \textcolor{\mycolor}{Furthermore, we observe a clear positive correlation between the number of demonstration videos and performance metrics.}
                        {These findings validate the efficacy of our approach in efficiently extracting and generalizing skills from a limited corpus of human demonstrations.} The results suggest a robust learning mechanism capable of distilling salient information from limited examples, thereby facilitating robust skill acquisition and transfer.
                        {Considering the trade-off between data availability and performance optimization, our empirical analysis suggests that a corpus of five demonstration videos provides an optimal balance}, \textcolor{\mycolor}{which facilitates sufficient data diversity to capture task variability while maintaining practical constraints imposed by data collection protocols and computational resource limitations.}
                        
                        \subsubsection{Comparison strategy} {Table \ref{Ablation} (d) presents a thorough analysis of the impact of the comparison strategy employed in skill adapters. Variants that rely exclusively on either visualized interactions or interaction values for constraint comparison exhibit a marked decrease in success rates}. The visual comparison approach facilitates semantic contrast within the VLMs, while interaction values provide fine-grained geometric information crucial for constraint representation.
                        \textcolor{\mycolor}{The experimental results demonstrate that our proposed strategy, which seamlessly integrates both visual and numerical comparisons, supports effective reasoning for the adaptation of both semantic and geometric constraints.} This integrated approach capitalizes on the complementary strengths of each comparison modality, facilitating a more holistic and nuanced understanding of the skill constraints.

                        \subsubsection{Number of iterations} {A systematic analysis is conducted to evaluate the impact of iteration count in the skill adapter and to determine the optimal parameter configuration, as illustrated in Table \ref{Ablation}(e)}. The results demonstrate that a reduction in the number of iterations to zero leads to a statistically significant decrease in performance metrics. Notably, robust performance is observed even with a single iteration, and subsequent iterations yield incremental improvements.
                        \textcolor{\mycolor}{These empirical findings indicate that the iterative approach substantially augments the effectiveness of skill adaptation, allowing VLMs to converge more efficiently on optimal solutions within the solution space and thereby enhancing both task performance and generalization}. Based on this evaluation of the performance-iteration relationship, the four-iteration configuration is determined to provide an optimal balance between computational efficiency and performance enhancement. {This configuration maximizes the benefits of the iterative approach while maintaining practical limitations on processing time and resource utilization.}
                        
                        

\begin{figure*}[t!]
    \setlength{\abovecaptionskip}{-0.13cm}
    \begin{center}
        \includegraphics[width=0.98\textwidth]{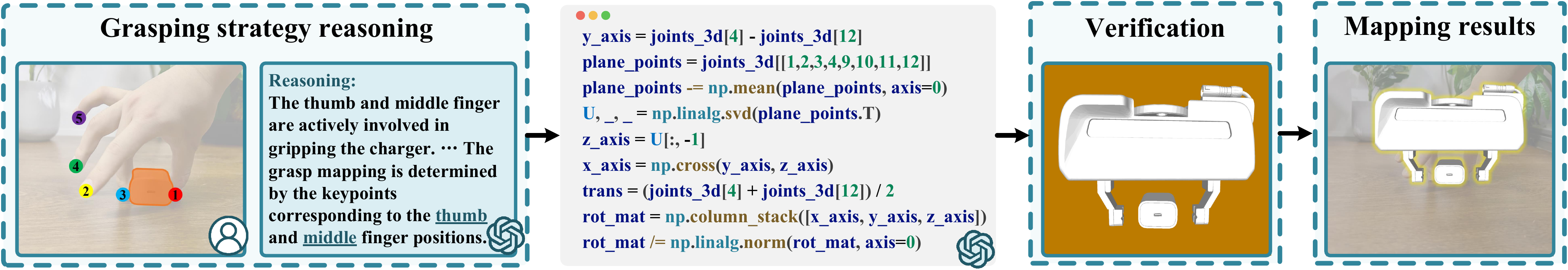}
    \end{center}
    \caption{{Illustration of enhanced grasping pose mapping method.}  }
    \label{Grapsing}
\end{figure*}

\begin{figure*}[t!]
    \setlength{\abovecaptionskip}{-0.13cm}
    \begin{center}
        \includegraphics[width=0.98\textwidth]{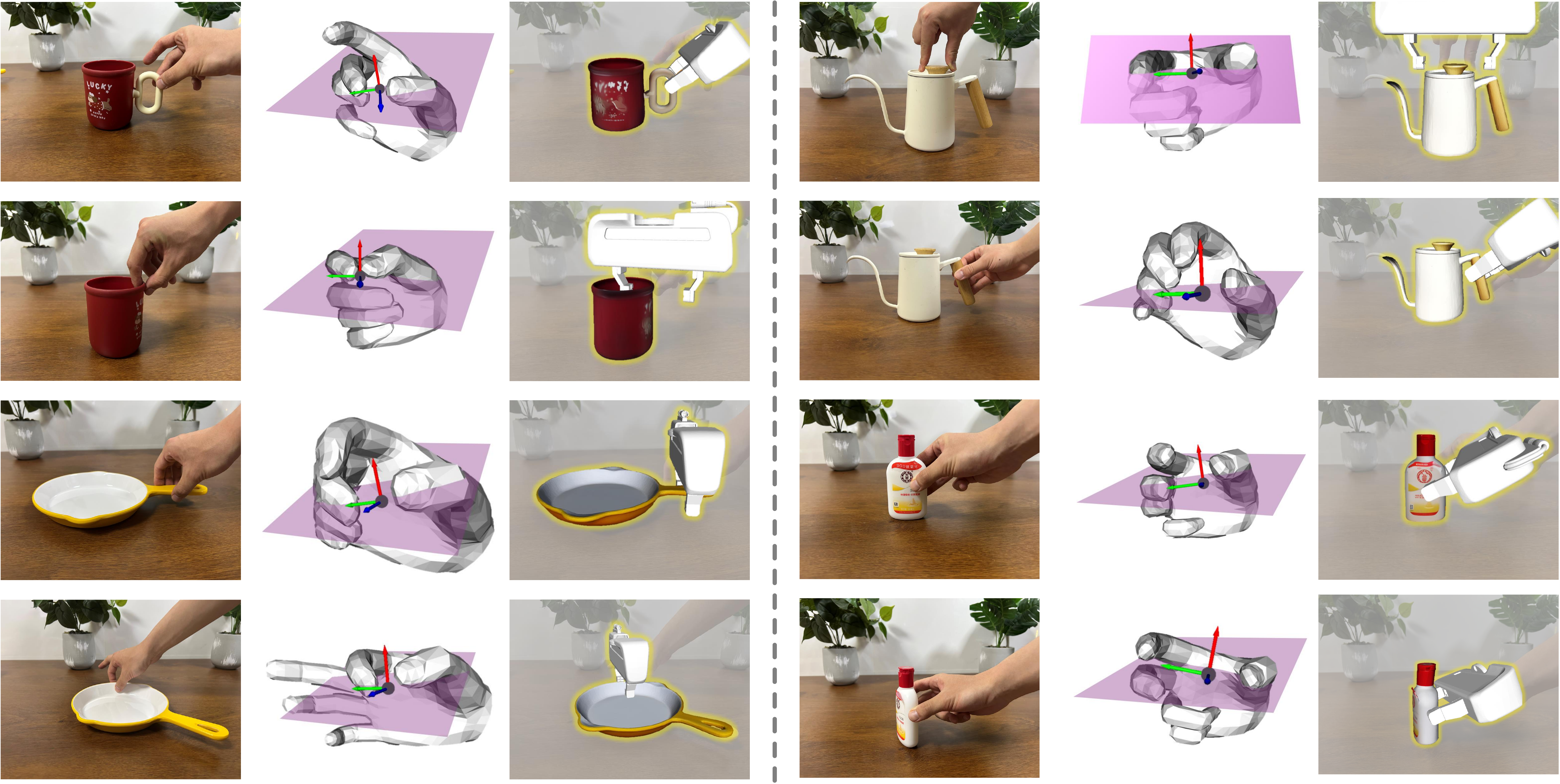}
    \end{center}
    \caption{{Examples of enhanced grasping pose mapping method. Our approach is applicable to diverse grasping situations.}  }
    \label{Grapsing_examples}
\end{figure*}
                        
                        \begin{figure*}[th!]
                            \setlength{\abovecaptionskip}{-0.13cm}
                            \begin{center}
                                \includegraphics[width=0.98\textwidth]{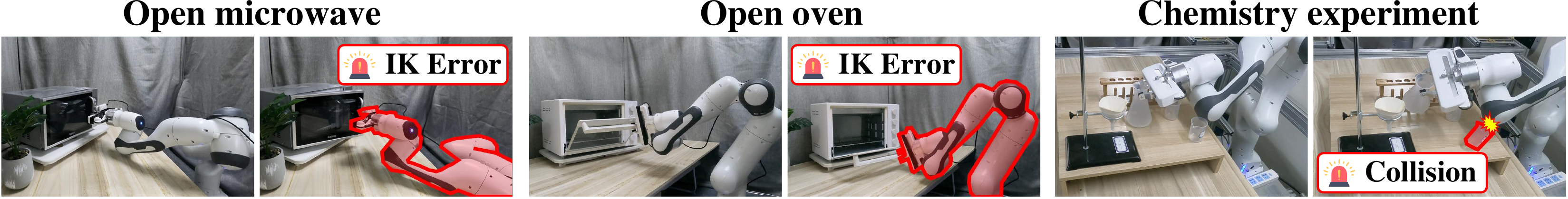}
                            \end{center}
                            \caption{\textcolor{black}{Examples of failure cases.}  }
                            \label{failure_case}
                        \end{figure*}

\subsubsection{Fine-grained action correction} \textcolor{black}The efficacy of fine-grained action correction is systematically evaluated, with results presented in Table \ref{Ablation}(f). Experimental results reveal a positive correlation between the number of iterations and the success rate, with an inflection point observed at two iterations. This inflection point represents an optimal equilibrium between real-time performance and task success rate, balancing computational efficiency with operational effectiveness.
{The implementation of fine-grained action correction proves particularly critical in tasks requiring high-precision manipulation, which are inherently susceptible to environmental perturbations and stochastic noise}. \textcolor{\mycolor}{The fine-grained action correction substantially enhances both the overall success rate and the robotic system's capacity to operate effectively in complex, dynamic environments, enabling it to consistently achieve high levels of precision and reliability across a wide spectrum of operational scenarios and task complexities.}

\begin{table*}[ht]
\caption{{Success rates on real-world manipulation experiments with TRELLIS method.}}
\label{table::TRELLIS}
\resizebox{\linewidth}{!}{
{\footnotesize
\begin{tabular}{>{\centering\arraybackslash}m{3cm}  *{6}{>{\centering\arraybackslash}m{1.8cm}}*{1}{>{\centering\arraybackslash}m{1.8cm}}}
\toprule
Methods &\thead{  Stack \\ block} & \thead{ Put fruit \\ on plate} & \thead{ Put tray \\ in oven} & \thead{ Sweep \\ table} & \thead{ Sauce \\ spread} & \thead{Overall } \\
\midrule
\textbf{Ours $\rm w/ \,TRELLIS$}& \textbf{0.70} & \textbf{0.80} & \textbf{0.80} & \textbf{0.80} & \textbf{0.70} & $\bm{0.76(\pm0.05)}$   \\
\rowcolor{gray!25} \textbf{Ours}& \textbf{0.80} & \textbf{0.90} & \textbf{0.70} & \textbf{0.90} & \textbf{0.60} & $\bm{0.78(\pm0.11)}$   \\
\bottomrule
\end{tabular}
}
}
\end{table*}

\subsubsection{3D model generation} Recent advancements in 3D AIGC, particularly exemplified by TRELLIS \citep{xiang2024structured}, have demonstrated promising outcomes in generating 3D object models from only sparse multi-view data or, notably, single-view inputs. \textcolor{\mycolor}{This substantial reduction in input requirements represents a significant advancement toward more accessible 3D model generation protocols}. We select five representative tasks and conduct experiments in real-world settings to evaluate our method's generalization capability to unseen environments using TRELLIS. \textcolor{\mycolor}{The experimental results, provided in Table \ref{table::TRELLIS}, corroborate that our \nickname maintains efficacious performance, using the AIGC method to generate 3D models.}

                        
                        \section{Discussion and Limitations} \label{Limitations}
                        
                        \subsection{Fundamental difference with planner methods}
                        While methods incorporating VLMs or LLMs as planners demonstrate the capability to execute similar tasks, such as the 'drawing' task in CaP and the 'spread sauce' task in our proposed method, {it is noteworthy that these approaches fail to acquire fine-grained motion skills.}
                        
                        As illustrated in Figure \ref{Comparison_with_Cap}, there exists a significant distinction in the fine-grained motion skill acquisition:
                            (a) In the drawing experiments of existing methods, the motion skill encompasses precise control to maintain continuous contact between the pen and the whiteboard surface, while ensuring adherence to the intended trajectory. This skill is implemented through a human-authored function, specifically, draw(pts\_2d).
                            (b) In comparison, our proposed \nickname framework demonstrates a marked advancement by enabling the autonomous acquisition of these fine-grained motion skills. {This capability represents a substantial improvement over existing methods, removing the need for pre-programmed, task-specific functions.}
                       \textcolor{\mycolor}{This distinction highlights the innovative character of our approach, which enables the emergence of sophisticated skills through learning paradigms, circumventing reliance on fixed, pre-specified behavioral patterns. Such capability profoundly enhances the adaptability and generalizability of robotic systems across diverse task environments.}
                        

\subsection{Hand grasp representation}

\textcolor{\mycolor}{To capture the rich variety of human grasping strategies, we attempt to develop an enhanced hand posture conversion framework to expand grasp representation, thereby accommodating a more diverse spectrum of human grasping techniques}. Our FMimic first identifies hand configurations and grasped objects, as delineated in our human-object interaction grounding module. Subsequently, we annotate each finger with a unique identifier within the image while simultaneously demarcating the grasped object using a mask representation, as illustrated in Figure \ref{Grapsing}. These annotated images are then processed by VLMs, which analyze the grasping strategy and generate appropriate hand-to-gripper mapping code. \textcolor{\mycolor}{To facilitate the code generation process, we provide the VLMs with our existing thumb-index finger mapping code, allowing the VLMs to focus on grasping strategy analysis.}

{To enhance the robustness of our approach, we present a verification module that evaluates mapping efficacy. Following the hand-to-gripper posture conversion, we project the resulting gripper configuration into 3D space in conjunction with the grasped object, utilizing the 3D models of the gripper and the grasped object. A suitable grasping posture should satisfy the condition that the object is positioned between the two gripper jaws while maintaining no contact between the object and the jaws, as illustrated in Figure \ref{Grapsing}. In the event of validation failure, the system re-queries the VLMs to generate an alternative mapping solution. When validation succeeds, the current mapping configuration is implemented. More examples are provided in Figure \ref{Grapsing_examples}.}
                        
\subsection{Real-world failure cases}
\textcolor{\mycolor}{Figure \ref{failure_case} depicts scenarios that present substantial challenges for resolution via VLM reasoning. These scenarios encompass:}
    (\uppercase\expandafter{\romannumeral1}) {The task execution may surpass the hardware limitations of the physical robot, inducing inverse kinematics (IK) errors. }
    (\uppercase\expandafter{\romannumeral2}) {Inadequate environmental perception heightens the likelihood of obstacle collisions, leading to task failure.} 
    Since the training datasets for VLMs exhibit a significant lack of data related to robot dynamics, these models lack associated knowledge, {exhibiting a restricted capacity for error analysis and encountering difficulties in inferring corrective strategies.} 

    {The current system relies on the point cloud data perceived by the main camera for obstacle avoidance, where the camera exhibiting superior perceptual range is designated as the primary sensing unit. \textcolor{\mycolor}{The acquired point cloud data, together with FMimic’s pose predictions, are provided as inputs to the Open Motion Planning Library (OMPL) \citep{sucan2012open} for trajectory generation.}
    To further enhance the robustness of our FMimic, we present two key improvements: (I) enhanced environmental perception through multi-camera reconstruction to achieve comprehensive spatial coverage, and (II) implementation of an advanced motion planning framework, such as CuRobo \citep{sundaralingam2023curobo}, to reduce IK failures and collision incidents.}
                        
                        \subsection{Limitations}
                        \textcolor{black}{
                            Despite the promising performance, \nickname still has several limitations, which are elaborated as follows:}
                        
                        \textcolor{black}{\textbf{Fine-grained scene understanding}. The integration of perceptual information from vision foundation models (VFMs) enhances the fine-grained scene understanding capabilities of \nickname. \textcolor{\mycolor}{Nevertheless, current VFMs continue to face difficulties in distinctly differentiating between various object components. This limitation impedes efficient skill acquisition and adaptation.} As depicted in Fig. \ref{Overview}, distinguishing the pan's surface from its handle remains challenging, {and hinders direct skill transfer from sauce spreading to pan brushing. We anticipate that future advancements in VFMs will address this challenge.}}
                        
                        \textcolor{black}{\textbf{Object reconstruction}. \textcolor{\mycolor}{We perform object reconstruction on previously unseen objects to enhance the generalizability of our method to novel environments}. However, (\uppercase\expandafter{\romannumeral1}) the limited number of object viewpoints in human manipulation videos obstructs object reconstruction solely from these videos, necessitating additional video recordings for reconstruction. With the rapid development of the 3D generation field, the ability of object reconstruction methods \citep{szymanowicz2024splatter, shen2023anything, pan2024fast} to generate 3D models from a single view or sparse views has significantly improved. \textcolor{\mycolor}{We anticipate that future applications will fully exploit these capabilities to efficiently and directly reconstruct high-quality 3D models from human demonstration videos. }(\uppercase\expandafter{\romannumeral2}) Due to the difficulty of depth sensors in accurately estimating the depth of certain objects, such as transparent or thin objects, depth estimation methods \citep{dai2022domain, yang2024depth,wu2018full, ihrke2010transparent} are typically required for object reconstruction. }

    \section{Conclusion} \label{Conclu}
    \textcolor{black}{In this paper, we present \nickname, \textcolor{\mycolor}{a novel framework that leverages foundation models to acquire fine-grained action-level skills from a limited set of human demonstration videos and robustly generalize them to previously unseen environments}. \nickname first extracts human-object interactions from human videos, distilling these interactions into keypoints and waypoints, subsequently learning skills through hierarchical constraint representations. {Furthermore, these skills are adapted to unseen environments via keypoint transfer and an iterative comparison strategy.} For high-precision tasks with stringent constraints, the skill refiner optimizes the interactions and pose estimation results, enhancing the ability of \nickname in even high-precision tasks. Extensive experiments conducted on various manipulation tasks, including challenging high-precision and long-horizon tasks, demonstrate the superior performance achieved by our \nickname, utilizing a limited number of human videos without requiring additional training, and exhibiting strong skill acquisition and adaptation capabilities. }

    \section{Funding}
    \textcolor{\mycolor}{This work was supported by the Natural Science Foundation of China under Grant 62233002, 62473050, 92370203, 624B2025.}

                        {\small
                            \bibliographystyle{agsm}
                            \bibliography{egbib}    
                        }

                    \end{document}